\documentclass{article} 
\usepackage{iclr2027_conference,times}

\usepackage{amsmath,amsfonts,bm}

\def\eqref#1{equation~\ref{#1}}

\def\1{\bm{1}}

\DeclareMathAlphabet{\mathsfit}{\encodingdefault}{\sfdefault}{m}{sl}
\SetMathAlphabet{\mathsfit}{bold}{\encodingdefault}{\sfdefault}{bx}{n}

\usepackage{hyperref}
\usepackage{url}
\usepackage{graphicx}
\usepackage{multirow}
\usepackage{tabularx}
\usepackage{booktabs}
\usepackage[T1]{fontenc}
\usepackage{listings}
\usepackage{upquote}
\usepackage[most]{tcolorbox}
\usepackage{subcaption}

\title{Time in the Residual Stream: How Language Models Encode Temporal Horizon}
\author{Vansh Gupta, \quad Justin Shenk \thanks{Code and dataset will be released upon publication} \\
Independent Researcher, Independent Researcher \\
\texttt{\{vansh1401876, shenk.justin\}@gmail.com}
}

\iclrfinalcopy 
\begin{document}

\maketitle

\begin{abstract}
Temporal Horizon understanding is crucial for Large Language Models (LLMs) 
to be able to execute temporal reasoning and planning tasks. However, 
its representational understanding remains fairly underexplored. We study 
temporal horizon representations in seven LLMs (1.7B–14B parameters) 
spanning four architecture families, comparing both within- and across-family behavior.
Projecting graded-horizon representations onto probe-derived temporal directions, 
we find that horizon is encoded approximately log-linearly at every depth 
(mean $R^2 > 0.90$ across models), a pattern analogous to the Weber-Fechner law of 
psychophysics. Geometric properties of these representations such as tortuosity, 
Centered Kernel Alignment and Betti numbers reveal consistent patterns which indicate 
progressive changes in temporal horizon geometry as the depth of the model increases.
Finally, comparing linear and manifold steering paths by Bhattacharyya energy, we find 
the two are indistinguishable at most depths, indicating that the simpler linear intervention 
loses little naturalness relative to manifold-based steering.
\end{abstract}

\section{Introduction}

What a model/agent does is largely set by the time scale over which it expects its actions to matter. The
same instruction demands a different plan at an afternoon than at a decade, and a model that
misjudges it over-weights immediate reward or defers action needed now. Yet it is rarely
stated, and must be inferred. Large Language Models (LLMs)
evaluated on temporal planning perform subpar \citep{valmeekam2023planningabilitiesofllms},
benchmarks quantify the gap \citep{fatemi2025testoftime,wei2025timemultilevelbenchmarktemporal},
frameworks improve it in smaller models \citep{liu2025timer1comprehensivetemporalreasoning}, and
models show present- versus future-oriented preferences
\citep{mazyaki2025temporalpreferenceslanguagemodels}. Horizon is also a clean object of study in
itself. Interpretability has mostly characterised discrete features and their circuits
\citep{wang2022interpretabilitywildcircuitindirect,olsson2022context}, whereas horizon is a
\emph{continuous} magnitude spanning five orders of magnitude and tied to no single token. It
therefore tests whether the linear, low-dimensional picture assumed by representation engineering
\citep{zou2025representationengineeringtopdownapproach,probetraining} extends to graded quantities,
and what shape a concept manifold \citep{wurgaft2026manifoldsteeringrevealsshared} takes when its
levels are ordered rather than categorical.

\begin{figure}[tb]
    \centering
    \begin{subfigure}[b]{0.32\linewidth}
        \centering
        \includegraphics[width=\linewidth]{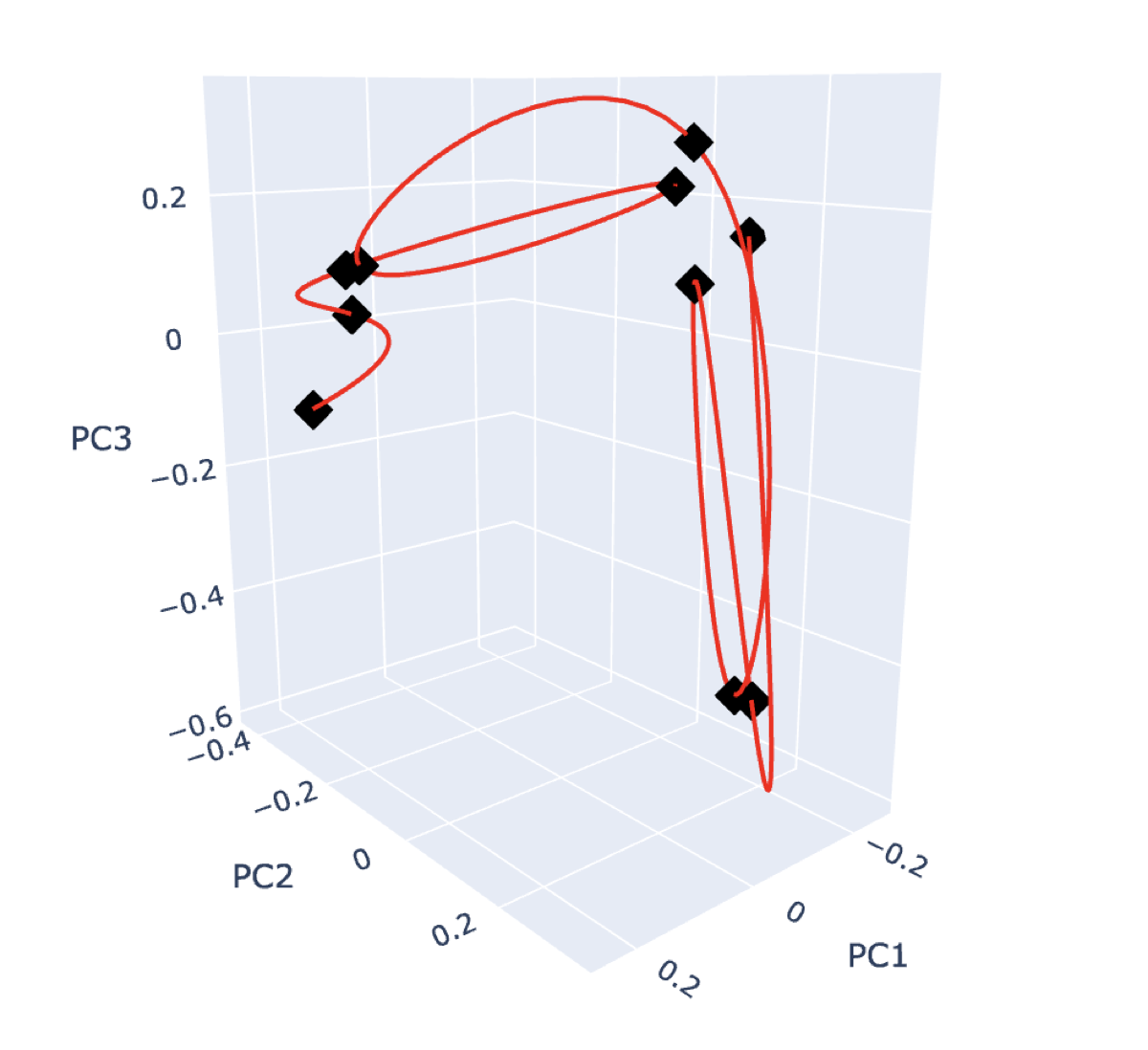}
        \caption{Layer 2 (early)}
        \label{fig:intro_manifold_layer2}
    \end{subfigure}
    \hfill
    \begin{subfigure}[b]{0.32\linewidth}
        \centering
        \includegraphics[width=\linewidth]{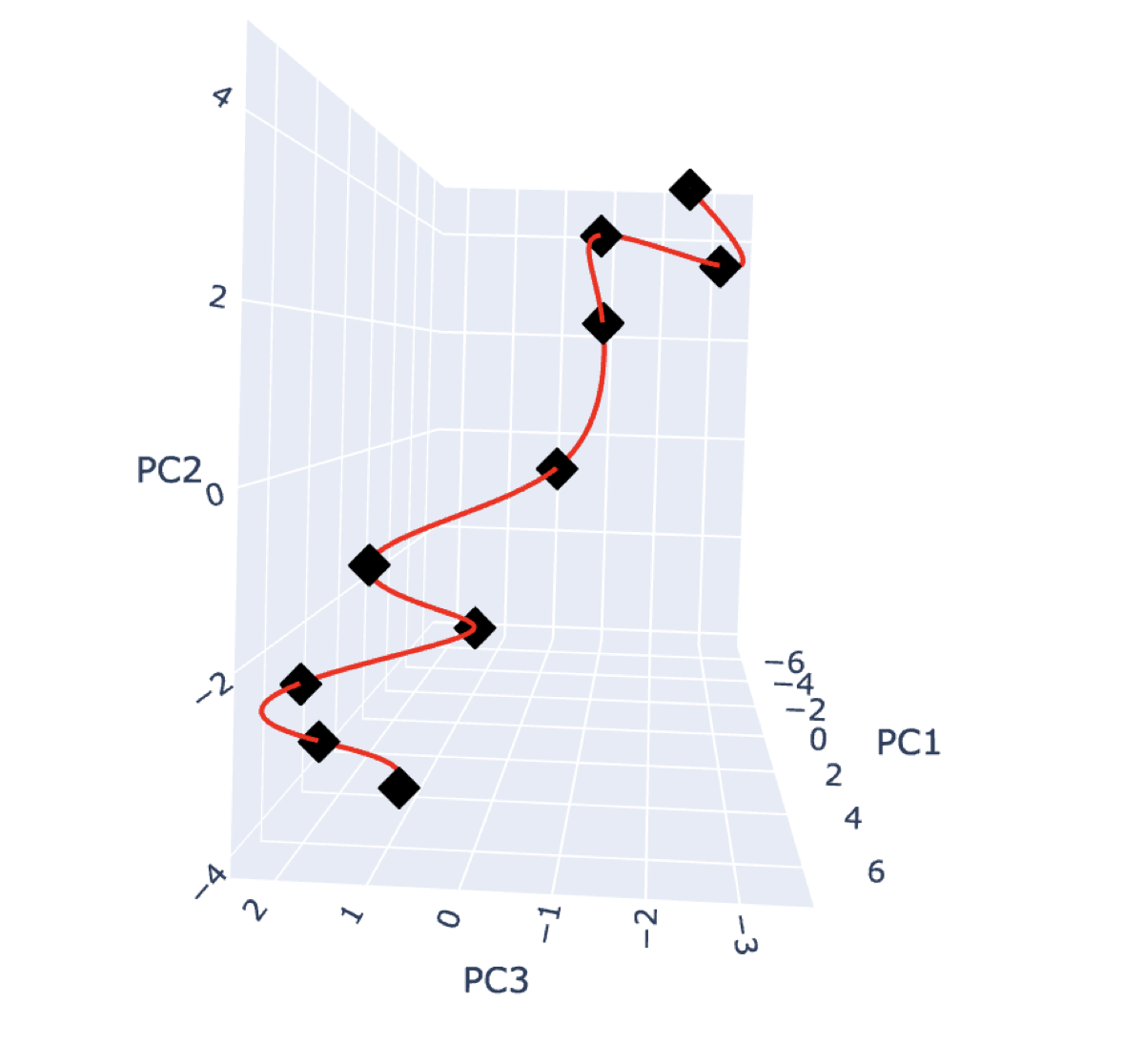}
        \caption{Layer 16 (middle)}
        \label{fig:intro_manifold_layer16}
    \end{subfigure}
    \hfill
    \begin{subfigure}[b]{0.32\linewidth}
        \centering
        \includegraphics[width=\linewidth]{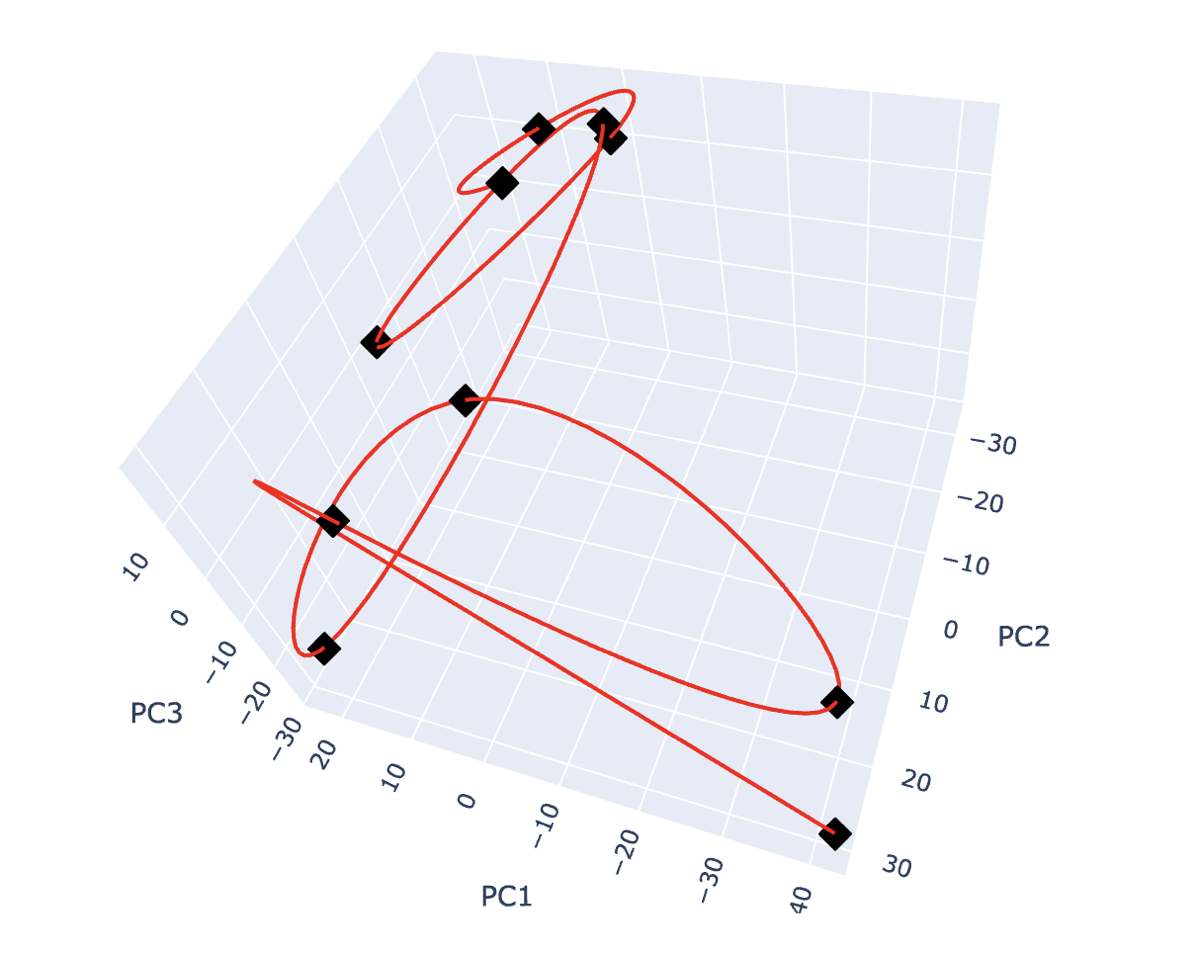}
        \caption{Layer 34 (late)}
        \label{fig:intro_manifold_layer34}
    \end{subfigure}
    \caption[Evolution of the temporal horizon manifold across depth in Qwen3-4B.]{Evolution of the temporal horizon manifold across depth in Qwen3-4B.
    Black diamonds are the mean representations of the ten graded horizon levels,
    spanning one day to five hundred years; the red curve is the fitted trajectory
    through them, projected onto the top three principal components. The arc is
    highly entangled and self-intersecting at layer 2, straightens into an ordered,
    near-one-dimensional curve by layer 16, and re-curves into a looping structure
    by layer 34. Note that
    the axis scales differ by roughly two orders of magnitude across the three
    panels, reflecting the monotone growth of activation norm with depth.}
    \label{fig:geometry_figures}
\end{figure}

Prior temporal studies mostly measure what models output; the work that does probe temporal
representations targets distance between calendar years \citep{li2025mindlanguagemodelsexhibit},
not duration. Consequently, this paper investigates the temporal horizon through both geometrical and
representational lenses (Figure~\ref{fig:geometry_figures}), across seven LLMs: Gemma2-2B
\citep{gemma2technicalrepoart}, Llama-3.1-8B \citep{grattafiori2024llama3herdmodels}, and Pythia-12B
\citep{pythiatechnicalreport}, spanning three families, along with Qwen3-1.7B, Qwen3-4B, Qwen3-8B and
Qwen3-14B \citep{qwen3technicalreport}. Both motivations meet in the answer: a horizon carried on a
recoverable direction can be read out to audit what a model inferred, and steered to correct it
\citep{zhu2025steeringriskpreferenceslarge}.

The contributions of this paper are as follows:
\begin{itemize}
    \item We characterise geometrically and representationally how the temporal horizon is encoded
    across seven LLMs and three families, as a testbed for continuous abstract magnitude.
    \item We establish that LLMs encode the temporal horizon in a log-linear manner, paralleling
    the Weber--Fechner law \citep{fechner1860elemente} in human perception.
    \item We compare manifold steering \citep{wurgaft2026manifoldsteeringrevealsshared} against
    linear steering and find the latter sufficient, so the horizon can be adjusted along a single
    direction rather than a fitted manifold.
\end{itemize}
\section{Related Works}

Recent studies evaluate temporal reasoning and planning capabilities in Large Language Models (LLMs). \citet{fatemi2025testoftime} introduces contamination-free synthetic benchmarks for temporal reasoning, while \citet{valmeekam2023planningabilitiesofllms} assesses autonomous and model-assisted planning performance. Extending these evaluations to complex scenarios, \citet{wei2025timemultilevelbenchmarktemporal} addresses intensive temporal information, evolving event dynamics, and social temporal dependencies.

Beyond benchmark reasoning, contemporary research investigates fine-grained temporal behaviors and model frameworks. \citet{herel2025timeawarenesslargelanguage} evaluates day-level temporal fact retention, showing base models often surpass instruction-tuned counterparts in time-sensitive recall. Addressing model efficiency, \citet{liu2025timer1comprehensivetemporalreasoning} equips $\sim$3B parameter models with temporal understanding, prediction, and generation capabilities, while \citet{mazyaki2025temporalpreferenceslanguagemodels} adapts psychological frameworks to examine present- versus future-oriented decision preferences. Closest to us, a psychophysical line imports the similarity-judgment paradigm: \citet{marjieh2025numberlargelanguagemodel} report logarithmic compression of \emph{number}, and \citet{li2025mindlanguagemodelsexhibit} extend this to \emph{calendar years}, finding judgments compress logarithmically about a subjective reference year and that layer-wise probes move from numeric to reference-centred codes with depth---a logarithmic mental line of the kind \citet{DEHAENE2003145} describes in humans. Both measure elicited similarity ratings between \emph{time points}; we instead measure the representation of \emph{duration}, and test it against an ordinal linear-in-rank baseline that separates metric magnitude from mere ordering.

Despite these advancements, the latent geometry encoding of the temporal horizon remains under-explored. Representation engineering \citep{zou2025representationengineeringtopdownapproach} provides a transparency framework for population-level representations, and latent steering techniques can effectively modify behavioral traits such as risk preference \citep{zhu2025steeringriskpreferenceslarge}. Our work bridges these lines of inquiry by mechanistically characterizing the latent geometry of temporal horizons and evaluating linear versus manifold steering methods \citep{wurgaft2026manifoldsteeringrevealsshared}.
\section{Methodological Overview}
 
\subsection{Dataset}

Within the experimental framework, two distinct datasets were developed to facilitate analysis. The first set, referred to as the \textbf{Horizon-Anchored} dataset, comprises three prompt variations for each scenario: ambiguous, short, and long. The ambiguous prompts incorporate essential scenario-relevant information, whereas the short and long prompts are designed to include temporal horizon information along with the exact same scenario information. The horizon is encoded in one of two ways: explicitly, through clearly defined anchor phrases naming the time scale, or implicitly, through cues from which the time scale can only be inferred.

The second set, termed the \textbf{Horizon-Graded} dataset, begins with a baseline scenario devoid of any temporal horizon information. This dataset's variations are crafted to systematically encode temporal information spanning 1 day to 500 years, thereby allowing a comprehensive exploration of the implications of varying temporal contexts on the scenarios presented. A detailed description of the datasets is present in Appendix~\ref{appendix:dataset_description}.

\section{Representational Alignment}

To analyze how short- and long-horizon representations evolve across network depth, we employ Centered Kernel Alignment (CKA) \citep{kornblith2019similarityneuralnetworkrepresentations,hsic}. CKA quantifies representational similarity between activation matrices $\mathbf{X} \in \mathbb{R}^{n \times p_1}$ and $\mathbf{Y} \in \mathbb{R}^{n \times p_2}$ by comparing their pairwise similarity structures in a manner invariant to orthogonal transformations and isotropic scaling. Formally, given Gram matrices $\mathbf{K} = \mathbf{X}\mathbf{X}^\top$ and $\mathbf{L} = \mathbf{Y}\mathbf{Y}^\top$, CKA is defined using the Hilbert-Schmidt Independence Criterion (HSIC):
\begin{equation}
    \mathrm{CKA}(\mathbf{K}, \mathbf{L}) = \frac{\mathrm{HSIC}(\mathbf{K}, \mathbf{L})}{\sqrt{\mathrm{HSIC}(\mathbf{K}, \mathbf{K}) \, \mathrm{HSIC}(\mathbf{L}, \mathbf{L})}}, \quad \text{where } \mathrm{HSIC}(\mathbf{K}, \mathbf{L}) = \frac{1}{(n-1)^2} \operatorname{tr}(\mathbf{K} \mathbf{H} \mathbf{L} \mathbf{H}),
\end{equation}
with centering matrix $\mathbf{H} = \mathbf{I}_n - \frac{1}{n} \mathbf{1}\mathbf{1}^\top$. The resulting score lies in $[0, 1]$, where $1$ denotes identical representational geometry and $0$ indicates orthogonal similarity structures. In our experiments, activations are extracted from the final token position at the output of each transformer block.

\begin{figure}[tb]
    \centering
    \begin{subfigure}[b]{1.0\linewidth}
        \centering
        \includegraphics[width=\linewidth]{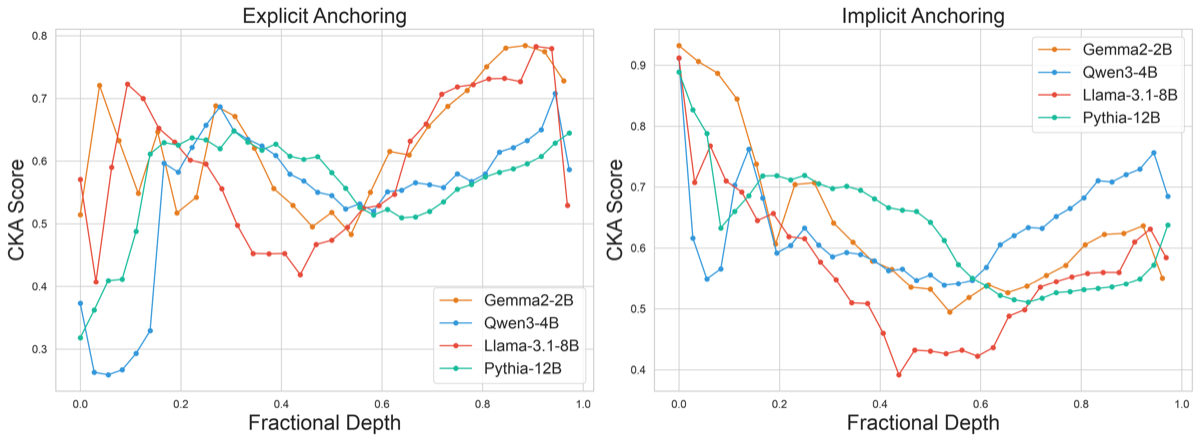}
        \caption{Across model families; Explicit (left) and Implicit (right) anchoring.}
        \label{fig:cross_family_cka}
    \end{subfigure}

    \begin{subfigure}[b]{1.0\linewidth}
        \centering
        \includegraphics[width=\linewidth]{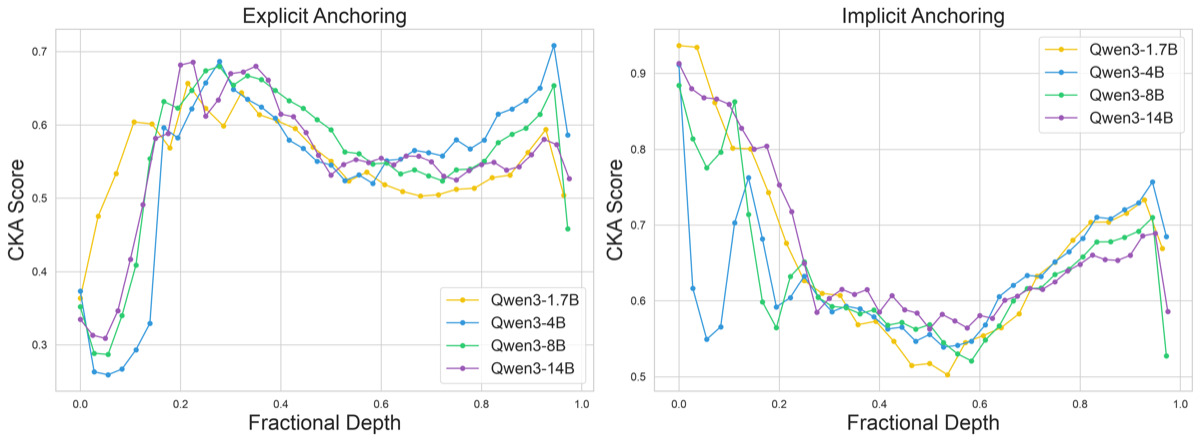}
        \caption{Within a model family; Explicit (left) and Implicit (right) anchoring.}
        \label{fig:within_family_cka}
    \end{subfigure}
    \caption[Centered Kernel Alignment (CKA) results for long and short horizon representations.]{The Centered Kernel Alignment (CKA) results show a clear pattern in the evolution of long and short horizon representations. Explicit prompt representations start low, peak as the model reaches the middle third, decline, and then rise again toward the end. In contrast, implicit prompt representations begin high, drop significantly, and then stabilize at a moderate level. This pattern is consistent across similar models and different families, with less variation seen in within-family curves compared to across-family curves.}
    \label{fig:representational_alignment_side_by_side}
\end{figure}

As depicted in Figure~\ref{fig:representational_alignment_side_by_side}, both explicit and implicit representations of temporal horizons demonstrate distinct transformations as they progress through the layers of the model. Notably, the analysis reveals a statistical trend: explicit representations begin at a lower value, then exhibit a peak and trough pattern, while implicit representations start at a higher level, move toward a trough, and ultimately stabilize in the midrange. Within-family representations, a more pronounced pattern is observed, characterized by a sudden decline toward the end. This phenomenon can be attributed to the final layers' focus on next-token prediction, which may suppress the processing of temporal information. Additional findings are detailed in Appendix~\ref{app:representation_analysis}.
\section{Logarithmic Temporal Encoding}
\label{sec:log-encoding}
We use the Horizon Graded dataset to analyze how the temporal horizon varies with the number of days. We project the mean representations of each graded horizon onto the temporal direction, which is derived from the probes trained in each layer by calculating \( W[0] - W[1] \), where \(W[0]\) and \(W[1]\) are the weights for the positive and negative classes, respectively. Since the probes were trained on a binary classification task to predict the temporal horizon, they serve as an ideal candidate for identifying the most appropriate temporal horizon direction for each layer.
\begin{table}[tb]
\centering
\caption{Mean and late-layer $R^2$ across models for the log-linear model.}
\label{tab:log_encoding_metrics}
\begin{tabular}{|l|c|c|}
\hline
\textbf{Model} & \textbf{Mean $R^2$ (log)} & \textbf{Late-layer $R^2$} \\ \hline
Qwen3-1.7B   & 0.923 & 0.936 \\ \hline
Qwen3-4B     & 0.944 & 0.974 \\ \hline
Qwen3-8B     & 0.936 & 0.956 \\ \hline
Qwen3-14B    & 0.954 & 0.983 \\ \hline
Llama-3.1-8B & 0.953 & 0.972 \\ \hline
Gemma2-2B    & 0.921 & 0.946 \\ \hline
Pythia-12B   & 0.905 & 0.905 \\ \hline
\end{tabular}
\end{table}

\begin{figure}[tb]
    \centering
    \begin{subfigure}[b]{0.9\linewidth}
        \centering
        \includegraphics[width=\linewidth]{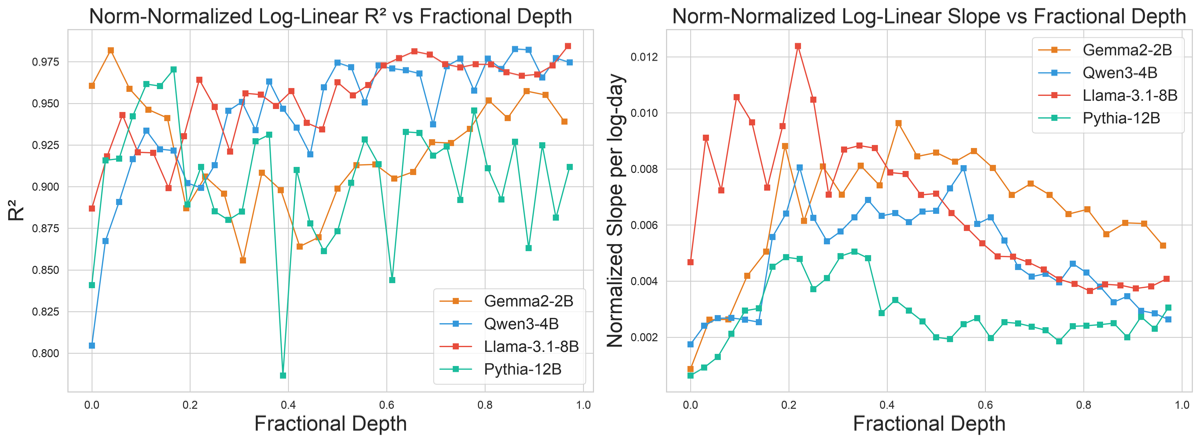}
        \caption{Across model families; $R^2$ (left) and normalized slope (right).}
        \label{fig:cross_family_log_encoding}
    \end{subfigure}

    \begin{subfigure}[b]{0.9\linewidth}
        \centering
        \includegraphics[width=\linewidth]{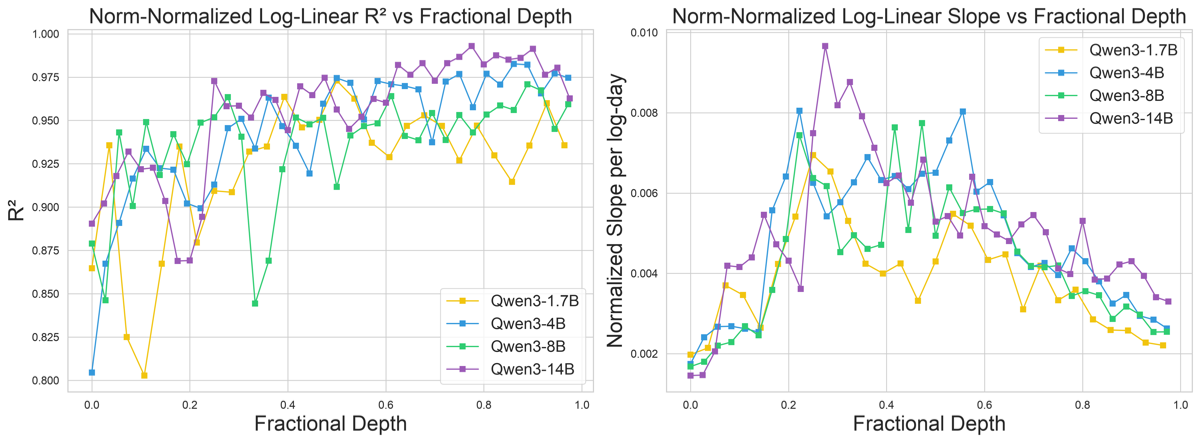}
        \caption{Within a model family; $R^2$ (left) and normalized slope (right).}
        \label{fig:within_family_log_encoding}
    \end{subfigure}
    \caption[Goodness of fit ($R^2$) vs depth of model and normalized slope for models within the same family.]{Goodness of fit ($R^2$) vs depth of model (left) and normalized slope of the log-linear graph vs model depth (right) for models within the same family. It is evident that as the model depth increases, the value of $R^2$ increases. The normalized slope is not monotonically increasing --- it is a hump. This shows that the intrinsic sensitivity of the representation to $\log(\text{horizon})$ is actually highest in the early-to-mid network and decays toward the output.}
    \label{fig:log_encoding_one_above_other}
\end{figure}

As presented in Figure~\ref{fig:log_encoding_one_above_other}, each layer exhibits a logarithmic relationship with the temporal horizon. This relationship is analogous to the Weber--Fechner law \citep{fechner1860elemente} in psychophysics, which suggests that the relationship between a physical stimulus and human perception follows a logarithmic pattern. In essence, as the physical intensity of a stimulus increases, the corresponding perceived sensation intensifies at a markedly slower rate.

It is noteworthy that the $R^2$ panel does not exhibit the same decay pattern. The $R^2$ values remain relatively stable or show an upward trend with depth (ranging from mid-0.9s to high-0.9s in the later layers), despite the normalized slope reaching its peak early and subsequently declining. Consequently, deeper layers maintain a clear log-linear shape; however, they do not possess as large a coefficient (relative to their own scale) as the early to mid layers. This indicates that while depth enhances the consistency of the fit, it does not yield a greater gain.

Table~\ref{tab:log_encoding_metrics} provides a summary of the average $R^2$ values across all layers. The majority of models demonstrate high $R^2$ scores (exceeding 0.9) in each layer, signifying that the temporal horizon is effectively encoded within the model's representations.

Furthermore, Appendix~\ref{appendix:log_encoding} contains a detailed statistical analysis of the Log Horizon model in comparison to four other models: power-law, linear-in-rank, piecewise-linear, and sigmoid-in-log.
\section{Geometrical and Topological Crossover}
To characterize the evolution of temporal horizon representations across network depth, we analyze the geometric and topological properties of layer-wise activation manifolds. Specifically, we quantify manifold curvature via \textit{tortuosity}, topological connectedness and loop structure via \textit{Betti numbers}, alongside activation norm scaling and centroid spacing.

\paragraph{Tortuosity ($\tau$).} 
Path tortuosity measures the geometric curvature and non-linear distortion of the temporal trajectory relative to a straight Euclidean line. For an ordered sequence of mean representation centroids $\{\boldsymbol{\mu}_1, \boldsymbol{\mu}_2, \dots, \boldsymbol{\mu}_K\}$ corresponding to graded temporal horizons $t_1 < t_2 < \dots < t_K$, we fit a natural cubic spline $\gamma : [1, K] \to \mathbb{R}^m$ that interpolates the centroids in horizon order, so that $\gamma(k) = \boldsymbol{\mu}_k$ at every integer parameter. Tortuosity is then the ratio of the arc length of this fitted manifold to the net displacement between its endpoints:
\begin{equation}
    \tau = \frac{L}{\|\boldsymbol{\mu}_K - \boldsymbol{\mu}_1\|_2},
    \qquad
    L = \int_{1}^{K} \left\lVert \gamma'(t) \right\rVert_2 \, dt .
\end{equation}
The arc length $L$ is evaluated numerically by sampling $\gamma$ at $N = 500$ equally spaced parameter values and accumulating the Euclidean distance between consecutive samples; the full construction is given in Appendix~\ref{app:manifold_construction}. A value of $\tau = 1$ corresponds to a perfectly straight, unidimensional trajectory in activation space, whereas $\tau > 1$ reflects local winding, curvature, or non-linear trajectory deviation. As representations mature in deeper layers, a reduction in $\tau$ toward unity indicates progressive geometric alignment of the temporal horizon axis.

\paragraph{Betti Numbers ($\beta_k$).} 
To examine the global topological structure of temporal activation point clouds independent of coordinate transformations, we employ persistent homology via Vietoris--Rips filtrations\citep{persistenthomology}. We track the primary Betti numbers across spatial filtration scales $\epsilon$:
\begin{itemize}
    \item \textbf{$0$-th Betti Number ($\beta_0$)}: Quantifies the number of connected components in the activation point cloud. Higher values of $\beta_0$ indicate isolated clusters or fragmented representations, while $\beta_0 = 1$ signifies a single cohesive representation manifold.
    \item \textbf{$1$-st Betti Number ($\beta_1$)}: Counts the number of one-dimensional topological holes or non-contractible loops. A value of $\beta_1 = 0$ indicates a simple, contractible curve, whereas $\beta_1 > 0$ reveals circular or looping sub-structures in the representation geometry.
\end{itemize}
Tracking persistent Betti barcodes across model depth allows us to determine whether temporal horizons undergo topological transitions---shifting from complex, multi-component loop topologies in early layers to a unified, contractible 1D manifold ($\beta_0 = 1, \beta_1 = 0$) in late layers.
Each of the variations of the temporal horizon indicator present for a datapoint in Horizon Graded Dataset is used to quantize the temporal horizon spectrum. These variations give us the 
centroids whose raw norm and spacing we calculate. 

\begin{figure}[tb]
    \centering
    \begin{subfigure}[b]{1.0\linewidth}
        \centering
        \includegraphics[width=\linewidth]{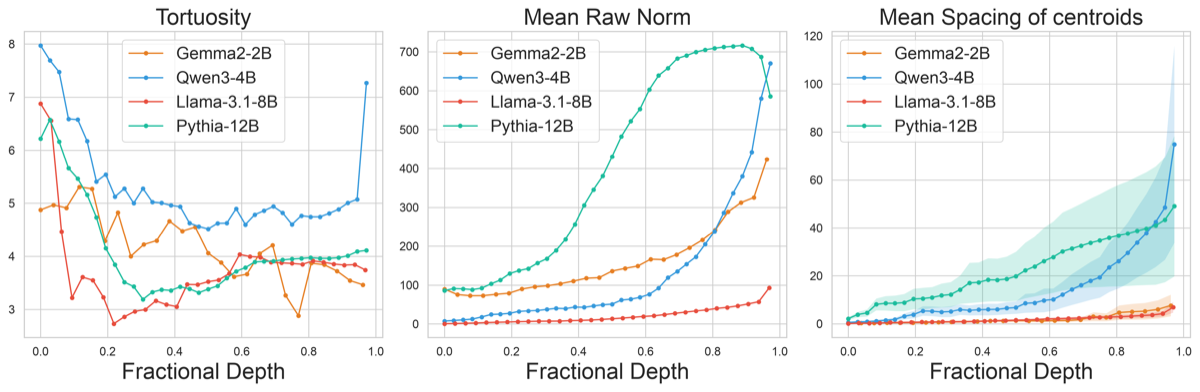}
        \caption{Across model families; activation norm, centroid spacing, and tortuosity.}
        \label{fig:cross_family_manifold_geometry}
    \end{subfigure}

    \begin{subfigure}[b]{1.0\linewidth}
        \centering
        \includegraphics[width=\linewidth]{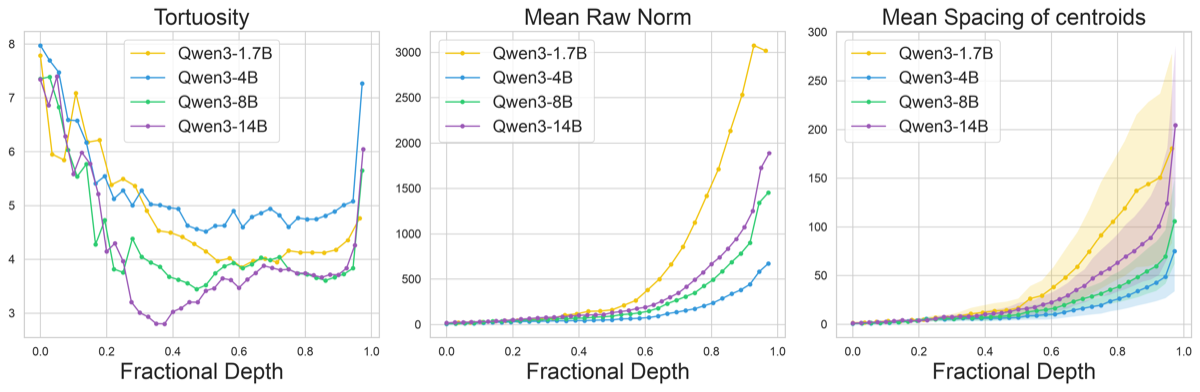}
        \caption{Within a model family; activation norm, centroid spacing, and tortuosity.}
        \label{fig:within_family_manifold_geometry}
    \end{subfigure}
    
    \caption{Layer-wise evolution of manifold geometry and topological properties across model depth for across-family (top) and within-family (bottom) evaluations.}
    \label{fig:manifold_geometry_combined}
\end{figure}

As illustrated in Figure~\ref{fig:manifold_geometry_combined}, both 
the mean activation norm and pairwise centroid spacing increase 
systematically across network depth. Path tortuosity is initially 
elevated in early layers ($\tau > 1.5$), reflecting non-linear 
representation curvature and local trajectory distortion. In 
intermediate layers, tortuosity undergoes a pronounced decline, 
converging to a stable, low plateau ($\tau \to 1$) across models 
within the same architecture family. Tortuosity never reaches $1$, which tells
us that temporal horizon geometry always contains loops and turns
and is never fully straight. Across heterogeneous models,
tortuosity remains tightly bounded within a narrow range. 
This empirical trajectory indicates that temporal representation 
centroids undergo progressive geometric untangling and manifold 
rectification across intermediate layers---forming a linear, 
low-tortuosity temporal axis---before exhibiting a slight upturn 
in terminal output-projection layers.

\begin{figure}[tb]
    \centering
    \includegraphics[width=0.99\linewidth]{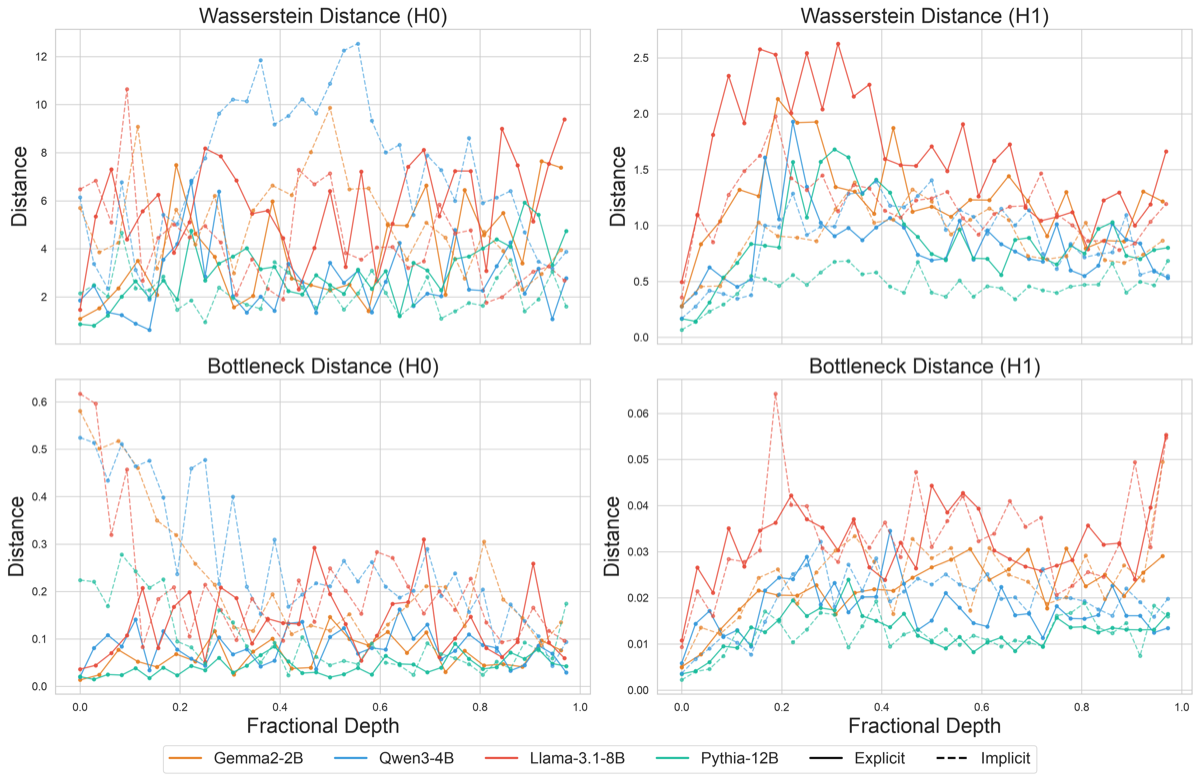}
    \caption{Persistent homology distance across normalized layer depth evaluated across heterogeneous model families.}
    \label{fig:across_persistent_homology}
\end{figure}

\begin{figure}[tb]
    \centering
    \includegraphics[width=0.99\linewidth]{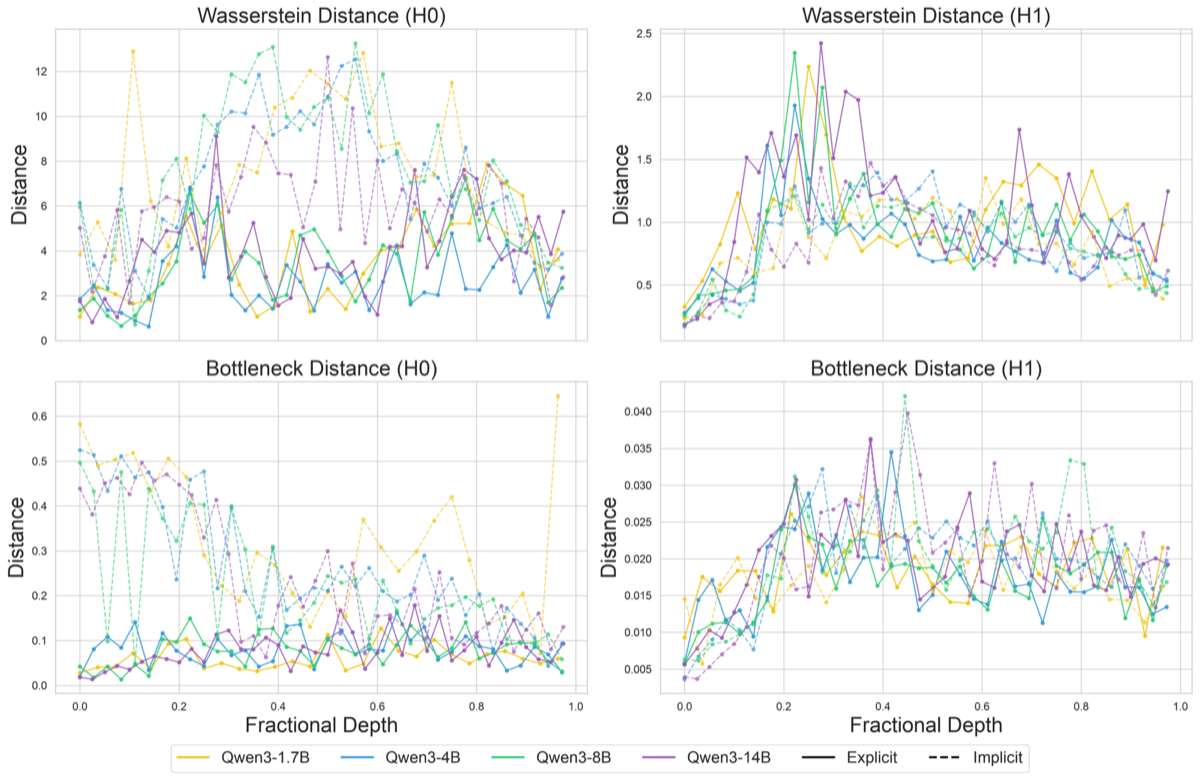}
    \caption{Persistent homology distance across normalized layer depth evaluated within individual model families.}
    \label{fig:within_persistent_homology}
\end{figure}

Figure~\ref{fig:across_persistent_homology} and Figure~\ref{fig:within_persistent_homology} report the
Wasserstein and bottleneck distances between the persistence diagrams of the shortest- and
longest-horizon point clouds, computed separately in degree $H_0$ and degree $H_1$ at every layer.
These distances measure how far apart the two horizon extremes are \emph{topologically}, and are
therefore complementary to the purely metric descriptors above.

Implicit anchoring (dashed) separates the horizon extremes more than explicit (solid) in $H_0$ across
all seven models, by $1.56$--$5.26\times$ in bottleneck distance, while $H_1$ reverses the ordering in
six of seven. The gap is transient rather than fixed: implicit bottleneck $H_0$ decays with depth in
every model, its final-third mean falling to $0.32$--$0.64$ of its first-third value against a flat or
mildly rising explicit curve, so the two converge by roughly $40\%$ depth, and the Wasserstein $H_1$
maximum correspondingly sits $0.23$ fractional depth deeper under implicit anchoring in six of seven
models. Implicit anchoring therefore postpones the depth at which the horizon distinction is
topologically resolved rather than leaving the manifold permanently more tangled. Beneath this
condition effect lies a shared profile: under explicit anchoring Wasserstein $H_1$ peaks at
$0.20$--$0.32$ fractional depth in all seven models and decays $37$--$79\%$ toward the output, with the
four Qwen3 models peaking at $0.23$--$0.28$ despite an eightfold spread in parameter count---topological
separation is maximal in the first third and governed by relative depth, not scale. These distances are
expressed in raw activation units, and activation norms grow by one to two orders of magnitude across
depth (Appendix~\ref{app:geometry_descriptors}); both trends run counter to that growth and are
consequently robust to the norm confound. Norm-normalized persistence and the remaining topological
diagnostics are given in Appendix~\ref{app:geometry_persistent_homology}.

\begin{figure}[tb]
    \centering
    \begin{subfigure}[b]{1.0\linewidth}
        \centering
        \includegraphics[width=\linewidth]{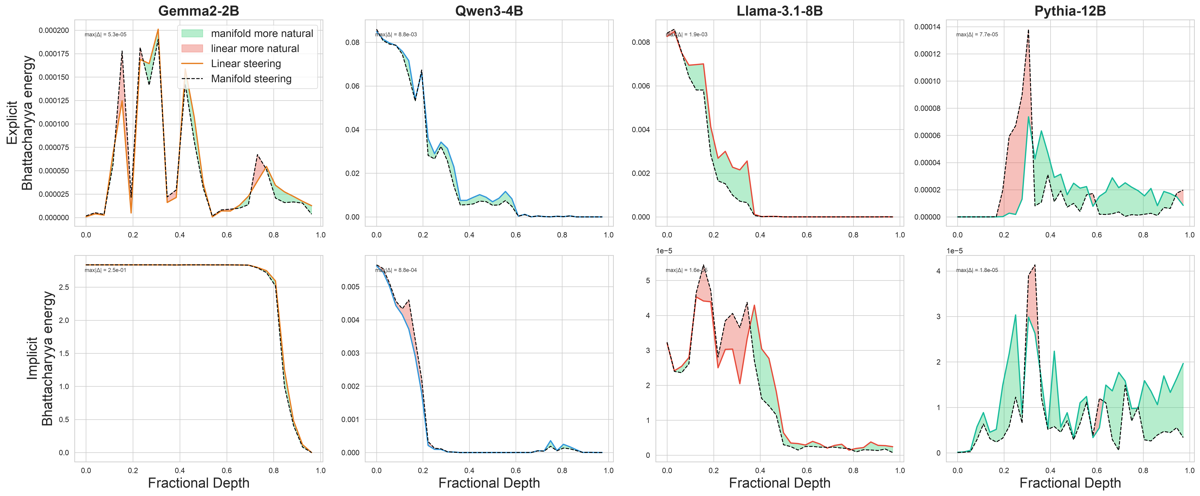}
        \caption{Across heterogeneous model families.}
        \label{fig:across_steering_energy}
    \end{subfigure}

    \begin{subfigure}[b]{1.0\linewidth}
        \centering
        \includegraphics[width=\linewidth]{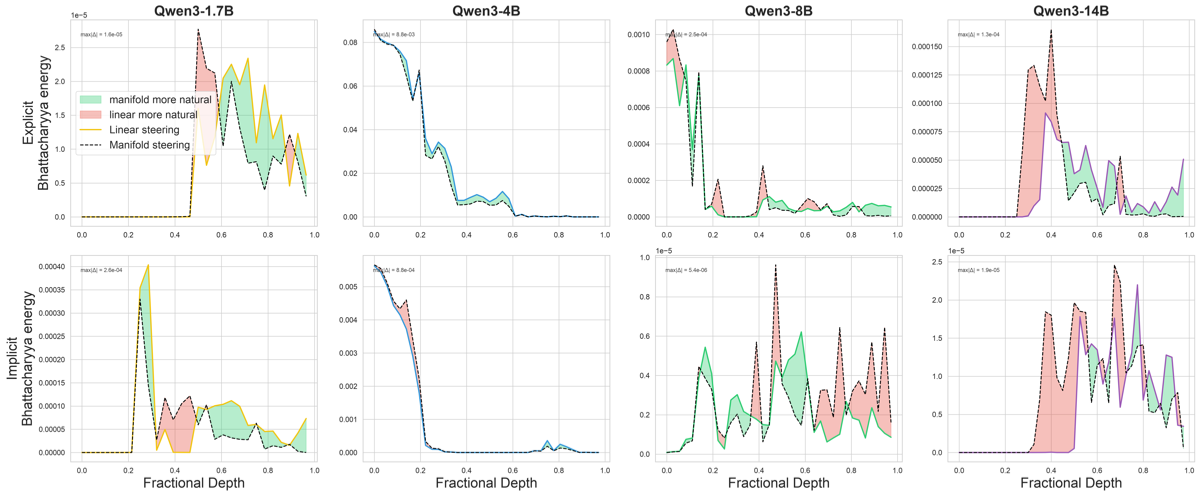}
        \caption{Within the Qwen3 model family.}
        \label{fig:within_steering_energy}
    \end{subfigure}

    \caption{Linear vs. manifold steering energy across normalized layer depth, evaluated across heterogeneous model families (top) and within the Qwen3 model family (bottom); in both, row 0 is explicit anchoring and row 1 is implicit anchoring. Green shading indicates depths where manifold steering is more natural; red shading indicates depths where linear steering is more natural.}
    \label{fig:steering_energy_combined}
\end{figure}

We use Bhattacharyya energy \citep{wurgaft2026manifoldsteeringrevealsshared} to measure the naturalness of
steering states in the models. Across all seven models and both anchoring conditions, the 
linear-interpolation and manifold-spline steering paths yield 
statistically indistinguishable naturalness energies at the 
overwhelming majority of intervention depths: in 9 of the 14 
(model, condition) pairs, the energy gap 
$\Delta E(L) = E_{\text{linear}}(L) - E_{\text{manifold}}(L)$ never 
exceeds $5\times10^{-4}$ in magnitude, consistent with numerical 
noise rather than a genuine geometric preference for either steering 
path. Five exceptions break this pattern. Gemma2-2B under implicit 
anchoring shows a dramatic preference for the manifold path 
($\max|\Delta E| = 0.248$, mean $\Delta E = +0.018$), while 
Qwen3-4B ($\max|\Delta E| = 8.8\times10^{-3}$) and Llama-3.1-8B 
($\max|\Delta E| = 1.9\times10^{-3}$) show smaller but consistent 
manifold preferences under explicit anchoring; conversely, Qwen3-4B 
reverses under implicit anchoring to a small linear preference 
($\max|\Delta E| = -8.8\times10^{-4}$), with Qwen3-8B showing a 
marginal, noise-adjacent linear preference under explicit anchoring 
($\max|\Delta E| = -2.5\times10^{-4}$). It 
indicates that the curvature of the fitted horizon manifold is
largely inconsequential to steering naturalness except in a small,
architecture-specific subset of cases. This provides secondary
evidence that the temporal horizon is close to linearly encoded in the model representations,
given that only 3 out of 14 (model, condition) pairs show a manifold preference.
It also establishes that linear steering should in fact be the go-to approach
when we try to steer the temporal horizon.

\section{Conclusion}
\label{sec:conclusions}

The primary discovery outlined in this paper is that the 
concept of temporal horizon is systematically developed 
throughout the intermediate to later stages of large 
language models (LLMs). Both representational and 
geometrical markers reveal a psychophysical law employed 
to encode the temporal horizon spectrum and an underlying mechanism that 
facilitates the encoding of temporal horizon across various 
families of models. The presence of both explicit and 
implicit temporal indicators compels these models to 
initially process representations in distinct ways. However, 
as the layers progress, the representations converge, 
allowing for a more uniform processing in the 
middle third and later third of the model. This indicates that the early differentiated 
handling of temporal information is crucial for achieving a 
consistent understanding as the model matures.

\section{Limitations}
\label{sec:limitations}

The experiments presented here constitute an extensive representational
investigation of the temporal horizon representations. Steering behavior has
only been explored from a quantitative perspective, which we deemed
sufficient to demonstrate steering responsiveness. However, a more
exhaustive claim would require a downstream qualitative behavioral
analysis. The emergence of temporal horizon understanding could also be
traced through the training dynamics, which we do not explore here
and which we believe is a self-sufficient research direction in its own right.
 
\section{Future Work}
\label{sec:future}

The most natural extensions of this work are: (i) replication across additional model families to test universality more stringently; (ii) resolving the encoding-versus-use dissociation reported in Appendix~\ref{app:head_convergence}, where the attention heads that most strongly represent the temporal horizon are almost disjoint from those that causally control it --- locating components that both encode and causally drive horizon-dependent downstream behavior remains open; and
(iii) studying the emergence of the temporal horizon during pre-training and how it is affected during post-training, and whether a geometry of time itself can be established.

\ificlrfinal
\section{Acknowledgement}
\label{sec:acknowledgement}

This research was done as part of the Alignment Research Cohort conducted by
SPAR (Supervised Program for Alignment Research). We would like to thank
them for providing us with the opportunity and financial support to
pursue this work.
\fi

\bibliography{references/iclr2026_conference}
\bibliographystyle{iclr2027_conference}

\appendix
\renewcommand{\thesection}{\Alph{section}}

\section{Dataset Description}
\label{app:dataset_description}
\label{appendix:dataset_description}

This section provides a detailed description of the synthetic dataset generated using Qwen3-32B \citep{qwen3technicalreport}, which served as the foundation for all conducted experiments. A structured template was followed to effectively isolate temporal horizon information, ensuring the delivery of precise results while minimizing potential noise from other elements of the prompt, including lexical structure and subject-object relationships.

Two datasets were constructed. The \textbf{Horizon-Anchored} dataset
(Appendix~\ref{app:horizon_anchored}) supports the binary short-versus-long
analyses --- probe training, representational alignment, and the causal
steering experiments --- while the \textbf{Horizon-Graded} dataset
(Appendix~\ref{app:horizon_graded}) supports the continuous analyses in which
the temporal horizon is swept across ten levels spanning five orders of
magnitude. Both were generated with Qwen3-32B served through vLLM\citep{vllm} using
bfloat16 precision, nucleus sampling with temperature $0.85$ and $p = 0.95$,
and a generation budget of 7000 tokens per call. Scenarios were requested in
batches of 50 per call and accumulated until the target corpus size was
reached. Every prompt template referred to below is reproduced verbatim in
Appendix~\ref{app:prompts}.

\subsection{Horizon-Anchored Dataset}
\label{app:horizon_anchored}

The dataset is structured according to a standardized template, encompassing five distinct domains: Personal Finance, Health Behavior, Career Decisions, Software Engineering, and Geopolitical Reasoning. Each data point within this dataset features three variations of prompts, with the temporal horizon expressed in two formats---either explicitly through specific keywords or implicitly using indicative phrases. Each data point is categorized as follows:

\begin{itemize}
    \item \textbf{Ambiguous:} These prompts outline the scenario without specifying the time horizon.
    \item \textbf{Long:} These prompts detail the scenario along with temporal horizon indicators that suggest a long-term outlook.
    \item \textbf{Short:} These prompts articulate the scenario with temporal horizon indicators that indicate a short-term perspective. 
\end{itemize}

This structure allows for a comprehensive analysis of ambiguous instances across diverse domains.

The two anchoring conditions, explicit and implicit, were generated as
independent corpora with separate prompt templates, so an instance belongs to
exactly one condition. Table~\ref{tab:dataset_composition} reports the size and
domain composition of each corpus. Because the generator occasionally emits a
scenario whose \texttt{short} field duplicates an earlier one, all downstream
experiments deduplicate on the \texttt{short} field; the resulting effective
corpus sizes are 7{,}469 explicit and 6{,}397 implicit instances, and these are
the counts on which every probe in this paper was trained.

\begin{table}[htbp]
\centering
\caption{Composition of the Horizon-Anchored dataset. Counts are numbers of
$\langle$ambiguous, short, long$\rangle$ triples. The final row reports the
effective size after deduplication on the \texttt{short} field, which is what
all downstream experiments consume.}
\label{tab:dataset_composition}
\begin{tabular}{|l|c|c|}
\hline
\textbf{Domain} & \textbf{Explicit} & \textbf{Implicit} \\ \hline
Personal Finance       & 1608 & 1424 \\ \hline
Health Behavior        & 1568 & 1424 \\ \hline
Career Decisions       & 1507 & 1419 \\ \hline
Software Engineering   & 1486 & 1404 \\ \hline
Geopolitical Reasoning & 1450 & 1393 \\ \hline
\textbf{Total generated}  & \textbf{7619} & \textbf{7064} \\ \hline
\textbf{After deduplication} & \textbf{7469} & \textbf{6397} \\ \hline
\end{tabular}
\end{table}

\paragraph{Explicit anchoring.}
In the explicit condition the \texttt{ambiguous} prompt is a complete,
horizon-neutral declarative sentence, and the \texttt{short} and \texttt{long}
variants are produced by inserting exactly one anchor phrase drawn from a
closed vocabulary, leaving every other token untouched. The permitted anchors
and their empirical frequencies are listed in
Table~\ref{tab:anchor_vocabulary}. Insertion occurs at one of two positions:
sentence-final (3784 instances), as in \textit{``The household should allocate
funds to emergency savings this afternoon.''}, or sentence-initial (3716
instances), as in \textit{``Today, the individual should track monthly
expenses.''}. The two positions are close to balanced by construction, and the
short and long variants of a given instance use the same position in
$82.7\%$ of cases.

The generation template, given in full as Listing~\ref{lst:prompt_explicit},
additionally forbids a drift vocabulary of urgency,
strategy, and frequency terms --- among them \textit{current}, \textit{crisis},
\textit{urgent}, \textit{temporary}, \textit{weekly}, \textit{sustainable},
\textit{strategic}, \textit{tactical}, \textit{permanent},
\textit{generational}, \textit{later}, \textit{eventually}, \textit{quickly},
\textit{soon}, \textit{immediately}, \textit{short-term}, \textit{long-term},
and \textit{ongoing}. This ensures the only signal distinguishing the short and
long variants is the temporal anchor itself, and not a correlated register of
urgency or strategic framing. Empirically the construction is tight: deleting
the anchor phrase recovers the \texttt{ambiguous} sentence verbatim in
$97.5\%$ of short variants and $96.9\%$ of long variants, and all three fields
terminate in a period in $100\%$ of instances.

\begin{table}[htbp]
\centering
\caption{Closed anchor vocabulary for the explicit condition, with the number of
instances using each phrase. Only these phrases are permitted; the residual
counts (119 short, 90 long) are instances where the generator deviated from the
prescribed vocabulary.}
\label{tab:anchor_vocabulary}
\begin{tabular}{|l|c|l|c|}
\hline
\textbf{Short anchor} & \textbf{Count} & \textbf{Long anchor} & \textbf{Count} \\ \hline
``this week''              & 1381 & ``in fifty years''            & 1603 \\ \hline
``by the end of the day''  & 1299 & ``in a decade''               & 1594 \\ \hline
``today''                  & 1225 & ``by the end of the century'' & 1270 \\ \hline
``this afternoon''         & 1036 & ``in five years''             & 1246 \\ \hline
``right now''              &  949 & ``next year''                 & 1035 \\ \hline
``this morning''           &  873 & ``over the next generation''  &  781 \\ \hline
``before tonight''         &  737 & ---                           &  --- \\ \hline
\end{tabular}
\end{table}

\paragraph{Implicit anchoring.}
In the implicit condition no temporal expression appears anywhere in the
prompt. The \texttt{ambiguous} field is a subject-and-verb stub such as
\textit{``The individual is''}, and the \texttt{short} and \texttt{long}
variants complete that stub by naming an \emph{activity} whose typical duration
is under one week or over one year respectively, so that the horizon is
recoverable only from world knowledge about the activity. The generation
template, given in full as Listing~\ref{lst:prompt_implicit}, enforces a banned
vocabulary that is a strict superset of the explicit
anchor set --- covering every time-denoting noun, adjective, and adverb
(\textit{time}, \textit{duration}, \textit{now}, \textit{soon}, \textit{hour},
\textit{day}, \textit{year}, \textit{decade}, \textit{brief},
\textit{permanent}, and so on) --- which guarantees that the surface vocabulary
of the two conditions is disjoint. Two further constraints keep the pair
comparable: the short and long completions must match within $\pm 2$ tokens
(satisfied in $93.4\%$ of instances), and only the activity noun phrase may
differ between them.

A design property worth stating explicitly is that the implicit stems are
deliberately generic and therefore heavily shared: the corpus contains 7064
instances but only 711 distinct \texttt{ambiguous} stems, with
\textit{``The individual is''} occurring 584 times and \textit{``The patient
is''} 511 times. This is a consequence of requiring the stem to name no
outcome at all; it means the implicit condition varies almost entirely in the
completion, and the \texttt{ambiguous} field acts as a near-constant baseline
rather than a per-instance control. The stem is a literal prefix of both
completions in $93.9\%$ of instances.

Table~\ref{tab:example_instances} shows one representative instance from each
condition, reproduced verbatim from the released corpora. The contrast between
the two rows illustrates the design: the explicit instance differs only by an
inserted anchor phrase, whereas the implicit instance differs only by the named
activity.

\begin{table}[htbp]
\centering
\caption{Representative instances from the Horizon-Anchored dataset, one per
anchoring condition, reproduced verbatim. Both are drawn from the Personal
Finance domain.}
\label{tab:example_instances}
\begin{tabularx}{\linewidth}{|l|l|X|}
\hline
\textbf{Condition} & \textbf{Field} & \textbf{Prompt} \\ \hline
\multirow{3}{*}{Explicit} & Ambiguous & The household should allocate funds to emergency savings. \\ \cline{2-3}
                          & Short     & The household should allocate funds to emergency savings \textbf{this afternoon}. \\ \cline{2-3}
                          & Long      & The household should allocate funds to emergency savings \textbf{in five years}. \\ \hline
\multirow{3}{*}{Implicit} & Ambiguous & The household is \\ \cline{2-3}
                          & Short     & The household is \textbf{covering seasonal utility costs}. \\ \cline{2-3}
                          & Long      & The household is \textbf{developing a tax-optimized estate plan}. \\ \hline
\end{tabularx}
\end{table}

Table~\ref{tab:field_lengths} summarizes the surface length of each field. The
explicit condition produces longer prompts overall because the ambiguous stem
is already a full sentence, while the implicit stem is only three words on
average. Within each condition the short and long variants are closely matched
in length, so any representational separation observed between them cannot be
attributed to a systematic difference in sequence length.

\begin{table}[htbp]
\centering
\caption{Surface length statistics (whitespace-delimited words) for each field
of the Horizon-Anchored dataset.}
\label{tab:field_lengths}
\begin{tabular}{|l|c|c|c|c|c|c|}
\hline
\multirow{2}{*}{\textbf{Field}} & \multicolumn{3}{c|}{\textbf{Explicit}} & \multicolumn{3}{c|}{\textbf{Implicit}} \\ \cline{2-7}
 & \textbf{Mean} & \textbf{Min} & \textbf{Max} & \textbf{Mean} & \textbf{Min} & \textbf{Max} \\ \hline
Ambiguous & 7.8  & 4 & 15 & 3.2 & 3 & 8  \\ \hline
Short     & 10.3 & 4 & 20 & 8.0 & 2 & 13 \\ \hline
Long      & 11.3 & 5 & 21 & 8.4 & 3 & 16 \\ \hline
\end{tabular}
\end{table}

\subsection{Behavioral Validation and Model Filtering}
\label{app:model_filtering}

Before any representational analysis, each model was checked for whether it can
read the temporal horizon behaviorally at all. Every \texttt{short} and every
\texttt{long} prompt was presented inside a three-shot classification template
(Listing~\ref{lst:prompt_model_eval})
asking the model to label the situation as \textit{short} or \textit{long}, and
the prediction was taken as the argmax over the next-token logits of the two
label tokens. This yields $2 \times 7619 = 15{,}238$ explicit and $2 \times
7064 = 14{,}128$ implicit trials per model. Triples classified correctly were
retained as the model-filtered corpus used by the rank and intrinsic
dimensionality experiments.

The resulting accuracies, reported in Table~\ref{tab:model_filtering}, double as
a zero-shot behavioral result. Two patterns are worth noting. First, implicit
anchoring is harder than explicit anchoring for six of the seven models, with
the gap reaching 17 points for Gemma2-2B and 11 points for Qwen3-1.7B; this is
consistent with the persistent-homology finding that implicit anchoring
postpones the depth at which the horizon distinction is topologically resolved.
Second, Pythia-12B is a clear outlier, performing near chance on explicit
anchoring ($62.7\%$) despite being among the largest models studied, and it is
the only model whose implicit accuracy exceeds its explicit accuracy. Pythia-12B
is likewise the weakest model on the log-linear goodness-of-fit reported in the
main text, so its behavioral weakness and its representational weakness
coincide.

\begin{table}[htbp]
\centering
\caption{Zero-shot short-versus-long classification accuracy under the
three-shot logit-comparison protocol. Accuracies are over 15{,}238 explicit and
14{,}128 implicit trials per model.}
\label{tab:model_filtering}
\begin{tabular}{|l|c|c|}
\hline
\textbf{Model} & \textbf{Explicit acc.} & \textbf{Implicit acc.} \\ \hline
Qwen3-1.7B   & 85.8\% & 75.2\% \\ \hline
Qwen3-4B     & 92.4\% & 89.4\% \\ \hline
Qwen3-8B     & 94.3\% & 89.2\% \\ \hline
Qwen3-14B    & 92.0\% & 88.9\% \\ \hline
Llama-3.1-8B & 94.8\% & 87.2\% \\ \hline
Gemma2-2B    & 87.2\% & 70.4\% \\ \hline
Pythia-12B   & 62.7\% & 69.7\% \\ \hline
\end{tabular}
\end{table}

\subsection{Horizon-Graded Dataset}
\label{app:horizon_graded}

The Horizon-Graded dataset replaces the binary short/long contrast with a
graded sweep. Each instance consists of a base scenario describing an event
that is being awaited, together with ten variations that are identical
word-for-word except for the temporal phrase naming when the event occurs; the
generation template that enforces this structure is given as
Listing~\ref{lst:prompt_graded}. The
ten horizon levels, the surface phrase realizing each level, and the numeric
day-value used for every log-linear fit in this paper are given in
Table~\ref{tab:graded_horizons}. The levels are approximately geometrically
spaced and span five orders of magnitude, from one day to five hundred years,
which is what makes them suitable for testing logarithmic versus linear
encoding.

\begin{table}[htbp]
\centering
\caption{The ten horizon levels of the Horizon-Graded dataset. The day-values in
the final column are the horizon magnitudes used as the independent variable in
all log-linear, power-law, and rank-based fits.}
\label{tab:graded_horizons}
\begin{tabular}{|l|l|r|}
\hline
\textbf{Level} & \textbf{Surface phrase} & \textbf{Days} \\ \hline
\texttt{1\_day}     & ``tomorrow''               & 1       \\ \hline
\texttt{1\_week}    & ``next week''              & 7       \\ \hline
\texttt{1\_month}   & ``next month''             & 30      \\ \hline
\texttt{6\_months}  & ``in six months''          & 182     \\ \hline
\texttt{1\_year}    & ``next year''              & 365     \\ \hline
\texttt{5\_years}   & ``in five years''          & 1{,}825   \\ \hline
\texttt{10\_years}  & ``in a decade''            & 3{,}650   \\ \hline
\texttt{50\_years}  & ``in fifty years''         & 18{,}250  \\ \hline
\texttt{100\_years} & ``in a century''           & 36{,}500  \\ \hline
\texttt{500\_years} & ``in five hundred years''  & 182{,}500 \\ \hline
\end{tabular}
\end{table}

The corpus contains 543 instances, of which 287 have distinct base scenarios;
experiments load the first 500 instances by default. A representative instance,
reproduced verbatim, is shown in Table~\ref{tab:graded_example}. Because the ten
variations of an instance differ only in the temporal phrase, the mean
representation of each horizon level can be compared directly against the
others without controlling for scenario content, which is precisely the
property the continuous interpolation and Weber--Fechner analyses rely on.

\begin{table}[htbp]
\centering
\caption{A representative instance from the Horizon-Graded dataset, reproduced
verbatim. All ten variations share an identical scenario stem and differ only
in the temporal phrase.}
\label{tab:graded_example}
\begin{tabularx}{\linewidth}{|l|X|}
\hline
\textbf{Level} & \textbf{Variation} \\ \hline
\texttt{1\_day}     & The family is eagerly waiting for their child to be born \textbf{tomorrow}. \\ \hline
\texttt{1\_week}    & The family is eagerly waiting for their child to be born \textbf{next week}. \\ \hline
\texttt{1\_month}   & The family is eagerly waiting for their child to be born \textbf{next month}. \\ \hline
\texttt{6\_months}  & The family is eagerly waiting for their child to be born \textbf{in six months}. \\ \hline
\texttt{1\_year}    & The family is eagerly waiting for their child to be born \textbf{next year}. \\ \hline
\texttt{5\_years}   & The family is eagerly waiting for their child to be born \textbf{in five years}. \\ \hline
\texttt{10\_years}  & The family is eagerly waiting for their child to be born \textbf{in a decade}. \\ \hline
\texttt{50\_years}  & The family is eagerly waiting for their child to be born \textbf{in fifty years}. \\ \hline
\texttt{100\_years} & The family is eagerly waiting for their child to be born \textbf{in a century}. \\ \hline
\texttt{500\_years} & The family is eagerly waiting for their child to be born \textbf{in five hundred years}. \\ \hline
\end{tabularx}
\end{table}

\subsection{Models Studied}
\label{app:models_studied}

All experiments were run over the seven models listed in
Table~\ref{tab:model_registry}, spanning three architecture families and a
parameter range from 1.7B to 14B. Four models come from the Qwen3 family and
support the within-family scaling comparisons; the remaining three, together
with Qwen3-4B, support the across-family comparisons. Layer counts differ across
models, so all layer-indexed results in this paper are reported against
fractional depth rather than absolute layer index.

\begin{table}[htbp]
\centering
\caption{The seven models studied, with their transformer block counts and
residual stream widths. Layer counts vary from 26 to 40, motivating the use of
fractional depth throughout the paper.}
\label{tab:model_registry}
\begin{tabular}{|l|l|c|c|}
\hline
\textbf{Model} & \textbf{Family} & \textbf{Layers} & \textbf{$d_{\text{model}}$} \\ \hline
Qwen3-1.7B   & Qwen3  & 28 & 2048 \\ \hline
Qwen3-4B     & Qwen3  & 36 & 2560 \\ \hline
Qwen3-8B     & Qwen3  & 36 & 4096 \\ \hline
Qwen3-14B    & Qwen3  & 40 & 5120 \\ \hline
Llama-3.1-8B & Llama  & 32 & 4096 \\ \hline
Gemma2-2B    & Gemma  & 26 & 2304 \\ \hline
Pythia-12B   & Pythia & 36 & 5120 \\ \hline
\end{tabular}
\end{table}

\subsection{Prompts}
\label{app:prompts}

This subsection reproduces, verbatim, the four prompt templates used to
construct and validate the datasets. Two are generation templates for the
Horizon-Anchored corpus, one under implicit anchoring
(Listing~\ref{lst:prompt_implicit}) and one under explicit anchoring
(Listing~\ref{lst:prompt_explicit}); one is the generation template for the
Horizon-Graded corpus (Listing~\ref{lst:prompt_graded}); and one is the
three-shot classification template used for the behavioral validation of
Appendix~\ref{app:model_filtering} (Listing~\ref{lst:prompt_model_eval}). In the
last of these, \texttt{\{prompt\}} is the placeholder into which each
\texttt{short} or \texttt{long} sentence is substituted, and the trailing
quotation mark is deliberate: it constrains the model to emit the label token
immediately.

\providecommand{\iclrroot}{}

\lstdefinestyle{promptbox}{
    basicstyle=\ttfamily\scriptsize,
    breaklines=true,
    breakautoindent=false,
    columns=fullflexible,
    keepspaces=true,
    showstringspaces=false,
    frame=none,
    aboveskip=0pt,
    belowskip=0pt,
    literate={—}{{\textemdash}}1 {±}{{$\pm$}}1 {→}{{$\rightarrow$}}1
}

\newtcolorbox[auto counter]{promptcontainer}[2][]{
    breakable,
    enhanced jigsaw,
    colback=white,
    colframe=black!65,
    boxrule=0.5pt,
    left=4pt, right=4pt, top=4pt, bottom=4pt,
    fonttitle=\bfseries\small,
    title={Listing~\thetcbcounter: #2},
    #1
}

\begin{promptcontainer}[label={lst:prompt_implicit}]{Horizon-Anchored generation prompt, implicit anchoring}
\begin{lstlisting}[style=promptbox]
[SYSTEM]
You are a deterministic data generation engine. Output valid JSON only.

[USER]
Generate 50 scenarios (10 per domain) testing IMPLICIT temporal horizons.
Domains: Personal Finance, Health Behavior, Career Decisions,
         Software Engineering, Geopolitical Reasoning.

### DEFINITION OF "IMPLICIT"
A horizon is implicit ONLY if it can be inferred from world-knowledge
about the named ACTIVITY or OUTCOME, with no time-denoting adjective,
adverb, or noun anywhere in the sentence. The reader should need
domain knowledge — not vocabulary — to tell short from long.

### DOMAINS (use neutral framing — do NOT bias toward a horizon)
1. Personal Finance: cash management vs. capital formation
2. Health Behavior: acute care vs. preventative care
3. Career Decisions: task execution vs. capability building
4. Software Engineering: incident response vs. system design
5. Geopolitical Reasoning: situational response vs. structural posture

### TASK
For each scenario produce {"domain", "ambiguous", "short", "long"}:
- "ambiguous": a neutral lead-in (a subject + verb stub) that names NO outcome.
- "short": completes the lead-in by naming an ACTIVITY whose typical
  duration is < 1 week, inferable from world knowledge alone.
- "long": completes the same lead-in by naming an ACTIVITY whose typical
  duration is > 1 year, inferable from world knowledge alone.

### HARD CONSTRAINTS (a generation that violates any of these is invalid)
1. BANNED VOCABULARY — none of these word stems, in any inflection,
   may appear in "short" or "long":
     time, temporal, horizon, term, duration, period, phase, stage,
     now, today, tonight, tomorrow, yesterday, soon, later, eventual,
     immediate, instant, present, current, ongoing, acute, urgent,
     near, far, future, past, hour, minute, day, week, month, year,
     decade, century, quarter, annual, weekly, monthly, daily,
     short, long, brief, lasting, permanent, perpetual, ephemeral,
     transient, temporary, momentary, generational, strategic,
     tactical, sustainable, structural, compounding, legacy,
     foundational, preventative, preventive.
   (This list is a SUPERSET of the explicit-condition keywords, so the
   implicit and explicit sets are guaranteed disjoint.)
2. LENGTH MATCH — "short" and "long" must be within ±2 tokens of each other.
3. MINIMAL DELTA — change only the activity/object noun phrase; subject and
   verb must be identical to "ambiguous".
4. NO DURATION NUMERALS — no digits, no number words ("two", "three", ...).
5. ACTIVITY, NOT ADJECTIVE — the contrast must rest on the ACTIVITY named
   (e.g. "patching a kernel CVE" vs "rearchitecting the persistence layer"),
   not on a duration-implying modifier.

### SELF-CHECK BEFORE EMITTING
For each pair, ask: "If I replaced the activity noun phrase in 'short'
with the one from 'long', would the sentence read as long-horizon?"
If yes, the contrast lives in the activity → keep.
If no (i.e. an adjective is carrying the contrast), discard and regenerate.

### EXAMPLES (note: no banned words; horizon falls out of the activity)
[
  {
    "domain": "Software Engineering",
    "ambiguous": "The platform team is",
    "short": "The platform team is rolling back a bad deployment.",
    "long":  "The platform team is migrating the monolith to microservices."
  },
  {
    "domain": "Health Behavior",
    "ambiguous": "The patient is",
    "short": "The patient is taking ibuprofen for a tension headache.",
    "long":  "The patient is enrolling in a cardiac rehabilitation program."
  },
  {
    "domain": "Personal Finance",
    "ambiguous": "The household is",
    "short": "The household is paying off this month's credit-card balance.",
    "long":  "The household is building a diversified retirement portfolio."
  },
  {
    "domain": "Career Decisions",
    "ambiguous": "The engineer is",
    "short": "The engineer is reviewing a teammate's pull request.",
    "long":  "The engineer is pursuing a PhD in distributed systems."
  },
  {
    "domain": "Geopolitical Reasoning",
    "ambiguous": "The ministry is",
    "short": "The ministry is recalling its ambassador over the incident.",
    "long":  "The ministry is negotiating a regional free-trade union."
  }
]

### OUTPUT
Start with [ and end with ]. No markdown fences. 50 objects.
\end{lstlisting}
\end{promptcontainer}

\begin{promptcontainer}[label={lst:prompt_explicit}]{Horizon-Anchored generation prompt, explicit anchoring}
\begin{lstlisting}[style=promptbox]
[SYSTEM]
You are a deterministic data generation engine. Output valid JSON only.

[USER]
Generate 50 scenarios (10 per domain) for these 5 domains:
Personal Finance, Health Behavior, Career Decisions,
Software Engineering, Geopolitical Reasoning.

### DOMAIN DEFINITIONS (horizon-neutral wording — do NOT bake in a horizon)
1. Personal Finance: cash management vs. capital formation
2. Health Behavior: acute care vs. preventative care
3. Career Decisions: task execution vs. capability building
4. Software Engineering: incident response vs. system design
5. Geopolitical Reasoning: situational response vs. structural posture
(Note: prior wording used "immediate", "tactical", "strategic" — these
words pre-cued horizon and are forbidden here.)

### HORIZON KEYWORDS (one semantic class only: explicit time anchors)
- Short anchors (allowed, exactly these forms):
    "right now", "today", "this morning", "this afternoon",
    "by the end of the day", "before tonight", "this week"
- Long anchors (allowed, exactly these forms):
    "next year", "in five years", "in a decade", "in fifty years",
    "by the end of the century", "over the next generation"

FORBIDDEN (drift into strategy / urgency / frequency / quality):
    current, crisis, urgent, urgently, temporary, weekly,
    sustainable, strategic, tactical, permanent, generational,
    later, eventually, quickly, soon, immediately, immediate,
    short-term, long-term, ongoing.
(These describe quality, urgency, or frequency rather than horizon
and would confound the probe.)

### TEMPLATE RULES (syntactically uniform)
- "ambiguous": a complete declarative sentence ending in "." with NO
  temporal anchor anywhere in it.
- "short" and "long": take the ambiguous sentence VERBATIM (no
  edits, no punctuation changes) and INSERT a keyword from the
  matching list at exactly one of two positions:
    POSITION A (sentence-final): "<ambiguous body> <keyword>."
    POSITION B (sentence-initial): "<keyword>, <ambiguous body>."
  Use POSITION A for half the scenarios and POSITION B for the other
  half, balanced across domains.
- The "short" and "long" sentences for one scenario MUST use the SAME
  position so that the only delta is the keyword itself.
- "short" and "long" must end in "." (not "?"), matching "ambiguous".
- Length match: choose short and long keywords whose token counts
  differ by at most 1 (e.g. pair "right now" with "next year", or
  "this week" with "in a decade").

### SELF-CHECK BEFORE EMITTING
For each {ambiguous, short, long} triple, verify:
  (a) deleting the keyword and the surrounding whitespace/comma from
      "short" yields "ambiguous" exactly, character-for-character;
  (b) same for "long";
  (c) no forbidden word appears in any field;
  (d) "ambiguous", "short", "long" all end in ".".
If any check fails, regenerate the triple.

### EXAMPLES
[
  {
    "domain": "Software Engineering",
    "ambiguous": "The platform team should patch the authentication service.",
    "short":     "The platform team should patch the authentication service right now.",
    "long":      "The platform team should patch the authentication service next year."
  },
  {
    "domain": "Personal Finance",
    "ambiguous": "The household should rebalance the brokerage portfolio.",
    "short":     "Today, the household should rebalance the brokerage portfolio.",
    "long":      "In a decade, the household should rebalance the brokerage portfolio."
  }
]

### OUTPUT
Return a single JSON list of 50 objects. Start with [, end with ].
No markdown fences. No commentary.
\end{lstlisting}
\end{promptcontainer}

\begin{promptcontainer}[label={lst:prompt_graded}]{Horizon-Graded generation prompt}
\begin{lstlisting}[style=promptbox]
You are an expert dataset generator for NLP interpretability research. I need you to generate 50 distinct base scenarios that involve waiting or anticipating a future event.

For each of the 50 scenarios, generate 10 variations. The variations must be IDENTICAL word-for-word, except for the explicit temporal phrase indicating when the event will happen.

The 10 temporal horizons must be exactly:

"tomorrow"
"next week"
"next month"
"in six months"
"next year"
"in five years"
"in a decade"
"in fifty years"
"in a century"
"in five hundred years"

### Output Format: 
JSON list of objects. 
Each object should have an "id", a "base_scenario", and a "variations" dictionary mapping the horizon string to the full sentence.

Example item: { "id": 1, "base_scenario": "The city council announced that the new hyper-loop transit system will be completed and open to the public ", "variations": { "1_day": "The city council announced that the new hyper-loop transit system will be completed and open to the public tomorrow.", "1_week": "The city council announced that the new hyper-loop transit system will be completed and open to the public next week.", ... "500_years": "The city council announced that the new hyper-loop transit system will be completed and open to the public in five hundred years." } }
Now generate the instances now as cleanly formatted JSON.
\end{lstlisting}
\end{promptcontainer}

\begin{promptcontainer}[label={lst:prompt_model_eval}]{Three-shot horizon classification prompt used for model filtering}
\begin{lstlisting}[style=promptbox]
For the given situation, classify the temporal horizon as "short" or "long".

Example 1
Situation: I want to make one cup of coffee.
Answer: "short"

Example 2
Situation: I want to become an engineer.
Answer: "long"

Example 3
Situation: {prompt}
Answer: "
\end{lstlisting}
\end{promptcontainer}

\newpage
\section{Representational Alignment Analysis}
\label{app:representation_analysis}

The main text characterizes the evolution of short- and long-horizon
representations through Centered Kernel Alignment alone. This appendix reports
the supporting analyses: the quality and specificity of the linear probes on
which the temporal direction is defined
(Appendix~\ref{app:probe_quality}), a battery of scale-dependent and
scale-invariant separability diagnostics measured at every layer
(Appendix~\ref{app:separability}), a modality analysis that explains why the
short/long separation never resolves into two visible peaks
(Appendix~\ref{app:modality}), and a sigmoid fit that quantifies how sharp the
transition actually is (Appendix~\ref{app:crossover_width}).

The synthesis across all of these diagnostics is a single claim: across all
seven models and both anchoring conditions, the short-to-long transition is a
\emph{smooth crossover}. Every
scale-invariant quantity that peaks does so within the first third-to-half of
the network, saturates below its theoretical bimodal ceiling, and does so
without any narrow one-to-two layer spike. The one robust condition difference
is a difference in \emph{timing}: explicit anchoring establishes short/long
separation almost immediately, while implicit anchoring accrues it gradually
over depth.

\subsection{Probe Quality and Selectivity Controls}
\label{app:probe_quality}

Every analysis in this paper that relies on a temporal direction---the
logarithmic encoding fits, the manifold geometry descriptors, and the causal
steering experiments---derives that direction from a linear probe trained at
each layer~\citep{probetraining}. The probes must therefore be shown to read genuine temporal
structure rather than an artifact of the classifier's capacity.

Table~\ref{tab:probe_accuracy} reports peak validation accuracy and the depth
at which it occurs. Probes reach $98.8$--$99.5\%$ accuracy in every model and
condition, so the short/long distinction is linearly decodable from the
residual stream essentially everywhere. The more informative quantity is
\emph{where} the peak sits. In all seven models the explicit peak occurs
earlier in the network than the implicit peak, and in four models it occurs
within the first fifth of the network: Gemma2-2B peaks at layer 1 of 26 and
Qwen3-14B at layer 3 of 40 under explicit anchoring, whereas their implicit
peaks fall at fractional depths $0.38$ and $0.53$ respectively. Explicit
anchoring is a surface cue that a probe can exploit immediately; implicit
anchoring requires the network to first infer the horizon from world knowledge
about the named activity, and only becomes maximally decodable once that
inference has been performed. This ordering is the probing counterpart of the
CKA asymmetry reported in the main text.

\begin{table}[htbp]
\centering
\caption{Peak probe validation accuracy and the fractional depth at which it is
attained, for each model and anchoring condition. In every model the explicit
peak precedes the implicit peak in depth.}
\label{tab:probe_accuracy}
\begin{tabular}{|l|c|c|c|c|}
\hline
\multirow{2}{*}{\textbf{Model}} & \multicolumn{2}{c|}{\textbf{Explicit}} & \multicolumn{2}{c|}{\textbf{Implicit}} \\ \cline{2-5}
 & \textbf{Peak acc.} & \textbf{Depth} & \textbf{Peak acc.} & \textbf{Depth} \\ \hline
Qwen3-1.7B   & 0.988 & 0.18 & 0.990 & 0.39 \\ \hline
Qwen3-4B     & 0.990 & 0.33 & 0.993 & 0.39 \\ \hline
Qwen3-8B     & 0.990 & 0.31 & 0.991 & 0.58 \\ \hline
Qwen3-14B    & 0.989 & 0.08 & 0.994 & 0.53 \\ \hline
Llama-3.1-8B & 0.992 & 0.28 & 0.995 & 0.31 \\ \hline
Gemma2-2B    & 0.989 & 0.04 & 0.991 & 0.38 \\ \hline
Pythia-12B   & 0.990 & 0.14 & 0.989 & 0.33 \\ \hline
\end{tabular}
\end{table}

\begin{figure}[htbp]
    \centering
    \begin{subfigure}[b]{0.9\linewidth}
        \centering
        \includegraphics[width=\linewidth]{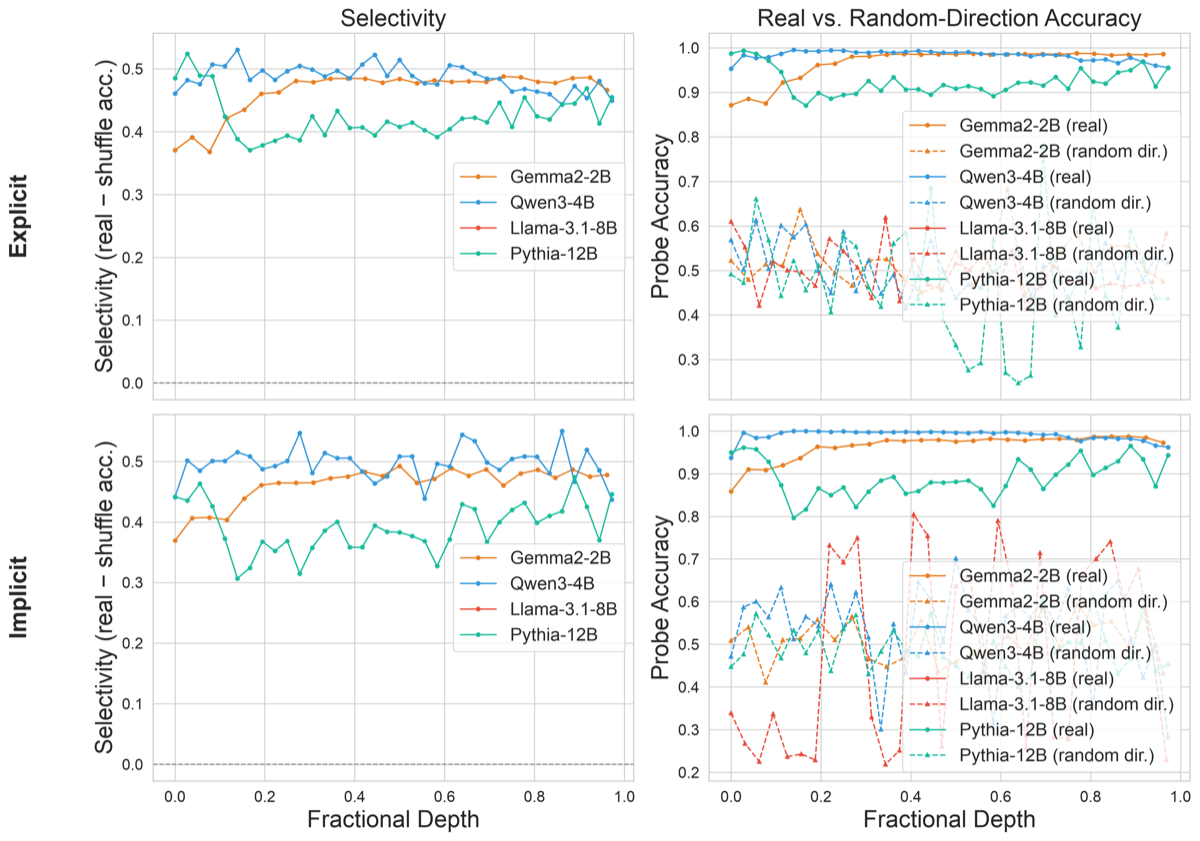}
        \caption{Across heterogeneous model families.}
        \label{fig:probe_selectivity_across}
    \end{subfigure}

    \vspace{0.3cm}

    \begin{subfigure}[b]{0.9\linewidth}
        \centering
        \includegraphics[width=\linewidth]{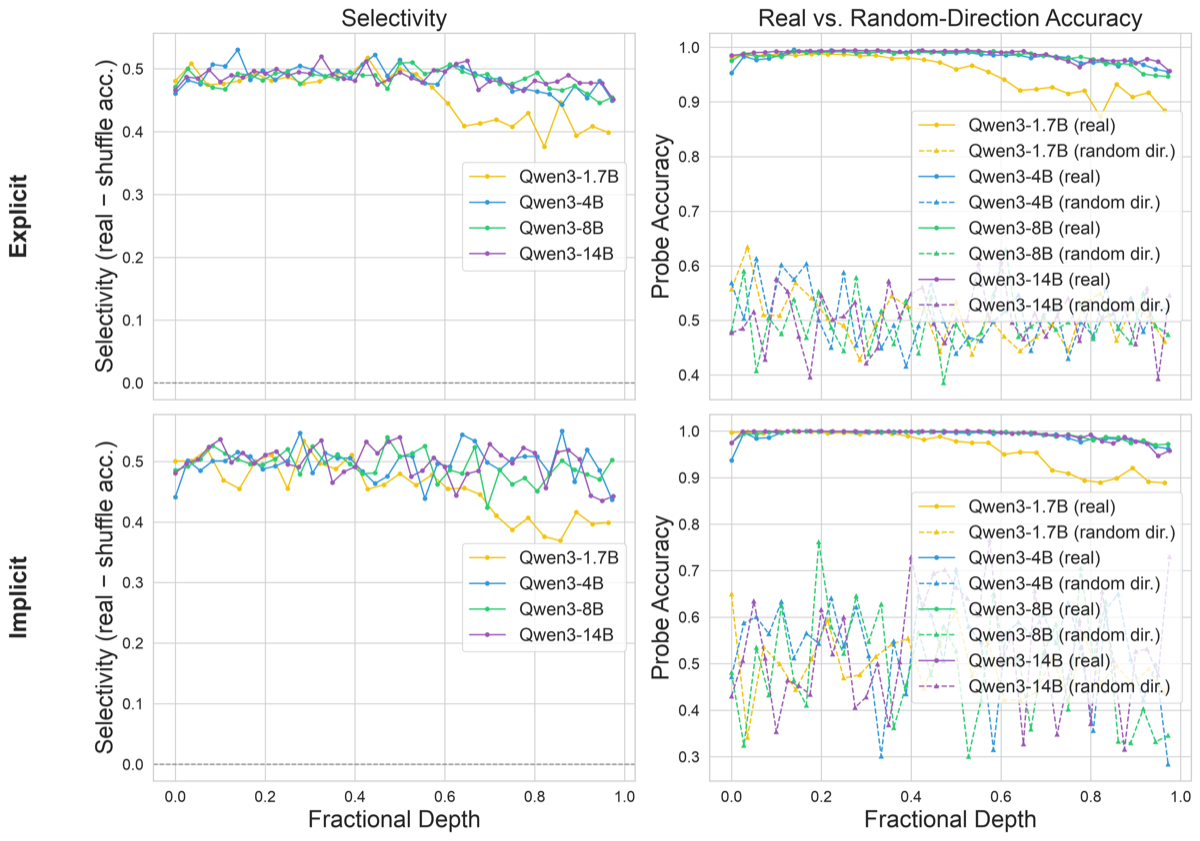}
        \caption{Within the Qwen3 model family.}
        \label{fig:probe_selectivity_within}
    \end{subfigure}

    \caption{Probe selectivity controls across normalized layer depth (top row of
    each panel: explicit anchoring; bottom row: implicit anchoring). Left:
    selectivity, defined as real minus shuffled-label accuracy, which stays far
    above zero at every depth. Right: real probe accuracy (solid) against
    random-direction accuracy (dashed), which fluctuates around chance
    throughout.}
    \label{fig:probe_selectivity}
\end{figure}

High accuracy alone does not establish that a probe has found a temporal
direction, so three controls were run at every layer, summarized in
Table~\ref{tab:probe_controls} and plotted against depth in
Figure~\ref{fig:probe_selectivity}. The \emph{shuffled-label} control retrains
the probe on randomly permuted labels and measures how much accuracy the
classifier can manufacture from capacity alone. The \emph{random-direction}
control projects activations onto a random unit vector instead of the learned
probe axis. The \emph{low-rank} control restricts the probe to a
low-dimensional subspace of the residual stream.

The shuffled-label and random-direction controls both sit at chance across
every model, condition, and layer, $0.494$--$0.505$ for shuffled labels and
$0.480$--$0.531$ for random directions, yielding a mean selectivity
(real minus shuffled accuracy) of $0.475$--$0.492$. Neither the probe's
capacity nor the ambient dimensionality of the residual stream can account for
any part of the observed accuracy. The low-rank control behaves differently and
is informative in its own right: restricting the probe to a low-rank subspace
still recovers $86$--$93\%$ accuracy, and does so more successfully under
explicit anchoring ($0.896$--$0.927$) than implicit ($0.858$--$0.877$). The
temporal horizon therefore occupies a genuinely low-dimensional, largely linear
subspace, and that subspace is slightly more compact when the horizon is
signaled by an explicit anchor phrase.

\begin{table}[htbp]
\centering
\caption{Probe selectivity controls, averaged over all layers. Shuffled-label
and random-direction accuracies sit at chance in every setting, while the
low-rank control retains most of the signal, indicating a low-dimensional
linear temporal subspace.}
\label{tab:probe_controls}
\begin{tabular}{|l|c|c|c|c|c|c|}
\hline
\multirow{2}{*}{\textbf{Model}} & \multicolumn{3}{c|}{\textbf{Explicit}} & \multicolumn{3}{c|}{\textbf{Implicit}} \\ \cline{2-7}
 & \textbf{Shuffle} & \textbf{Rand.\ dir.} & \textbf{Low-rank} & \textbf{Shuffle} & \textbf{Rand.\ dir.} & \textbf{Low-rank} \\ \hline
Qwen3-1.7B   & 0.504 & 0.499 & 0.907 & 0.498 & 0.503 & 0.867 \\ \hline
Qwen3-4B     & 0.496 & 0.499 & 0.919 & 0.498 & 0.515 & 0.877 \\ \hline
Qwen3-8B     & 0.497 & 0.522 & 0.915 & 0.500 & 0.522 & 0.865 \\ \hline
Qwen3-14B    & 0.502 & 0.508 & 0.927 & 0.496 & 0.505 & 0.866 \\ \hline
Llama-3.1-8B & 0.497 & 0.531 & 0.896 & 0.494 & 0.511 & 0.858 \\ \hline
Gemma2-2B    & 0.505 & 0.504 & 0.917 & 0.502 & 0.500 & 0.874 \\ \hline
Pythia-12B   & 0.499 & 0.480 & 0.925 & 0.502 & 0.492 & 0.866 \\ \hline
\end{tabular}
\end{table}

\subsection{Separability Diagnostics Across Depth}
\label{app:separability}

\begin{figure}[htbp]
    \centering
    \begin{subfigure}[b]{\linewidth}
        \centering
        \includegraphics[width=\linewidth]{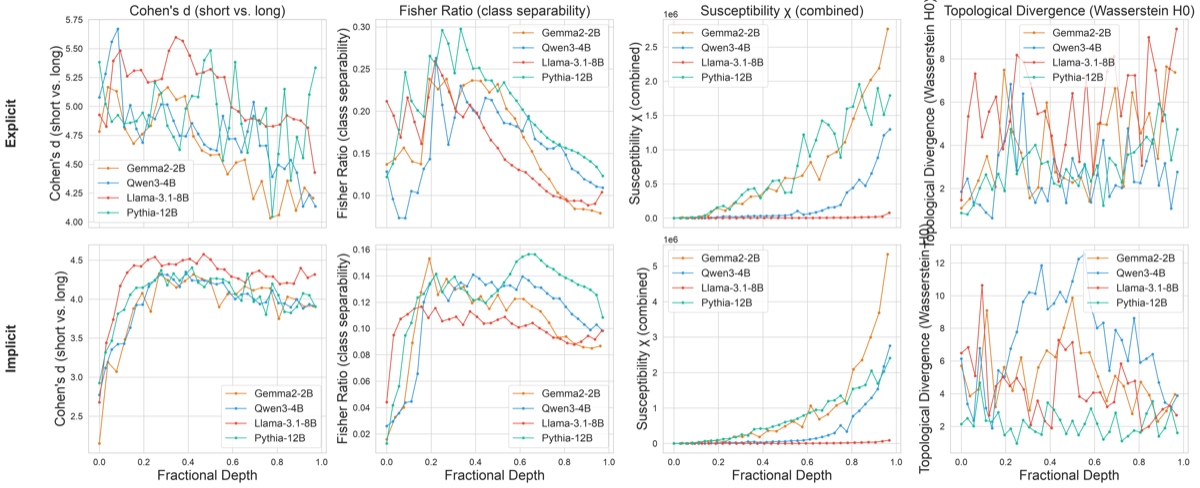}
        \caption{Across heterogeneous model families.}
        \label{fig:phase_diagnostics_across}
    \end{subfigure}

    \vspace{0.3cm}

    \begin{subfigure}[b]{\linewidth}
        \centering
        \includegraphics[width=\linewidth]{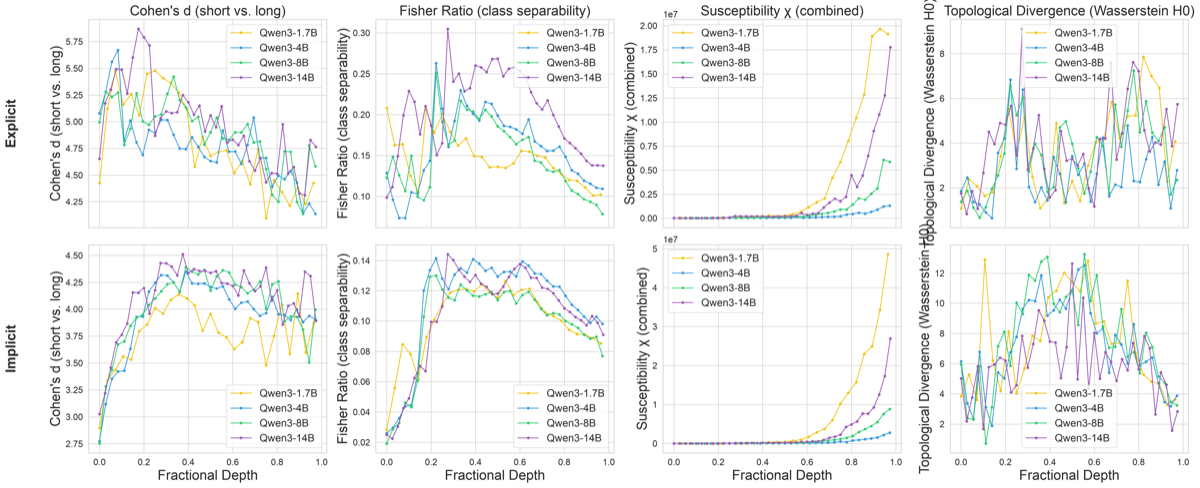}
        \caption{Within the Qwen3 model family.}
        \label{fig:phase_diagnostics_within}
    \end{subfigure}

    \caption{Extended separability diagnostics across normalized layer depth
    (row 0: explicit anchoring; row 1: implicit anchoring). From left to right:
    Cohen's $d$, Fisher discriminant ratio, susceptibility $\chi$, and
    Wasserstein-$H_0$ topological divergence. The monotonic late-layer growth of
    $\chi$ is an activation-norm artifact rather than a critical point.}
    \label{fig:phase_diagnostics}
\end{figure}

Figure~\ref{fig:phase_diagnostics} tracks four complementary measures of
short/long separability as a function of fractional depth: Cohen's $d$\citep{cohensd_metric} between
the two classes along the probe axis, the Fisher discriminant ratio \citep{fisher1936discriminant}
(between-class over within-class variance), the statistical-physics
susceptibility $\chi = N \cdot \mathrm{Var}(m)$ of the order parameter
$m = \langle h, \hat{w}\rangle$ \citep{magneticsusceptibility}, and the Wasserstein-$H_0$ topological
divergence between the short and long point clouds.

\paragraph{Fisher ratio and Cohen's $d$.}
Both quantities peak within the first third of the network
(fractional depth $\approx 0.19$--$0.33$) and decline thereafter, in every model
and both conditions. Linear separability is therefore constructed early-to-mid
network and then partially relinquished, not accumulated monotonically toward
the output. Explicit anchoring achieves roughly twice the peak Fisher ratio of
implicit anchoring ($\approx 0.24$--$0.30$ versus $\approx 0.12$--$0.16$),
echoing the probe-depth asymmetry of
Appendix~\ref{app:probe_quality}. Within the Qwen3 family, peak separability
grows monotonically with scale (explicit: $0.21$ at 1.7B rising to $0.30$ at
14B), so scale amplifies the magnitude of the separation. Across families the
shape is preserved without exception; Pythia-12B peaks latest and highest
(fractional depth $0.33$) and Gemma2-2B earliest ($0.19$).

\paragraph{Susceptibility is dominated by a norm artifact.}
The susceptibility panel of Figure~\ref{fig:phase_diagnostics} appears to show
a dramatic critical point at the output layers, with $\chi$ reaching magnitudes
of $10^6$ and above at fractional depth $0.83$--$0.97$. An out-of-the-box behavior
toward the end of the model can be attributed to the model allocating
its efforts toward output token computation rather than toward abstract
concept computation. It should not be confused with phase-transition-like
behavior~\citep{stanley1971phase}. Both $\chi$ and the
associated specific-heat proxy scale with raw activation variance, and the raw
activation norm grows by roughly two orders of magnitude across depth (see the
manifold geometry results). The scale-invariant Binder cumulant 
$U_4 = 1 - \langle m_c^4\rangle / 3\langle m_c^2\rangle^2$ \citep{bindercumulant} removes this
confound, and it tells the opposite story: it peaks mid-network ($\approx
0.53$--$0.57$ for explicit, $\approx 0.45$--$0.48$ for implicit), saturates
\emph{below} the bimodal ceiling of $2/3$, and then declines. By the standard
criterion, a broad $\chi$ peak together with $U_4$ rising smoothly and
saturating below $2/3$, every model exhibits a crossover rather than a phase
transition. The same caution applies to any unnormalized magnitude-based
quantity in this work.

\subsection{Modality of the Horizon Distribution}
\label{app:modality}

Two tests of modality were applied to the combined short-plus-long projection
onto the probe axis, and they disagree in a way that is itself the result.
Hartigan's dip test\citep{hartigandiptest} rejects unimodality at every layer, in every model, and in
both conditions, with the dip statistic roughly twice as large for explicit
anchoring ($\approx 0.08$--$0.10$) as for implicit ($\approx 0.05$--$0.06$).
Sarle's bimodality coefficient\citep{pfister2013bimodality}, by contrast, never exceeds $\approx 0.23$
anywhere, far below its bimodality threshold of $0.555$, and by that criterion
the distribution is unimodal everywhere.

The contradiction resolves as follows. The two horizon classes are strongly
separated in the mean (Cohen's $d$ is around $5$) but their within-class
spread is larger still, so the mixture has one dominant mode with a pronounced
shoulder rather than two resolved peaks. The dip test is sensitive enough to
detect the shoulder; the moment-based coefficient, which depends on skewness
and kurtosis, is not. That the separation is real but never resolves into two
visible modes is the single clearest piece of evidence that the short-to-long
transition is a crossover rather than a two-phase split, and it is consistent
across every model and condition examined.

Jensen-Shannon divergence\citep{jsd} between the short and long projection histograms
completes the picture. JSD is high and close to its $\log 2 \approx 0.693$
ceiling at essentially every depth ($\approx 0.62$--$0.67$), confirming that the
two marginals are strongly distinguishable throughout the network. Its
condition-dependent shape mirrors CKA: explicit anchoring is nearly flat with a
shallow mid-network peak, while implicit anchoring starts lower ($\approx
0.48$--$0.59$ at the input) and rises to the same plateau by mid-depth.
Llama-3.1-8B attains the highest peak ($0.667$ explicit, $0.669$ implicit) and
Gemma2-2B the lowest and earliest, but the near-ceiling plateau itself is
architecture-independent.

\subsection{Crossover Width}
\label{app:crossover_width}

\begin{figure}[htbp]
    \centering
    \begin{subfigure}[b]{0.85\linewidth}
        \centering
        \includegraphics[width=\linewidth]{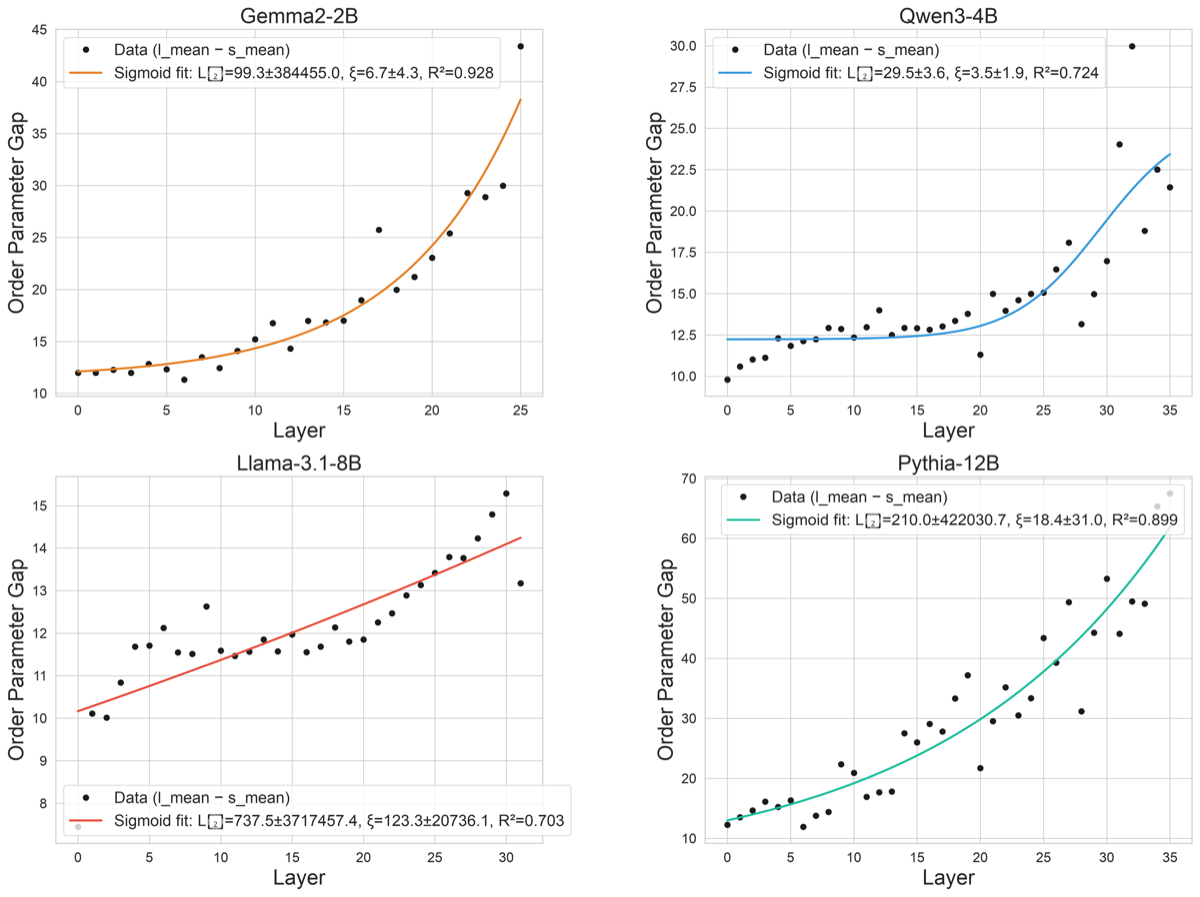}
        \caption{Across heterogeneous model families.}
        \label{fig:crossover_width_across}
    \end{subfigure}

    \vspace{0.3cm}

    \begin{subfigure}[b]{0.85\linewidth}
        \centering
        \includegraphics[width=\linewidth]{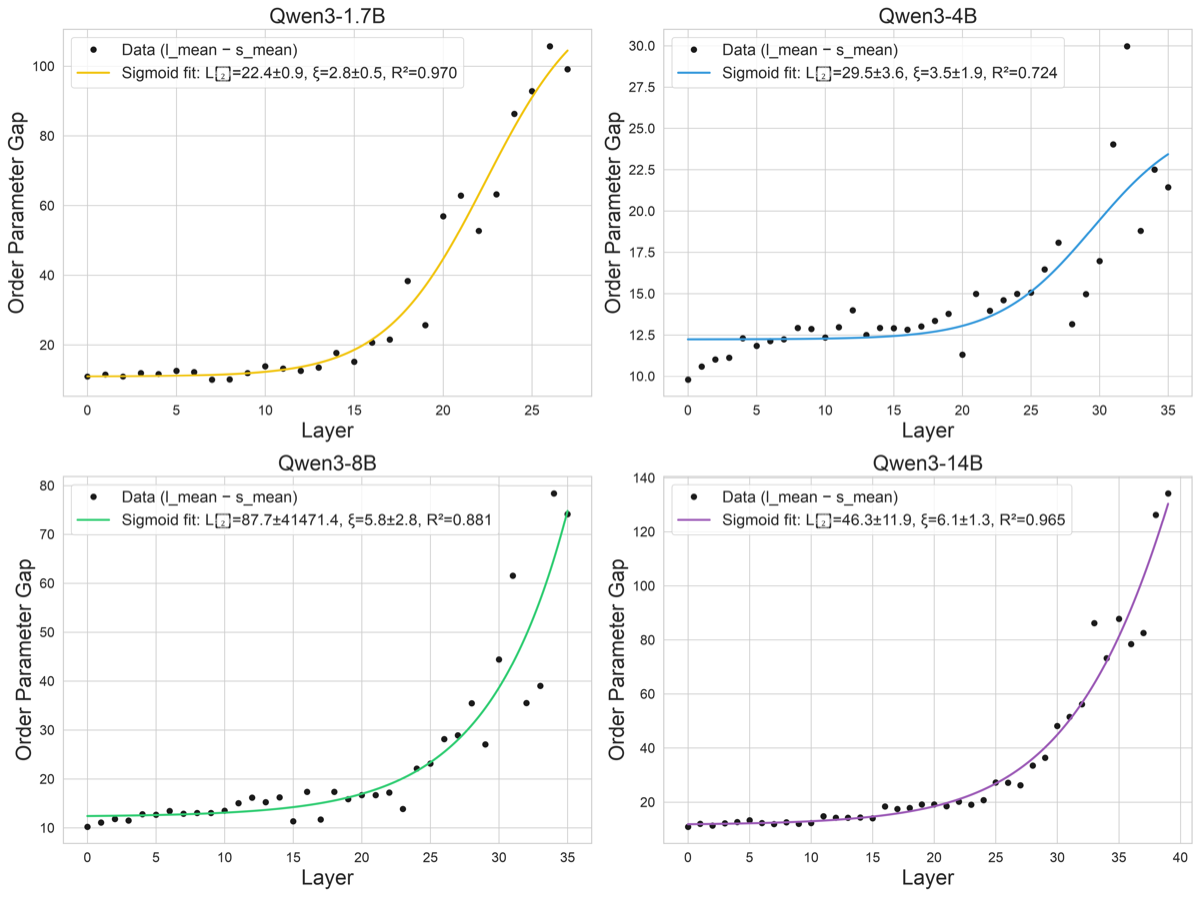}
        \caption{Within the Qwen3 model family.}
        \label{fig:crossover_width_within}
    \end{subfigure}

    \caption{Logistic fits to the explicit order-parameter gap against layer
    index. Points are the measured gap $\ell_{\text{mean}} - s_{\text{mean}}$;
    curves are the fitted sigmoid with parameters annotated. In most models the
    gap is still on the rising arm of the sigmoid at the final layer.}
    \label{fig:crossover_width}
\end{figure}

To quantify how sharp the transition is, a logistic function
\begin{equation}
    g(\ell) = \frac{A}{1 + \exp\!\left(-(\ell - L_0)/\xi\right)} + B
\end{equation}
was fitted to the explicit order-parameter gap
($\ell_{\text{mean}} - s_{\text{mean}}$) as a function of layer index $\ell$.
Here $L_0$ is the crossover midpoint, the layer at which the gap is half-open,
and $\xi$ is the crossover width in layers, with small $\xi$ denoting an abrupt
switch and large $\xi$ a gradual ramp. Fitted parameters are given in
Table~\ref{tab:crossover_width} and the fits themselves are shown in
Figure~\ref{fig:crossover_width}.

\begin{table}[htbp]
\centering
\caption{Logistic fit to the explicit order-parameter gap. $L_0$ is the
crossover midpoint and $\xi$ the crossover width, both in layers; $L_0/n$ and
$\xi/n$ express them as fractions of model depth. Values of $L_0/n > 1$
indicate that the fitted midpoint lies beyond the final layer, so the gap is
still rising at the output.}
\label{tab:crossover_width}
\begin{tabular}{|l|r|r|r|r|r|r|}
\hline
\textbf{Model} & \textbf{$L_0$} & \textbf{$L_0/n$} & \textbf{$\xi$} & \textbf{$\xi/n$} & \textbf{$\xi$ rel.\ err.} & \textbf{$R^2$} \\ \hline
Qwen3-1.7B   &  22.4 & 0.80  &   2.83 & 0.10 & 0.18 & 0.970 \\ \hline
Qwen3-4B     &  29.5 & 0.82  &   3.47 & 0.10 & 0.55 & 0.724 \\ \hline
Qwen3-8B     &  87.7 & 2.43  &   5.79 & 0.16 & 0.48 & 0.881 \\ \hline
Qwen3-14B    &  46.3 & 1.16  &   6.14 & 0.15 & 0.22 & 0.965 \\ \hline
Llama-3.1-8B & 737.5 & 23.05 & 123.31 & 3.85 & 168  & 0.703 \\ \hline
Gemma2-2B    &  99.3 & 3.82  &   6.72 & 0.26 & 0.64 & 0.928 \\ \hline
Pythia-12B   & 210.0 & 5.83  &  18.42 & 0.51 & 1.68 & 0.899 \\ \hline
\end{tabular}
\end{table}

The dominant result is that $L_0$ lands at or beyond the final layer for five
of the seven models, with fitted midpoint uncertainties of $10^4$ to $10^6$
layers --- that is, the midpoint is an unconstrained extrapolation rather than a
measured quantity. Mechanistically this means the short/long gap is still
rising at the output and never saturates within the observed depth: the
logistic is being fitted to its lower, still-climbing arm. This is the same
conclusion reached independently by the modality analysis and by CKA, now
expressed quantitatively. Only the two smallest Qwen models place $L_0$ inside
the network, at fractional depth $\approx 0.80$--$0.82$, and only there should
the midpoint be regarded as trustworthy. The width $\xi$ is the more robustly
estimated parameter and carries the interpretation below.

Within the Qwen3 family the crossover \emph{broadens} monotonically with scale:
$\xi$ rises from $2.83$ layers at 1.7B through $3.47$ and $5.79$ to $6.14$
layers at 14B. Larger models therefore build the short/long separation over a
wider band of depth rather than a sharper one, which combines with the Fisher
result to give a consistent picture of scale amplifying the magnitude of
separation while spreading it more gradually across layers. Fit quality is best
at the extremes of the sweep ($R^2 = 0.970$ for Qwen3-1.7B and $0.965$ for
Qwen3-14B) and weakest for Qwen3-4B ($R^2 = 0.724$), whose order-parameter gap
is the noisiest in the family.

Across families, crossover sharpness does not track model size. Qwen3-4B has
the sharpest reliable crossover at $\xi/n = 0.10$, Gemma2-2B is moderately
broader at $0.26$, and Pythia-12B is the broadest well-defined case at
$\xi/n = 0.51$, ramping its short/long gap across essentially half its depth.
Llama-3.1-8B should not be read as an extremely wide crossover: its fit is
degenerate, with $\xi = 123$ layers spanning nearly four times its own depth, a
relative error of $168\times$, and the worst $R^2$ in the sweep. The logistic
has collapsed into a near-linear ramp over the observed range, meaning
Llama-3.1-8B has no localizable crossover at all --- its order-parameter gap
rises roughly linearly with depth and never bends. This is the crossover-width
signature of the same behavior Llama-3.1-8B exhibits elsewhere in this work,
where it consistently shows the flattest CKA profile and the least
representational reorganization across depth.

\subsection{Causal and Mechanistic Analyses}
\label{app:mechanistic}

Every analysis presented so far is descriptive: it establishes that the
temporal horizon is linearly encoded in the residual stream and characterizes
where along depth that encoding is built. It does not establish that the
encoding is causally \emph{used}. This subsection reports three families of
intervention experiments that address the causal question --- activation
patching along the temporal axis
(Appendix~\ref{app:activation_patching}), circuit-level node attribution
(Appendix~\ref{app:acdc}), and per-head metric convergence
(Appendix~\ref{app:head_convergence}).

The headline result is a dissociation between encoding and use. The temporal
axis is cleanly causal precisely where the representational diagnostics say
separability is being constructed, and dilutes toward the output; the circuit
that carries the signal is an MLP backbone rather than a set of specialist
heads; and at the head level the attention heads that best \emph{encode} the
horizon are almost disjoint from those that causally \emph{use} it.

\subsubsection{Activation Patching Along the Temporal Axis}
\label{app:activation_patching}

Two interventions were run at every layer. \emph{Full patching} replaces the
entire residual stream at layer $L$ in a short-horizon run with the
corresponding activations from a long-horizon run, and measures the resulting
output shift\citep{activationpatching}. \emph{Directional patching} is the controlled counterpart: it
defines a single temporal axis per layer as the normalized class-mean
difference,
$\hat{d}_{\text{temp}}(L) = \widehat{\mathrm{mean}_{\text{long}}(L) - \mathrm{mean}_{\text{short}}(L)}$,
and transplants only the scalar projection onto that axis,
\begin{equation}
    h_{\text{patched}} = h_{\text{short}} + \left[\langle h_{\text{long}}, \hat{d}\rangle - \langle h_{\text{short}}, \hat{d}\rangle\right] \hat{d},
\end{equation}
leaving every orthogonal component untouched. The quantity of interest is the
ratio of directional to full effect, which measures the fraction of the total
causal effect that a single linear temporal direction reproduces. A ratio near
one means the horizon is carried by one clean linear direction at that depth; a
ratio near zero means the full-patch effect is dominated by content orthogonal
to the temporal axis.

\begin{figure}[htbp]
    \centering
    \begin{subfigure}[b]{0.9\linewidth}
        \centering
        \includegraphics[width=\linewidth]{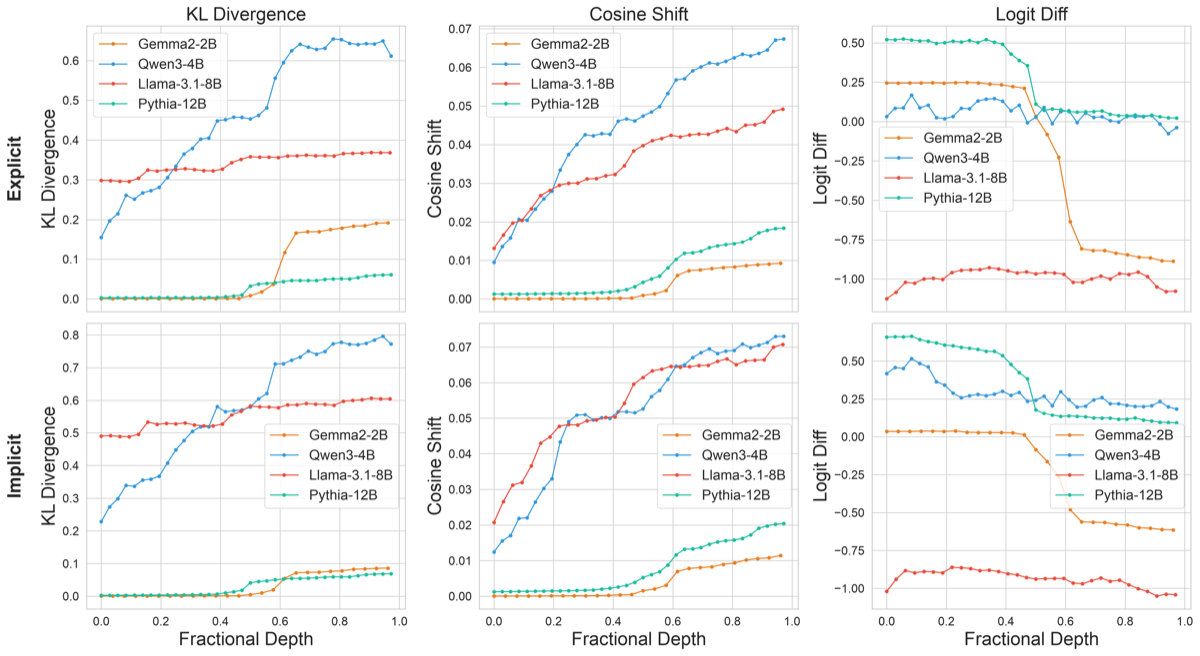}
        \caption{Across heterogeneous model families.}
        \label{fig:activation_patching_across}
    \end{subfigure}

    \vspace{0.3cm}

    \begin{subfigure}[b]{0.9\linewidth}
        \centering
        \includegraphics[width=\linewidth]{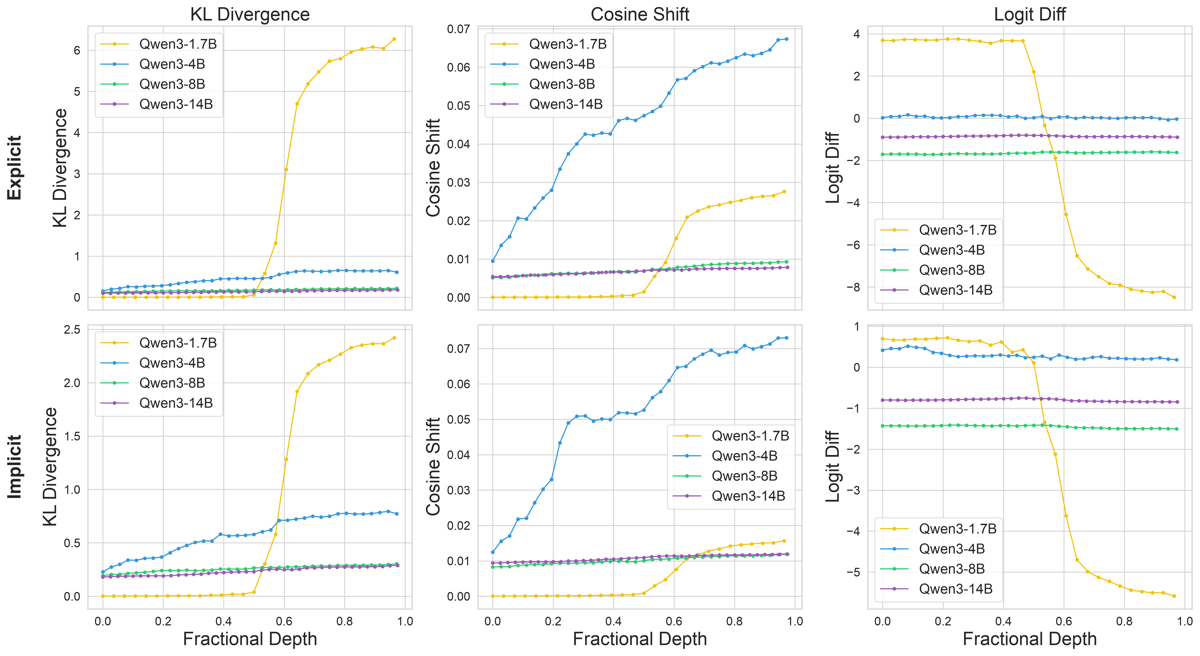}
        \caption{Within the Qwen3 model family.}
        \label{fig:activation_patching_within}
    \end{subfigure}

    \caption{Full-patch causal effect against normalized layer depth. The effect
    grows monotonically with depth in every model and condition, reflecting the
    reduced opportunity for downstream layers to wash out a late-layer
    intervention.}
    \label{fig:activation_patching}
\end{figure}

Full-patch effect size rises with depth universally: the Spearman correlation
between full-patch KL divergence and depth is at least $+0.69$ for every model
and condition, and at least $+0.95$ for most, as shown in
Figure~\ref{fig:activation_patching}. This is expected --- a patch applied
closer to the output has fewer remaining layers in which to be re-processed or
washed out --- and is therefore not itself evidence about temporal structure.

\begin{figure}[htbp]
    \centering
    \begin{subfigure}[b]{0.9\linewidth}
        \centering
        \includegraphics[width=\linewidth]{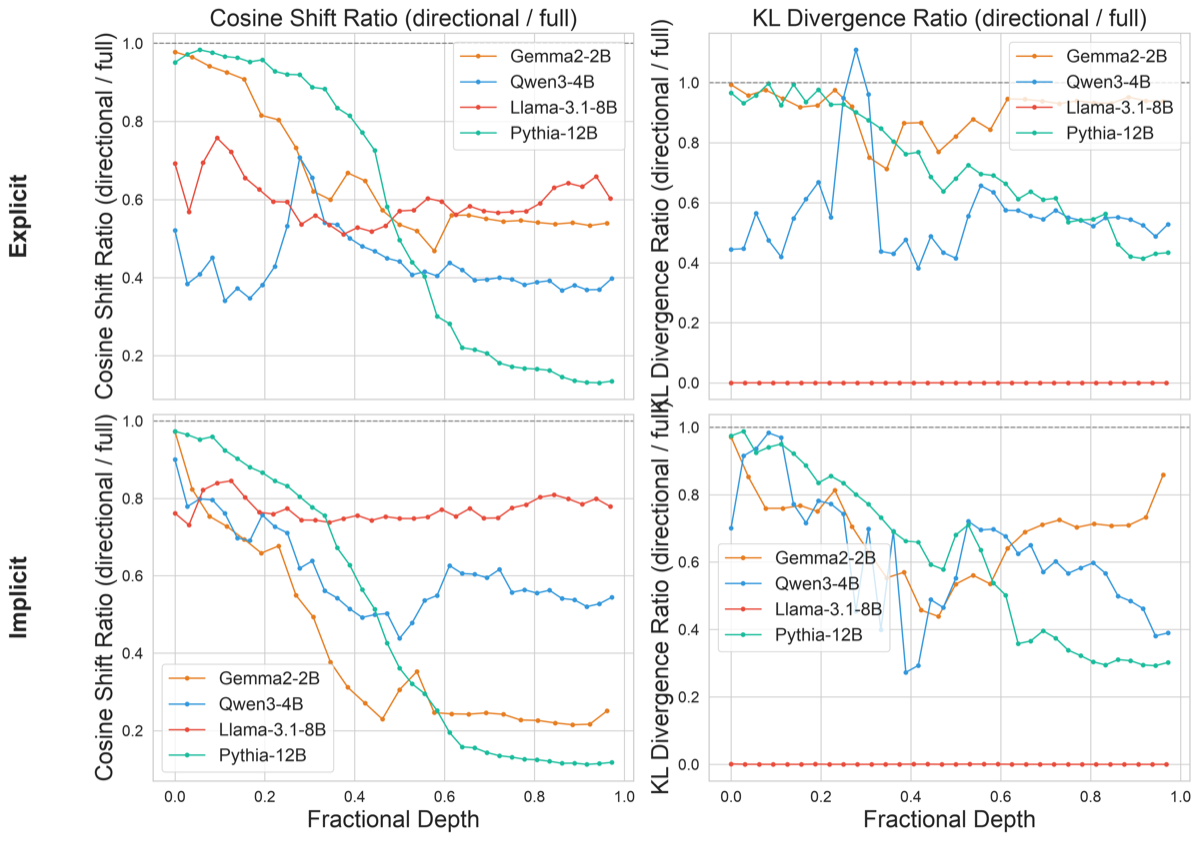}
        \caption{Across heterogeneous model families.}
        \label{fig:directional_patching_across}
    \end{subfigure}

    \vspace{0.3cm}

    \begin{subfigure}[b]{0.9\linewidth}
        \centering
        \includegraphics[width=\linewidth]{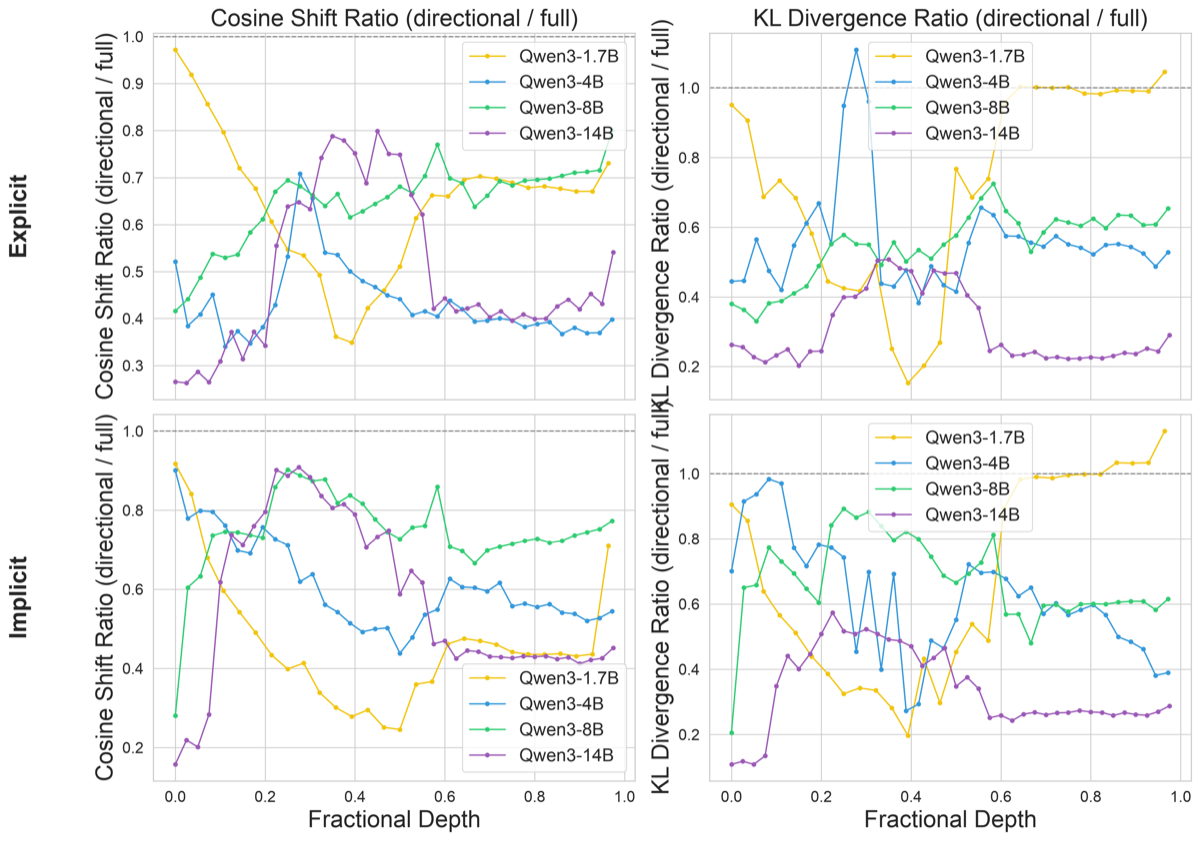}
        \caption{Within the Qwen3 model family.}
        \label{fig:directional_patching_within}
    \end{subfigure}

    \caption{Directional versus full activation patching across normalized layer
    depth. The directional intervention transplants only the scalar projection
    onto the temporal axis; the ratio of the two curves measures how much of the
    causal effect a single linear direction reproduces.}
    \label{fig:directional_patching}
\end{figure}

The informative quantity is the directional/full ratio, reported in
Table~\ref{tab:patching_ratio} and plotted in
Figure~\ref{fig:directional_patching}. In the models where the trend is
strongest, the ratio falls sharply with depth: the temporal effect is highly
linear early-to-mid network, where a single direction reproduces a large share
of the full-patch effect, and that share shrinks toward the output as the full
patch increasingly drags along non-temporal lexical, positional, and length
content that the directional patch excludes by construction. This is the
causal echo of the separability results in
Appendix~\ref{app:separability}: the temporal axis is cleanest exactly where
Fisher ratio and Cohen's $d$ say the separation is being built.

The trend is, however, \emph{not} universal, and we report it as such.
Pythia-12B and Gemma2-2B show it decisively in both conditions
($\rho \le -0.86$), Qwen3-4B moderately, and Qwen3-8B trends in the opposite
direction on the cosine ratio. Within the Qwen family the sign of $\rho$ differs
between adjacent model sizes, so the effect is not a simple function of scale.
What is consistent across all four Qwen sizes is the underlying mechanism:
directional KL keeps rising with depth in absolute terms in every case, so the
temporal axis never stops being causal --- it is diluted rather than destroyed.

\begin{table}[htbp]
\centering
\caption{Spearman correlation between the directional/full cosine ratio and
layer depth. Negative values indicate that a single linear temporal direction
accounts for a shrinking share of the total causal effect toward the output.
The effect is strong for Pythia-12B and Gemma2-2B but is not universal.}
\label{tab:patching_ratio}
\begin{tabular}{|l|r|r|}
\hline
\textbf{Model} & \textbf{$\rho$ (explicit)} & \textbf{$\rho$ (implicit)} \\ \hline
Pythia-12B   & $-0.99$ & $-1.00$ \\ \hline
Gemma2-2B    & $-0.86$ & $-0.92$ \\ \hline
Qwen3-4B     & $-0.35$ & $-0.62$ \\ \hline
Qwen3-1.7B   & $-0.07$ & $-0.16$ \\ \hline
Llama-3.1-8B & $-0.08$ & $+0.19$ \\ \hline
Qwen3-14B    & $+0.22$ & $-0.35$ \\ \hline
Qwen3-8B     & $+0.83$ & $-0.10$ \\ \hline
\end{tabular}
\end{table}

\paragraph{The Llama-3.1-8B anomaly.}
Llama-3.1-8B requires a caveat that should be stated before any of its
directional-patching numbers are used. Its directional KL divergence is
essentially zero at every layer, on the order of $10^{-9}$ to $10^{-8}$,
against $10^{-3}$ to $10^{-1}$ for every other model, giving a
directional/full KL ratio of approximately zero at all depths --- while its
full-patch effect is of ordinary magnitude. The class-mean-difference axis
therefore carries almost no causal effect in Llama-3.1-8B alone. Two
explanations remain live and we cannot presently distinguish them: either
Llama-3.1-8B genuinely encodes horizon along a direction that diverges sharply
from the class-mean difference, which would be consistent with its flat CKA
profile and its degenerate crossover fit, or the Llama directional-patching run
suffers a scale or normalization defect. We flag this as requiring a targeted
re-run and do not lean on Llama-3.1-8B in the ratio narrative above.

\subsubsection{Circuit-Level Attribution}
\label{app:acdc}

Using ACDC\citep{wang2022interpretabilitywildcircuitindirect}, each attention head and each layer MLP was scored by the drop in a normalized
fidelity metric when that node's activation is patched from the long-horizon
run into the short-horizon run, and a node is flagged as in-circuit when its
metric drop is at least $0.01$. It is important to be precise about what this
does and does not constitute: membership is determined purely by this
magnitude threshold, so what is reported here is \emph{single-node attribution
thresholding, not an iterative edge-pruning circuit search}. Circuit size is
consequently sensitive to how many nodes sit just above the threshold.

Table~\ref{tab:acdc_composition} gives the resulting composition. The dominant
structural fact is MLP saturation: the threshold admits the large majority of
MLPs in the Qwen, Llama, and Pythia models --- between $72\%$ and $100\%$ ---
but only a small minority of attention heads, between $2.9\%$ and $9.8\%$. MLP
membership is therefore close to non-selective, and it is the attention heads
that discriminate. The natural reading is that the MLP stack carries the
temporal signal while heads decorate it. Within the Qwen family the circuit
sparsifies monotonically with scale, from $14.1\%$ of all nodes at 1.7B to
$4.6\%$ at 14B, with the attention-head recruitment falling from $9.8\%$ to
$2.9\%$; larger models localize the temporal computation into a smaller
relative footprint.

\begin{table}[htbp]
\centering
\caption{Composition of the thresholded temporal circuit under explicit
anchoring. ``Recruited'' columns give the fraction of all MLPs and all attention
heads admitted at the $0.01$ threshold. Gemma2-2B is attention-dominated and
Pythia-12B is heavily bloated relative to the rest.}
\label{tab:acdc_composition}
\begin{tabular}{|l|r|r|r|r|}
\hline
\textbf{Model} & \textbf{Circuit \% of nodes} & \textbf{MLPs recr.} & \textbf{Heads recr.} & \textbf{MLP share} \\ \hline
Qwen3-1.7B   & 14.1\% &  82\% &  9.8\% & 34\% \\ \hline
Qwen3-4B     &  9.1\% &  89\% &  6.6\% & 30\% \\ \hline
Qwen3-8B     &  7.1\% &  86\% &  4.6\% & 37\% \\ \hline
Qwen3-14B    &  4.6\% &  72\% &  2.9\% & 38\% \\ \hline
Llama-3.1-8B &  7.8\% &  78\% &  5.6\% & 30\% \\ \hline
Gemma2-2B    & 17.1\% &  19\% & 16.8\% & 12\% \\ \hline
Pythia-12B   & 28.6\% & 100\% & 26.8\% &  9\% \\ \hline
\end{tabular}
\end{table}

Two architectures break the pattern. Gemma2-2B is the compositional outlier:
only about a fifth of its MLPs enter the circuit against about a sixth of its
attention heads, giving an MLP share of just $12\%$, so its temporal circuit is
attention-dominated rather than MLP-saturated. Pythia-12B is bloated, with
$28.6\%$ of all nodes admitted including every MLP and more than a quarter of
all attention heads, four to six times the head recruitment of any other model.
Llama-3.1-8B, despite being an outlier on nearly every other metric in this
work, is unremarkable here.

Table~\ref{tab:acdc_depth} reports where the circuits sit and how stable they
are between conditions. Circuits are concentrated mid-to-late, at mean
fractional depth $0.48$ to $0.65$, and within the Qwen family drift later with
scale. Explicit and implicit circuits overlap heavily in six of seven models,
with Jaccard similarity between $0.71$ and $0.86$, and the shared core is the
MLP backbone: the temporal MLP circuit is essentially condition-invariant, and
the explicit/implicit differences live almost entirely in the attention heads.
Pythia-12B is again the exception, at Jaccard $0.38$, with an explicit circuit
roughly twice the size of its implicit one.

\begin{table}[htbp]
\centering
\caption{Circuit depth distribution and explicit/implicit stability. Jaccard
similarity is computed between the explicit and implicit circuits. Pythia-12B
is the only condition-unstable model.}
\label{tab:acdc_depth}
\begin{tabular}{|l|c|c|c|}
\hline
\textbf{Model} & \textbf{Mean frac.\ depth} & \textbf{Late-third share} & \textbf{Jaccard (exp., imp.)} \\ \hline
Qwen3-1.7B   & 0.58 & 37\% & 0.82 \\ \hline
Qwen3-4B     & 0.65 & 50\% & 0.82 \\ \hline
Qwen3-8B     & 0.65 & 57\% & 0.77 \\ \hline
Qwen3-14B    & 0.65 & 55\% & 0.86 \\ \hline
Llama-3.1-8B & 0.52 & 23\% & 0.75 \\ \hline
Gemma2-2B    & 0.65 & 40\% & 0.71 \\ \hline
Pythia-12B   & 0.48 / 0.65 & 29\% / 51\% & 0.38 \\ \hline
\end{tabular}
\end{table}

\begin{figure}[htbp]
    \centering
    \begin{subfigure}[b]{0.9\linewidth}
        \centering
        \includegraphics[width=\linewidth]{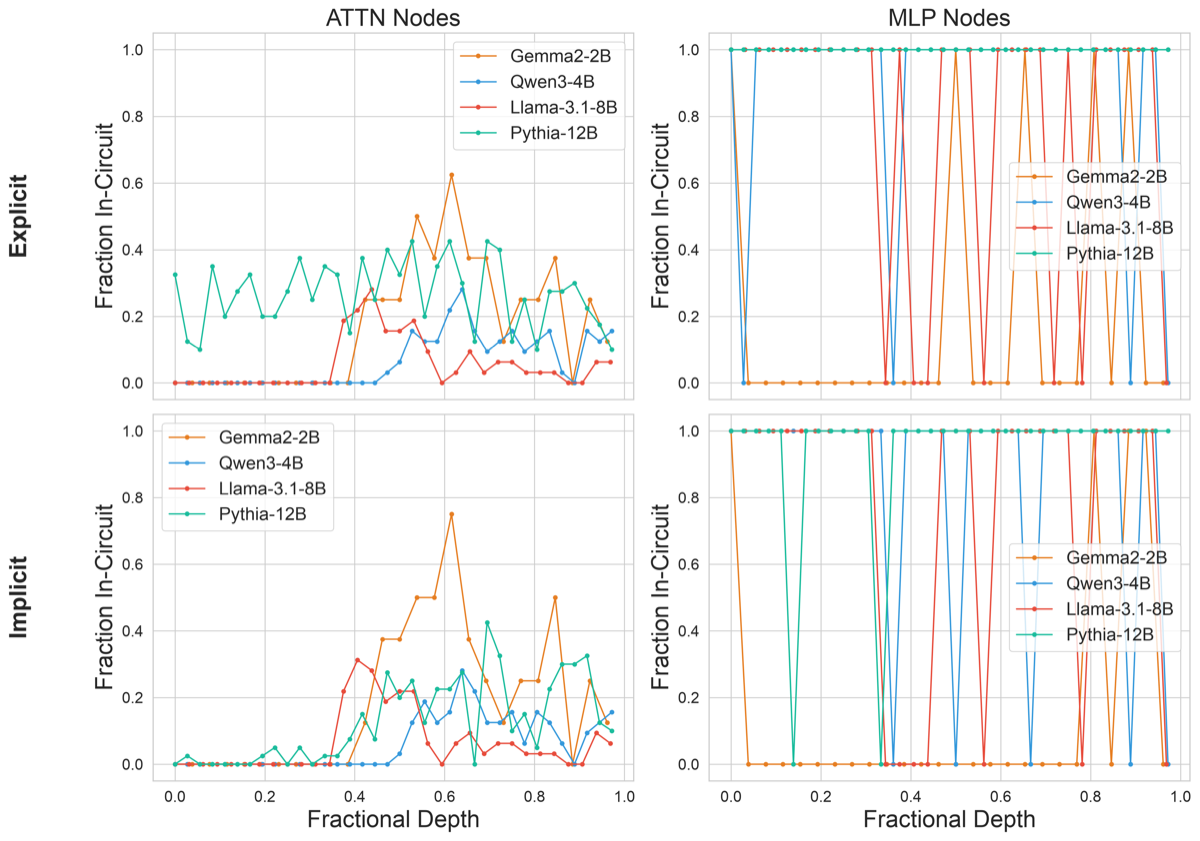}
        \caption{Across heterogeneous model families.}
        \label{fig:acdc_composition_across}
    \end{subfigure}

    \vspace{0.3cm}

    \begin{subfigure}[b]{0.9\linewidth}
        \centering
        \includegraphics[width=\linewidth]{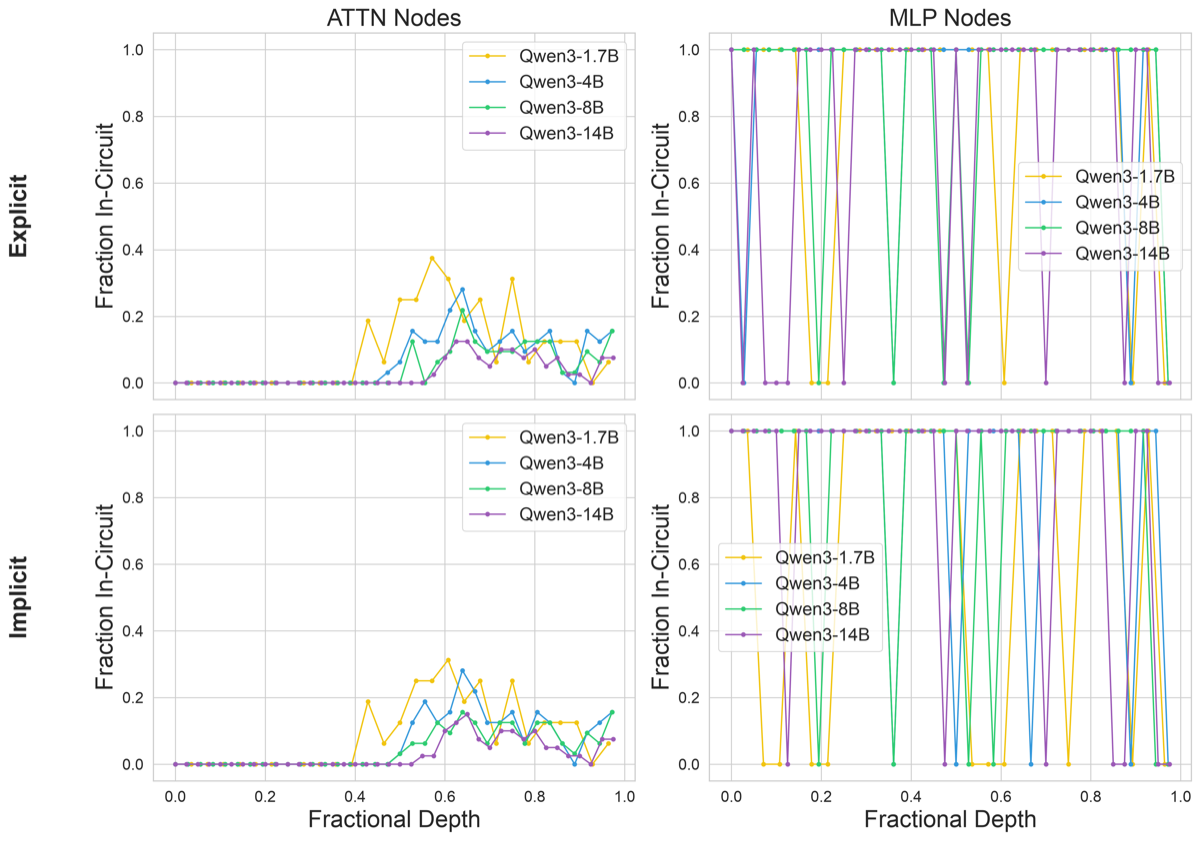}
        \caption{Within the Qwen3 model family.}
        \label{fig:acdc_composition_within}
    \end{subfigure}

    \caption{Composition and depth distribution of the thresholded temporal
    circuit. The circuit is MLP-saturated in all models except Gemma2-2B, and
    sparsifies with scale within the Qwen3 family.}
    \label{fig:acdc_composition}
\end{figure}

\paragraph{Caveats.}
Three artifacts temper any strong reading of these circuits and should be
reported alongside them. First, as noted above, the circuit is
threshold-defined rather than search-defined, so circuit size across models
should not be over-interpreted. Second, the embedding MLP is the single largest
node in most models, carrying $10$--$17\%$ of the total positive drop in every
Qwen model; this is the standard input-node boundary effect, since patching it
corrupts the input representation wholesale, and it inflates the apparent
importance of layer $0$. Third, per-node drop signs are unstable across conditions and should
not be used as a cross-condition quantity.

\begin{figure}[htbp]
    \centering
    \begin{subfigure}[b]{0.9\linewidth}
        \centering
        \includegraphics[width=\linewidth]{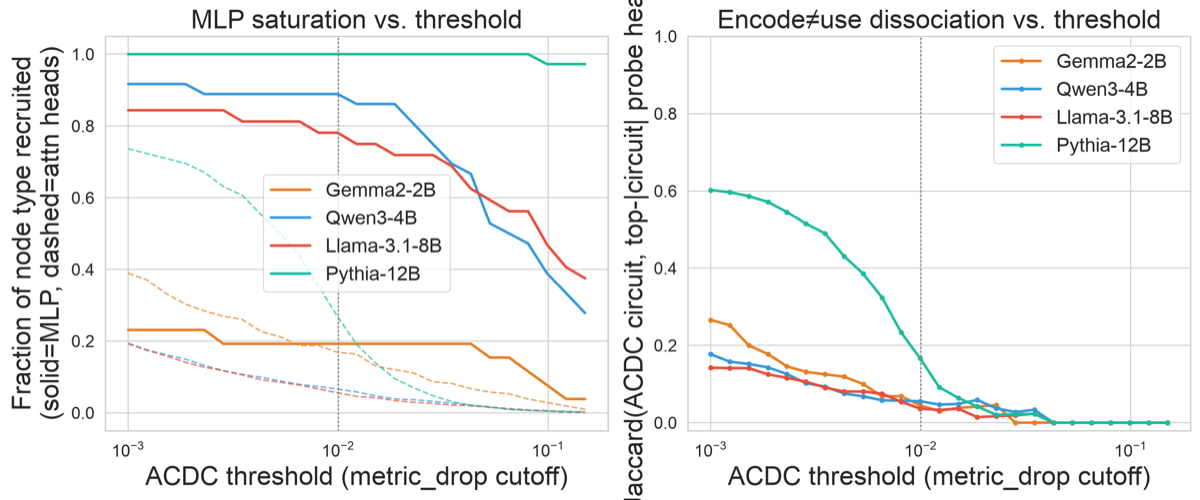}
        \caption{Across heterogeneous model families.}
        \label{fig:acdc_threshold_across}
    \end{subfigure}

    \vspace{0.3cm}

    \begin{subfigure}[b]{0.9\linewidth}
        \centering
        \includegraphics[width=\linewidth]{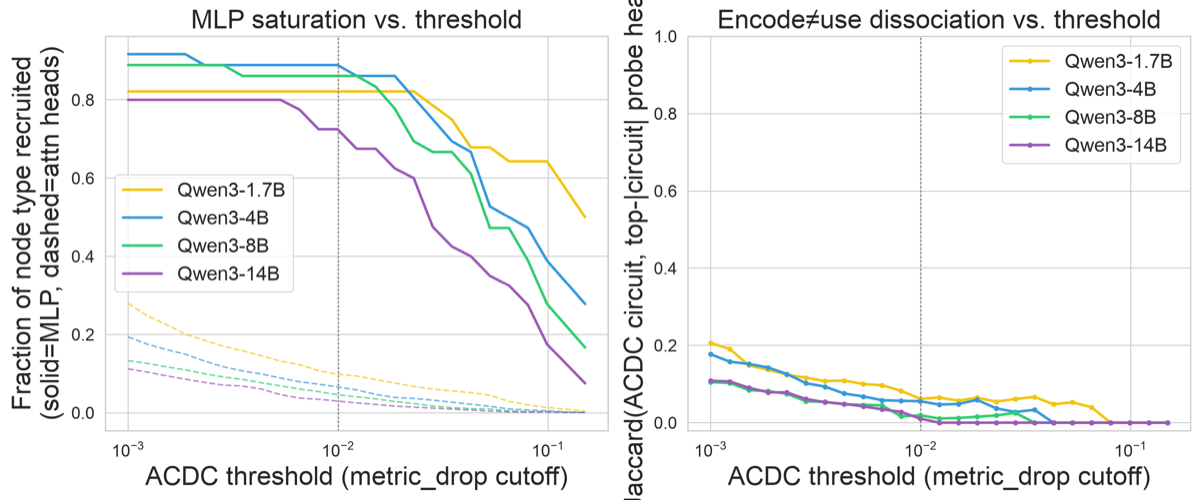}
        \caption{Within the Qwen3 model family.}
        \label{fig:acdc_threshold_within}
    \end{subfigure}

    \caption{Threshold-sweep robustness check over metric-drop thresholds
    $0.001$--$0.15$. Solid curves give the fraction of MLPs recruited and dashed
    curves the fraction of attention heads; the separation between them
    confirms that MLP saturation is not an artifact of the chosen threshold.}
    \label{fig:acdc_threshold}
\end{figure}

To check that the conclusions do not depend on the fixed threshold, it was
swept logarithmically over $0.001$--$0.15$ in 25 steps for all seven models
(Figure~\ref{fig:acdc_threshold}). MLP saturation is robust across the entire
range for every model except Gemma2-2B, whose attention-dominated exception is
likewise not a threshold artifact. The encode-versus-use dissociation developed
below is robust at every threshold tested for six of seven models. Pythia-12B is
the exception: at thresholds below $0.005$ its circuit and its top
probe-accuracy heads substantially agree, and the dissociation only appears once
the threshold exceeds roughly $0.02$. The dissociation should therefore be read
as a finding that holds generally for Qwen, Gemma, and Llama, and that Pythia
exhibits only once a stricter threshold isolates its causal core.

\subsubsection{Head-Level Convergence: Encoding Versus Causal Use}
\label{app:head_convergence}

Attention heads were ranked by six per-head metrics and the Spearman rank
correlation taken between every pair\citep{olsson2022context}. Rank correlation is scale-invariant and
so is directly comparable across grids ranging from 208 heads in Gemma2-2B to
1600 in Qwen3-14B. The metrics divide a priori into two families. The
\emph{representational} family asks whether a head encodes the horizon: probe
accuracy on the head's output, attribution of the head onto the temporal axis,
and the shift in the head's attention entropy between conditions. The
\emph{causal} family asks whether the head is used: the metric drop when the
head is patched, and the KL shift under single-head ablation.

Table~\ref{tab:head_convergence} gives the result. Both families are internally
coherent --- probe accuracy and attribution correlate at $0.35$ to $0.69$ in
every model and pull entropy shift along with them, while the two causal
metrics correlate at $0.23$ to $0.66$ in six of seven models --- but the two
families are close to orthogonal to each other. Every cross-family correlation
falls between $0.04$ and $0.31$. A head can score highly on the
representational axis and contribute almost nothing causally, and the reverse.

\begin{table}[htbp]
\centering
\caption{Spearman rank correlations between per-head metrics under explicit
anchoring. The representational block (first three columns) and the causal
block (fourth column) are each internally coherent, while the cross-family
correlations (last two columns) are uniformly weak.}
\label{tab:head_convergence}
\begin{tabular}{|l|r|r|r|r|r|r|}
\hline
\multirow{2}{*}{\textbf{Model}} & \multicolumn{3}{c|}{\textbf{Representational}} & \textbf{Causal} & \multicolumn{2}{c|}{\textbf{Cross-family}} \\ \cline{2-7}
 & \textbf{pr.\ $\sim$ at.} & \textbf{pr.\ $\sim$ en.} & \textbf{at.\ $\sim$ en.} & \textbf{ACDC $\sim$ abl.} & \textbf{pr.\ $\sim$ ACDC} & \textbf{pr.\ $\sim$ abl.} \\ \hline
Qwen3-1.7B   & 0.49 & 0.23 & 0.23 & $0.56$    & 0.20 & 0.14 \\ \hline
Qwen3-4B     & 0.64 & 0.24 & 0.34 & $0.46$    & 0.31 & 0.22 \\ \hline
Qwen3-8B     & 0.35 & 0.18 & 0.22 & $0.38$    & 0.15 & 0.09 \\ \hline
Qwen3-14B    & 0.65 & 0.21 & 0.33 & $0.23$    & 0.20 & 0.07 \\ \hline
Llama-3.1-8B & 0.69 & 0.49 & 0.52 & $0.52$    & 0.24 & 0.18 \\ \hline
Gemma2-2B    & 0.60 & 0.46 & 0.57 & $0.66$    & 0.04 & 0.22 \\ \hline
Pythia-12B   & 0.66 & 0.59 & 0.58 & $-0.12$   & 0.09 & 0.19 \\ \hline
\end{tabular}
\end{table}

The dissociation is starker still at the extremes.
Table~\ref{tab:head_jaccard} reports the Jaccard overlap between the ten most
causally important heads and the ten most probe-decodable heads: it is
approximately zero almost everywhere, and exactly zero in the majority of
cases. The single most causally important head is essentially never among the
most probe-decodable heads. This is not an artifact of ties or saturation --- the
underlying probe accuracies have genuine spread, with medians between $0.93$
and $0.98$ and more than twelve distinct values among the top fifteen heads ---
so the disjoint top-sets are real.

\begin{table}[htbp]
\centering
\caption{Jaccard overlap between the top-10 causally important heads and the
top-10 representationally selective heads. Overlap is near zero in every model
and both conditions.}
\label{tab:head_jaccard}
\begin{tabular}{|l|c|c|c|}
\hline
\textbf{Model} & \textbf{ACDC $\cap$ probe (exp.)} & \textbf{ACDC $\cap$ probe (imp.)} & \textbf{ACDC $\cap$ attrib.\ (exp.)} \\ \hline
Qwen3-1.7B   & 0.05 & 0.05 & 0.00 \\ \hline
Qwen3-4B     & 0.00 & 0.00 & 0.00 \\ \hline
Qwen3-8B     & 0.00 & 0.00 & 0.00 \\ \hline
Qwen3-14B    & 0.00 & 0.11 & 0.00 \\ \hline
Llama-3.1-8B & 0.00 & 0.00 & 0.00 \\ \hline
Gemma2-2B    & 0.00 & 0.18 & 0.00 \\ \hline
Pythia-12B   & 0.00 & 0.05 & 0.00 \\ \hline
\end{tabular}
\end{table}

Two further observations qualify the picture. Within the Qwen family,
representational coherence is stable across the scale sweep while causal
coherence \emph{decays} monotonically with scale, the correlation between the
two causal metrics falling from $0.56$ at 1.7B to $0.23$ at 14B. As models
grow, circuit patching and single-head ablation increasingly disagree, which is
consistent with larger models spreading the causal effect across more,
partly redundant heads so that no single-head measure captures it cleanly. The
representational-versus-causal gap is nonetheless present at every size, so the
dissociation is a property of the architecture rather than something that
emerges only with scale. Separately, Pythia-12B is a causal-incoherence
outlier: it is the only model whose two causal metrics correlate negatively,
even though its representational family is the tightest in the sweep. Pythia-12B
represents the horizon cleanly in many heads without yielding any coherent
causal circuit.

Taken together, these results support a specific and somewhat cautionary
conclusion. The temporal horizon is robustly encoded, in a low-dimensional
linear subspace, distributed across many attention heads. But the heads that
encode it are not the heads that use it, and the causal work is carried largely
by an MLP backbone that is nearly condition-invariant and nearly
non-selective. Identifying a small set of interpretable ``temporal specialist
heads'' that both represent and causally control horizon-dependent behavior is
therefore not supported by the present evidence, and remains open work.

\newpage
\section{Logarithmic Encoding Analysis}
\label{appendix:log_encoding}
This section outlines the additional experiments and statistical 
validations conducted to support a key claim of this paper: the 
temporal horizon is encoded logarithmically with respect to the 
number of days. Five distinct model fittings were evaluated to 
identify the most suitable statistical model for representing the 
temporal horizon. The experimental details remain consistent with 
those discussed in Section~\ref{sec:log-encoding}.
\begin{table}[htbp]
\centering
\caption{Candidate functional forms tested for modeling temporal horizon representations, along with their mathematical definitions and number of free parameters.}
\label{tab:candidate_models}
\begin{tabular}{|l|c|c|}
\hline
\textbf{Model} & \textbf{$f(x)$} & \textbf{Free params} \\ \hline
\textbf{Log-linear} (Weber--Fechner) & $a \cdot \log(x) + b$ & 2 \\ \hline
\textbf{Power-law} (Stevens\citep{stevenslaw}) & $a \cdot x^\alpha + b$ & 3 \\ \hline
\textbf{Linear-in-rank} & $a \cdot \text{rank} + b$ & 2 \\ \hline
\textbf{Piecewise-linear} (2-segment) & $a \cdot \log(x) + b \text{ for } x \le x_0, \text{ else } c \cdot \log(x) + d$ & 4 \\ \hline
\textbf{Sigmoid-in-log} & $\frac{A}{1 + \exp\left(-\frac{\log(x) - \mu}{s}\right)} + B$ & 4 \\ \hline
\end{tabular}
\end{table}

Table~\ref{tab:candidate_models} displays five distinct functional 
forms that were evaluated for modeling representations of temporal 
horizons. For each layer, we fitted these five models to the equation 
\texttt{mean\_projection $\sim$ f(horizon\_days)}. The term ``mean
projection'' denotes the averaging of representations for all
examples that share a common temporal horizon (i.e., horizon days). 
These averages are projected onto the temporal direction derived 
from the probe weights, enabling us to compare the fit with the 
logarithm of the number of days.

Model selection was performed using the Akaike Information Criterion 
(AIC)\citep{aic} and the Bayesian Information Criterion (BIC)\citep{bic}. Bootstrapping 
across prompts (500 resamples) was conducted to derive confidence 
intervals for $R^2$, slope parameters, $\Delta\mathrm{AIC}$, and 
$\Delta\mathrm{BIC}$.
\begin{itemize}
    \item \textbf{AIC and BIC Selection Metrics}: Both metrics 
    evaluate model parsimony by balancing goodness-of-fit against 
    parameter complexity, penalizing over-parameterized models to 
    prevent selection driven purely by overfitting.
    \item \textbf{Penalty Dynamics}: Whereas AIC imposes a constant 
    penalty per parameter ($2k$), BIC scales the penalty 
    logarithmically with sample size ($k \ln n$). Consequently, 
    for sample sizes $n \ge 8$, BIC penalizes additional complexity 
    more strictly than AIC, favoring parsimonious functional forms 
    (e.g., log-linear) unless a complex model demonstrates a 
    substantial improvement in fit.
    \item \textbf{Decision Thresholds}: Following established 
    information-theoretic conventions, differences in information 
    criteria ($\Delta = \mathrm{IC}_{\text{competitor}} - \mathrm{IC}_{\text{log-linear}}$) 
    are interpreted as:
    \begin{itemize}
        \item $\Delta > +6$: Decisive evidence favoring the log-linear 
        model over the competitor.
        \item $\Delta < -6$: Decisive evidence favoring the competing 
        model over log-linear.
        \item $-6 \le \Delta \le +6$: Inconclusive evidence, indicating 
        no strong preference for either model.
    \end{itemize}
\end{itemize}

Tables~\ref{tab:delta_aic} and \ref{tab:delta_bic} report the 
empirical $\Delta\mathrm{AIC}$ and $\Delta\mathrm{BIC}$ values 
across evaluated language model architectures. Across all models, 
the log-linear functional form is consistently preferred over 
linear-in-rank and power-law alternatives ($\Delta > 0$). While more 
complex candidate models (such as piecewise-linear) achieve higher 
raw likelihoods on select architectures, the information-theoretic 
parameter penalties in AIC and BIC demonstrate that the parsimonious 
log-linear fit is the most statistically justified model overall.

\begin{table}[htbp]
\centering
\caption{$\Delta\mathrm{AIC}$ model comparison relative to the log-linear baseline. Negative values indicate a better fit than log-linear; values $\Delta < -6$ indicate decisive evidence in favor of the competitor.}
\label{tab:delta_aic}
\begin{tabular}{|l|c|c|c|c|}
\hline
\textbf{Model} & \textbf{Linear-in-rank} & \textbf{Piecewise-linear} & \textbf{Power-law} & \textbf{Sigmoid-in-log} \\ \hline
Gemma2-2B     & 3.38 & -19.40 & 2.00 &  -7.88 \\ \hline
Llama-3.1-8B  & 4.77 & -19.37 & 2.00 & -12.99 \\ \hline
Pythia-12B    & 2.98 & -19.58 & 2.00 &  -9.09 \\ \hline
Qwen3-1.7B    & 3.36 & -12.70 & 2.00 &  -9.39 \\ \hline
Qwen3-14B     & 2.59 & -10.69 & 1.96 &  -8.82 \\ \hline
Qwen3-4B      & 2.76 & -12.64 & 2.00 &  -9.84 \\ \hline
Qwen3-8B      & 3.73 & -14.71 & 2.00 & -10.61 \\ \hline
\end{tabular}
\end{table}

\begin{table}[htbp]
\centering
\caption{$\Delta\mathrm{BIC}$ model comparison relative to the log-linear baseline. Negative values indicate a better fit than log-linear; values $\Delta < -6$ indicate decisive evidence in favor of the competitor.}
\label{tab:delta_bic}
\begin{tabular}{|l|c|c|c|c|}
\hline
\textbf{Model} & \textbf{Linear-in-rank} & \textbf{Piecewise-linear} & \textbf{Power-law} & \textbf{Sigmoid-in-log} \\ \hline
Gemma2-2B     & 3.38 & -18.79 & 2.30 &  -7.27 \\ \hline
Llama-3.1-8B  & 4.77 & -18.76 & 2.31 & -12.38 \\ \hline
Pythia-12B    & 2.98 & -18.98 & 2.30 &  -8.49 \\ \hline
Qwen3-1.7B    & 3.36 & -12.10 & 2.30 &  -8.79 \\ \hline
Qwen3-14B     & 2.59 & -10.09 & 2.26 &  -8.22 \\ \hline
Qwen3-4B      & 2.76 & -12.03 & 2.30 &  -9.24 \\ \hline
Qwen3-8B      & 3.73 & -14.11 & 2.31 & -10.01 \\ \hline
\end{tabular}
\end{table} 

A language model representation is defined as 
\textbf{Weber--Fechner-valid} when all three of the following 
conditions hold simultaneously:
\begin{enumerate}
    \item \textbf{Goodness of Fit}: Mean log-linear $R^2 > 0.70$, establishing that the temporal representation axis is strongly logarithmic.
    \item \textbf{Superiority over Ordinal Rival}: $\Delta\mathrm{BIC}$ relative to the \textit{linear-in-rank} model is positive ($\Delta\mathrm{BIC} > 0$), demonstrating that metric temporal spacing---rather than ordinal rank alone---governs the representation space.
    \item \textbf{Superiority over Power-Law Rival}: $\Delta\mathrm{BIC}$ relative to the \textit{power-law} model is non-negative ($\Delta\mathrm{BIC} \ge 0$), confirming that Stevens-style power scaling is not preferred over logarithmic scaling.
\end{enumerate}

The two 4-parameter functional forms (piecewise-linear and
sigmoid-in-log) are \textit{strict supersets} of the log-linear model. We
report them for contextual baseline evaluation but exclude them
from the parsimony tests, because they introduce secondary saturation or
breakpoint mechanisms rather than altering the fundamental
logarithmic relationship.

\begin{table}[htbp]
\centering
\caption{Validation metrics for Weber--Fechner (logarithmic) temporal horizon encoding across all seven evaluated models (excluding the \textit{Valid} status column).}
\label{tab:weber_fechner_validation}
\small
\begin{tabular}{|l|c|c|c|c|c|c|c|}
\hline
& & & \multicolumn{2}{c|}{$\Delta\mathrm{AIC}$} & \multicolumn{2}{c|}{$\Delta\mathrm{BIC}$} & \\
\cline{4-7}
\textbf{Model} & \textbf{Mean $R^2$} & \textbf{Late $R^2$} & \textbf{vs. Rank} & \textbf{vs. Power} & \textbf{vs. Rank} & \textbf{vs. Power} & \textbf{Log Win \%} \\ \hline
Qwen3-1.7B   & 0.923 & 0.936 & +3.36 & +2.00 & +3.36 & +2.30 & 100\% \\ \hline
Qwen3-4B     & 0.944 & 0.974 & +2.76 & +2.00 & +2.76 & +2.30 &  94\% \\ \hline
Qwen3-8B     & 0.936 & 0.956 & +3.73 & +2.00 & +3.73 & +2.31 & 100\% \\ \hline
Qwen3-14B    & 0.954 & 0.983 & +2.59 & +1.96 & +2.59 & +2.26 &  92\% \\ \hline
Llama-3.1-8B & 0.953 & 0.972 & +4.77 & +2.00 & +4.77 & +2.31 & 100\% \\ \hline
Gemma2-2B    & 0.921 & 0.946 & +3.38 & +2.00 & +3.38 & +2.30 & 100\% \\ \hline
Pythia-12B   & 0.905 & 0.905 & +2.98 & +2.00 & +2.98 & +2.30 &  97\% \\ \hline
\end{tabular}
\end{table}

The empirical results in Table~\ref{tab:weber_fechner_validation} 
demonstrate strong validation of logarithmic temporal encoding across
three key dimensions:

\begin{enumerate}
    \item \textbf{Goodness of Fit for the Log Law}: Across all layers 
    and evaluated architectures, the pooled log-linear fit achieves 
    $R^2 = \mathbf{0.934}$. Every individual model achieves a mean
    $R^2 > 0.90$, with fit quality rising to $R^2 \approx 0.91\text{--}0.98$
    in the deepest quarter of layers where temporal directions are 
    cleanest. Even Pythia-12B (the lowest-scoring architecture) 
    exhibits a mean $R^2 = 0.905$, indicating that logarithmic 
    encoding is universal.

    \item \textbf{Metric Spacing Outperforms Ordinal Ranking (Core Weber--Fechner Test)}: The log-linear model outperforms the \textit{linear-in-rank} baseline---which considers only ordinal horizon order rather than metric log-distance---in \textbf{97.4\%} of all layer-wise fits, yielding positive $\Delta\mathrm{AIC}$ and $\Delta\mathrm{BIC}$ values across all models ($+2.6$ to $+4.8$). Notably, $\Delta\mathrm{AIC}$ and $\Delta\mathrm{BIC}$ match identically for linear-in-rank because both models share the same parameter count ($k = 2$), causing complexity penalty terms to cancel ($\Delta k = 0$). This establishes that representations encode the numerical \textit{magnitude} of elapsed time on a logarithmic scale rather than mere ordinal rank.

    \item \textbf{Logarithmic Scaling Outperforms Power-Law Scaling (Weber--Fechner vs.\ Stevens)}: While the 3-parameter power-law model achieves near-identical raw fit quality to three decimal places, its fitted exponent collapses toward the logarithmic limit, rendering the additional parameter redundant. Consequently, log-linear is preferred by BIC in \textbf{100\%} of fits ($\Delta\mathrm{BIC} \approx +2.3$ across all models) and by AIC in the vast majority of fits ($\Delta\mathrm{AIC} \approx +2.0$). Because AIC imposes a constant penalty of $2$ for the extra parameter while BIC imposes a penalty of $\ln(n) \approx 2.3$ ($n=10$ horizons per fit), BIC displays a strictly stronger preference for log-linear by a constant difference of $\approx 0.3$. Thus, Stevens-style power-law scaling is disfavored under both criteria.
\end{enumerate}

The only candidate functional forms that yield higher raw fit metrics 
(piecewise-linear $R^2 \approx 0.99$ and sigmoid-in-log $R^2 \approx 
0.97$) represent \textbf{strict supersets} of the log-linear model. 
Their additional parameters capture mild \textbf{saturation at 
multi-century temporal horizons}, bending the representation curve 
slightly in late layers. This represents a secondary refinement 
\textit{on top of} the fundamental logarithmic law rather than an 
alternative to it. 

Therefore, we can safely conclude that temporal horizon is encoded 
logarithmically (\textit{Weber--Fechner}) universally across model 
architectures and parameter scales, representing the most 
statistically justified functional form once model parsimony is 
accounted for under both AIC and BIC. 
\newpage
\section{Geometrical Analysis}
\label{app:geometrical_analysis}

The main text presents the manifold-geometry and
persistent-homology results in summary form. This appendix supplies the
procedure by which the temporal manifold is constructed and visualized
(Appendix~\ref{app:manifold_construction}), the
per-model quantities behind those figures
(Appendix~\ref{app:geometry_descriptors}), the isometry analysis that is the
one geometric descriptor computed for both anchoring conditions
(Appendix~\ref{app:geometry_isometry}), the persistent-homology diagnostics not
shown in the main text (Appendix~\ref{app:geometry_persistent_homology}), the
intrinsic-dimensionality and rank analyses
(Appendix~\ref{app:geometry_intrinsic_dim}), and supplementary steering results
including the control that motivates the causal readout used in the main text
(Appendix~\ref{app:geometry_steering}).

A caveat governs this entire appendix. The mean activation norm grows
monotonically with depth in every model, by factors between $5\times$ and
$192\times$ (Table~\ref{tab:raw_norm}). Any unnormalized magnitude-based
quantity therefore inherits a strong depth trend that has nothing to do with
temporal structure. Where a metric is norm-sensitive we report the
norm-normalized version and say so explicitly.

\subsection{Construction of the Temporal Manifold}
\label{app:manifold_construction}

The manifold visualizations shown in Figure~\ref{fig:geometry_figures} and all
of the geometric descriptors reported below are derived from a single
construction, which follows the concept-manifold procedure of
\citep{wurgaft2026manifoldsteeringrevealsshared}: concept centroids are computed
in activation space, a cubic spline is fitted through them in concept order, and
geodesic quantities are measured as cumulative Euclidean distance along that
spline. This subsection states the construction precisely as implemented.

\paragraph{Activation collection.}
Let $K = 10$ be the number of graded horizon levels and let $S_k$ denote the set
of Horizon-Graded prompts realizing level $k$, ordered from shortest
(\texttt{1\_day}) to longest (\texttt{500\_years}). For a given layer $\ell$ we
take the residual-stream activation at the output of transformer block $\ell$,
read at the \emph{final token position} only:
$h^{(\ell)}(x) \in \mathbb{R}^{d_{\text{model}}}$. Restricting to the final
position means each prompt contributes a single vector, so the ten levels are
directly comparable across a fixed scenario stem.

\paragraph{Concept centroids.}
For each level we form the centroid, the mean activation over all prompts
realizing that level:
\begin{equation}
    \boldsymbol{\mu}^{(\ell)}_k = \frac{1}{|S_k|} \sum_{x \in S_k} h^{(\ell)}(x),
    \qquad k = 1, \dots, K .
\end{equation}
Because the ten variations of a Horizon-Graded instance are identical
word-for-word apart from the temporal phrase, averaging over instances cancels
scenario content and leaves the horizon as the dominant axis of variation
between centroids.

\paragraph{Dimensionality reduction.}
Principal component analysis is fitted \emph{per layer} on the pooled
activations of all levels, $\bigcup_{k} S_k$, rather than on the ten centroids
themselves; the centroids are then projected into the resulting basis. Fitting on
the pooled samples means the retained components capture the directions of
greatest variance in the layer's representation of the whole graded set, not
merely the directions that happen to separate ten averaged points. Writing
$P_\ell$ for the projection, the reduced centroids are
$\tilde{\boldsymbol{\mu}}^{(\ell)}_k = P_\ell \boldsymbol{\mu}^{(\ell)}_k$. Two
settings are used: $m = 3$ components for the visualizations, so the manifold
can be rendered in three dimensions, and $m = 10$ components for every
quantitative descriptor reported in this appendix.

\paragraph{Manifold fitting.}
A natural cubic spline $\gamma_\ell : [1, K] \to \mathbb{R}^m$ is fitted
through the ordered reduced centroids, parameterized by level index, so that
$\gamma_\ell(k) = \tilde{\boldsymbol{\mu}}^{(\ell)}_k$ exactly at every integer
parameter. The spline interpolates rather than regresses: it passes through
every centroid, and the curvature between consecutive centroids is the minimal
curvature consistent with a twice-differentiable curve through all of them. The
red curve in Figure~\ref{fig:geometry_figures} is $\gamma_\ell$ for the
indicated layer, and the black diamonds are the
$\tilde{\boldsymbol{\mu}}^{(\ell)}_k$.

\paragraph{Arc length and geodesic distance.}
The spline is sampled at $N$ equally spaced parameter values
$t_1 < \dots < t_N$ spanning $[1, K]$, and arc length is accumulated as the
Euclidean polyline length through those samples:
\begin{equation}
    s(t_j) = \sum_{i=1}^{j-1} \left\lVert \gamma_\ell(t_{i+1}) - \gamma_\ell(t_i) \right\rVert_2 ,
    \qquad
    L_\ell = s(t_N),
    \label{eq:arclength}
\end{equation}
which converges to $\int_1^{K} \lVert \gamma_\ell'(t) \rVert_2 \, dt$ as
$N \to \infty$. We use $N = 500$ for all reported descriptors. The geodesic
distance between two levels is the arc length of the spline between them,
\begin{equation}
    d_{\text{geo}}(i, j) = \bigl| s(i) - s(j) \bigr| ,
\end{equation}
obtained by interpolating $s$ at the integer parameters. This along-manifold
distance --- rather than the straight-line chord --- is what enters the isometry
analysis of Appendix~\ref{app:geometry_isometry}.

\paragraph{Descriptors.}
Tortuosity is the ratio of the fitted manifold's arc length to the
straight-line distance between its endpoints,
\begin{equation}
    \tau_\ell = \frac{L_\ell}{\bigl\lVert \tilde{\boldsymbol{\mu}}^{(\ell)}_K - \tilde{\boldsymbol{\mu}}^{(\ell)}_1 \bigr\rVert_2} ,
    \label{eq:tortuosity_impl}
\end{equation}
with $\tau_\ell = 1$ for a perfectly straight trajectory and $\tau_\ell > 1$ for
a curved one; it is undefined when the endpoints coincide. Centroid spacing is
summarized by the mean and coefficient of variation of the consecutive chord
lengths $\lVert \tilde{\boldsymbol{\mu}}_{k+1} - \tilde{\boldsymbol{\mu}}_k \rVert_2$,
where a low coefficient of variation indicates the horizon is encoded in roughly
equal steps per level and a high one indicates compression at one end. The mean
raw norm of Table~\ref{tab:raw_norm} is computed on the \emph{unreduced}
activations, $\frac{1}{K}\sum_k \frac{1}{|S_k|}\sum_{x \in S_k} \lVert h^{(\ell)}(x) \rVert_2$,
so that it measures the true scale of the residual stream rather than the scale
of its projection.

Equation~\ref{eq:tortuosity_impl} is identical to the definition given in the
main text; both measure arc length along the fitted spline via
Equation~\ref{eq:arclength}, and this is the quantity reported in
Table~\ref{tab:tortuosity} and throughout the paper. We note for clarity that
this spline arc length upper-bounds the simpler piecewise-linear alternative
that sums the chords between consecutive centroids, since an interpolating cubic
bows outward between the points it passes through; the two coincide only when
the trajectory is exactly straight, in which case $\tau = 1$.

\subsection{Manifold Descriptors: Tortuosity and Activation Norm}
\label{app:geometry_descriptors}

Table~\ref{tab:tortuosity} reports path tortuosity $\tau$ averaged within each
third of the network, together with its Spearman correlation\citep{spearmancorrelation} against depth. The
qualitative pattern described in the main text holds across the Qwen family:
tortuosity is maximal at the embedding layer, falls to a mid-network trough, and
re-curves slightly toward the output. The embedding-layer peak is close to
scale-invariant across the Qwen sweep, at $\tau \approx 7.3$--$8.0$ for all four
sizes, but the rate of decline flattens monotonically with scale, with $\rho$
moving from $-0.74$ at 1.7B to $-0.24$ at 14B. Larger models keep the
short-to-long arc more uniformly curved through depth rather than aggressively
straightening it in the middle.

Two architectures depart from the pattern. Gemma2-2B is the cleanest monotone
straightening in the sweep ($\rho = -0.87$) and remains the least curved overall.
Llama-3.1-8B is the only model whose tortuosity \emph{rises} with depth
($\rho = +0.28$): it straightens sharply out of the embeddings and then
re-curves toward the output, ending more tortuous than its middle layers. This
is the same directional anomaly Llama-3.1-8B exhibits in its CKA profile, its
activation patching, and its crossover fit.

\begin{table}[htbp]
\centering
\caption{Path tortuosity $\tau$ averaged over each third of the network, with
the Spearman correlation against depth and the values at the first and final
layers. Computed on the explicit runs. Llama-3.1-8B is the only model with a
positive depth trend.}
\label{tab:tortuosity}
\begin{tabular}{|l|r|r|r|r|r|r|}
\hline
\textbf{Model} & \textbf{$\tau$ early} & \textbf{mid} & \textbf{late} & \textbf{$\rho(\tau, \text{depth})$} & \textbf{$\tau_{L_0}$} & \textbf{$\tau_{\text{final}}$} \\ \hline
Qwen3-1.7B   & 6.14 & 4.29 & 4.17 & $-0.74$ & 7.79 & 4.76 \\ \hline
Qwen3-4B     & 6.18 & 4.76 & 5.04 & $-0.57$ & 7.97 & 7.27 \\ \hline
Qwen3-8B     & 5.32 & 3.74 & 3.95 & $-0.54$ & 7.35 & 5.65 \\ \hline
Qwen3-14B    & 5.31 & 3.25 & 3.96 & $-0.24$ & 7.34 & 6.05 \\ \hline
Llama-3.1-8B & 4.01 & 3.40 & 3.88 & $+0.28$ & 6.88 & 3.74 \\ \hline
Gemma2-2B    & 4.81 & 4.22 & 3.65 & $-0.87$ & 4.87 & 3.46 \\ \hline
Pythia-12B   & 4.84 & 3.50 & 3.98 & $-0.12$ & 6.22 & 4.11 \\ \hline
\end{tabular}
\end{table}

Table~\ref{tab:raw_norm} quantifies the norm growth referred to above. It is the
most universal geometric signal in this work: the correlation with depth is at
least $+0.96$ in every model and exactly $+1.00$ across the entire Qwen sweep.
The magnitude of the blow-up, however, splits the architectures into two groups
that differ by more than an order of magnitude. Qwen and Llama inflate the
residual norm by roughly $90$--$190\times$ over depth, whereas Gemma2-2B and
Pythia-12B barely move, at $5\times$ and $7\times$ respectively. Within the Qwen
family the growth factor does not track model size --- 1.7B is the steepest at
$192\times$ and 4B the mildest at $90\times$ --- so it is better read as an
architectural property of the residual design than as a scale effect. Absolute
scale differs by two orders of magnitude across models as well, which is why
only scale-invariant summaries are comparable across families.

\begin{table}[htbp]
\centering
\caption{Mean $L_2$ norm of the concept centroids, averaged over each third of
the network, with the first-layer and final-layer values and the resulting
growth factor. Norm growth is monotone in every model but its magnitude splits
the architectures into two groups.}
\label{tab:raw_norm}
\begin{tabular}{|l|r|r|r|r|}
\hline
\textbf{Model} & \textbf{$\rho$} & \textbf{$\lVert L_0 \rVert$} & \textbf{$\lVert L_{\text{final}} \rVert$} & \textbf{Growth} \\ \hline
Qwen3-1.7B   & $+1.00$ & 15.7 & 3015.6 & $192\times$ \\ \hline
Qwen3-4B     & $+1.00$ &  7.5 &  670.2 & $90\times$  \\ \hline
Qwen3-8B     & $+1.00$ & 10.5 & 1452.6 & $139\times$ \\ \hline
Qwen3-14B    & $+1.00$ & 16.7 & 1887.4 & $113\times$ \\ \hline
Llama-3.1-8B & $+1.00$ &  0.7 &   93.1 & $136\times$ \\ \hline
Gemma2-2B    & $+0.99$ & 89.9 &  423.9 & $5\times$   \\ \hline
Pythia-12B   & $+0.96$ & 85.8 &  585.1 & $7\times$   \\ \hline
\end{tabular}
\end{table}

\subsection{Isometry of the Temporal Manifold}
\label{app:geometry_isometry}

\begin{figure}[htbp]
    \centering
    \begin{subfigure}[b]{0.9\linewidth}
        \centering
        \includegraphics[width=\linewidth]{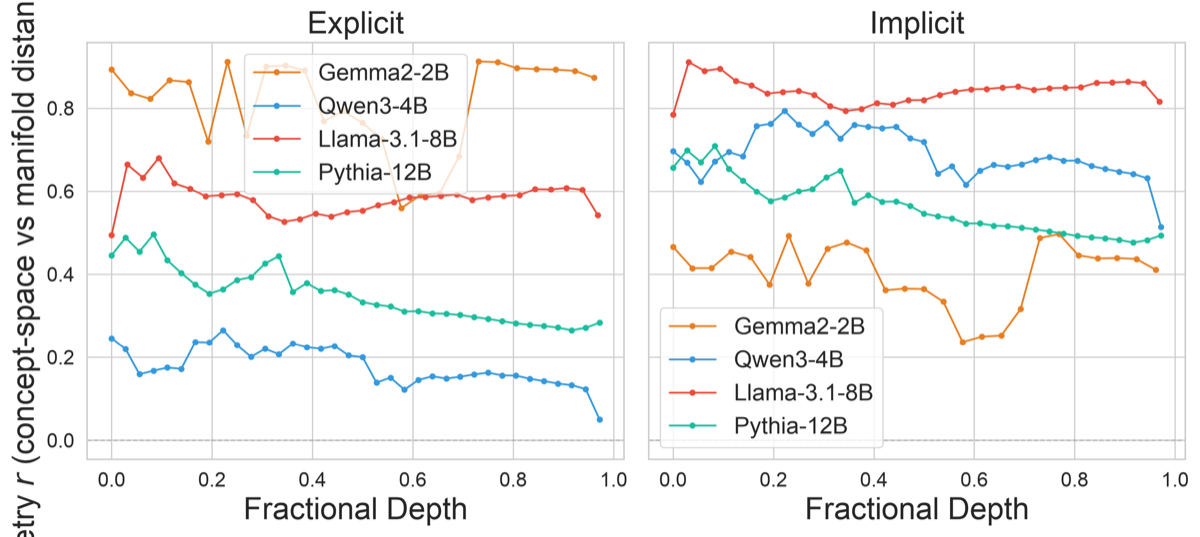}
        \caption{Across heterogeneous model families.}
        \label{fig:isometry_across}
    \end{subfigure}

    \vspace{0.3cm}

    \begin{subfigure}[b]{0.9\linewidth}
        \centering
        \includegraphics[width=\linewidth]{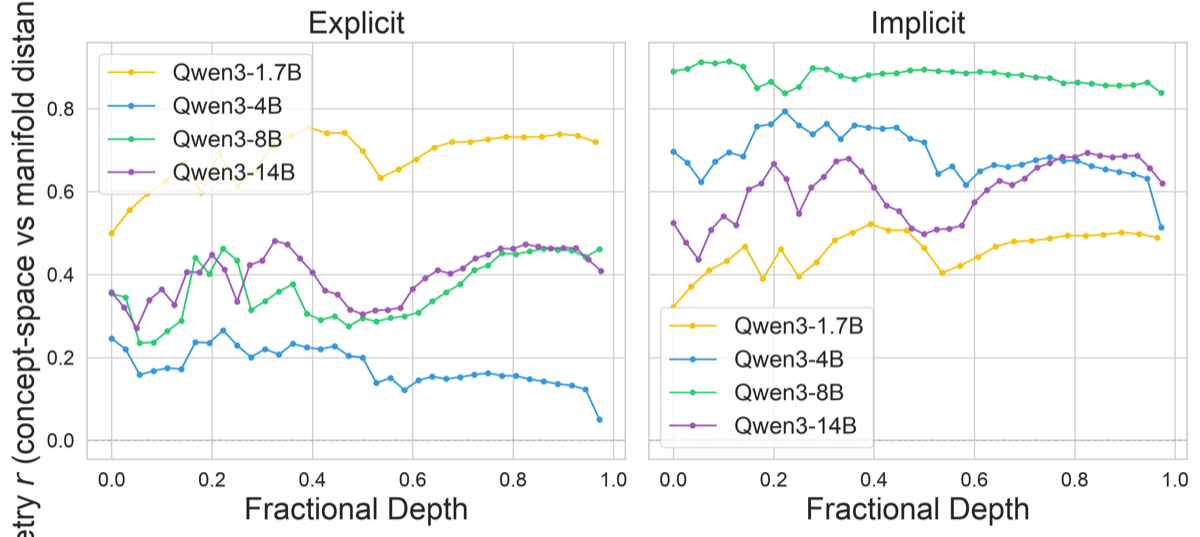}
        \caption{Within the Qwen3 model family.}
        \label{fig:isometry_within}
    \end{subfigure}

    \caption{Isometry coefficient across normalized layer depth for both
    anchoring conditions. Higher values indicate that pairwise concept distances
    are better preserved along the fitted manifold.}
    \label{fig:isometry}
\end{figure}

The isometry coefficient measures the correlation between pairwise distances in
concept space and pairwise distances along the fitted manifold\citep{wurgaft2026manifoldsteeringrevealsshared}. A value of one
means the layer preserves concept geometry isometrically; a value near zero
means the geometry is scrambled. This is the only geometric descriptor computed
for both anchoring conditions, so it is where the explicit-versus-implicit
contrast can be examined directly. Results are given in
Table~\ref{tab:isometry} and plotted against depth in
Figure~\ref{fig:isometry}.

\begin{table}[htbp]
\centering
\caption{Isometry coefficient averaged over all layers, with the Spearman
correlation against depth, for both anchoring conditions. The final column is
the explicit minus implicit difference; negative values indicate that implicit
anchoring preserves concept geometry better.}
\label{tab:isometry}
\begin{tabular}{|l|r|r|r|r|r|}
\hline
\multirow{2}{*}{\textbf{Model}} & \multicolumn{2}{c|}{\textbf{Explicit}} & \multicolumn{2}{c|}{\textbf{Implicit}} & \multirow{2}{*}{\textbf{Expl.\ $-$ Impl.}} \\ \cline{2-5}
 & \textbf{Mean} & \textbf{$\rho$} & \textbf{Mean} & \textbf{$\rho$} & \\ \hline
Qwen3-1.7B   & 0.68 & $+0.65$ & 0.46 & $+0.58$ & $+0.22$ \\ \hline
Qwen3-4B     & 0.18 & $-0.76$ & 0.69 & $-0.56$ & $-0.51$ \\ \hline
Qwen3-8B     & 0.36 & $+0.55$ & 0.88 & $-0.55$ & $-0.52$ \\ \hline
Qwen3-14B    & 0.40 & $+0.48$ & 0.60 & $+0.61$ & $-0.20$ \\ \hline
Llama-3.1-8B & 0.58 & $-0.04$ & 0.84 & $+0.12$ & $-0.26$ \\ \hline
Gemma2-2B    & 0.81 & $+0.08$ & 0.40 & $-0.16$ & $+0.41$ \\ \hline
Pythia-12B   & 0.35 & $-0.96$ & 0.56 & $-0.97$ & $-0.21$ \\ \hline
\end{tabular}
\end{table}

No single depth direction is universal, even within a family. Three of the four
Qwen sizes build explicit isometry with depth, while Qwen3-4B is a sharp
exception that loses it, ending at just $0.14$. The more striking pattern is a
condition asymmetry: in five of seven models implicit anchoring preserves
concept geometry better than explicit. Qwen3-8B is the extreme case, remaining
nearly isometric under implicit anchoring (mean $0.88$) while scrambled under
explicit (mean $0.36$). Read alongside the finding that implicit anchoring
builds class separation gradually over depth, this suggests implicit
representations do not merely separate the horizon classes later --- they lay
them out in a more geometry-faithful arrangement.

Two models stand apart. Gemma2-2B preserves explicit concept geometry best and
most stably, with a mean of $0.81$ and an essentially flat depth trend,
consistent with it being the least-reorganizing model on other diagnostics.
Pythia-12B is the isometry-collapse outlier: it is the only model whose geometry
degrades monotonically with depth in \emph{both} conditions
($\rho = -0.96$ and $-0.97$), so its deep layers actively scramble the
short-to-long distance structure.

\paragraph{Data caveat.} The Gemma2-2B isometry files were produced by an
earlier version of the analysis and report a single Pearson $p$-value rather
than the separate Pearson\citep{Pearson1895} and Mantel\citep{manteltest} $p$-values recorded for the other six
models. The isometry coefficients themselves are directly comparable; only the
significance testing differs.

\subsection{Additional Persistent-Homology Results}
\label{app:geometry_persistent_homology}

The main text reports only the overlaid Wasserstein\citep{Panaretos_2019wasserstein} and bottleneck distances
between the short- and long-horizon persistence diagrams. Four further
diagnostics were computed and are reported here. Throughout, $H_0$ denotes
connected components and $H_1$ denotes one-dimensional loops, and the point
clouds are $L_2$-normalized before the Vietoris--Rips\citep{carlsson2009topology} filtration so that only
direction and shape, not raw magnitude, drive the topology. The results were
calculated using the open-source libraries ripser~\citep{ctralie2018ripser,Bauer2021Ripser}
and persim~\citep{persim}.

\begin{figure}[htbp]
    \centering
    \begin{subfigure}[b]{0.9\linewidth}
        \centering
        \includegraphics[width=\linewidth]{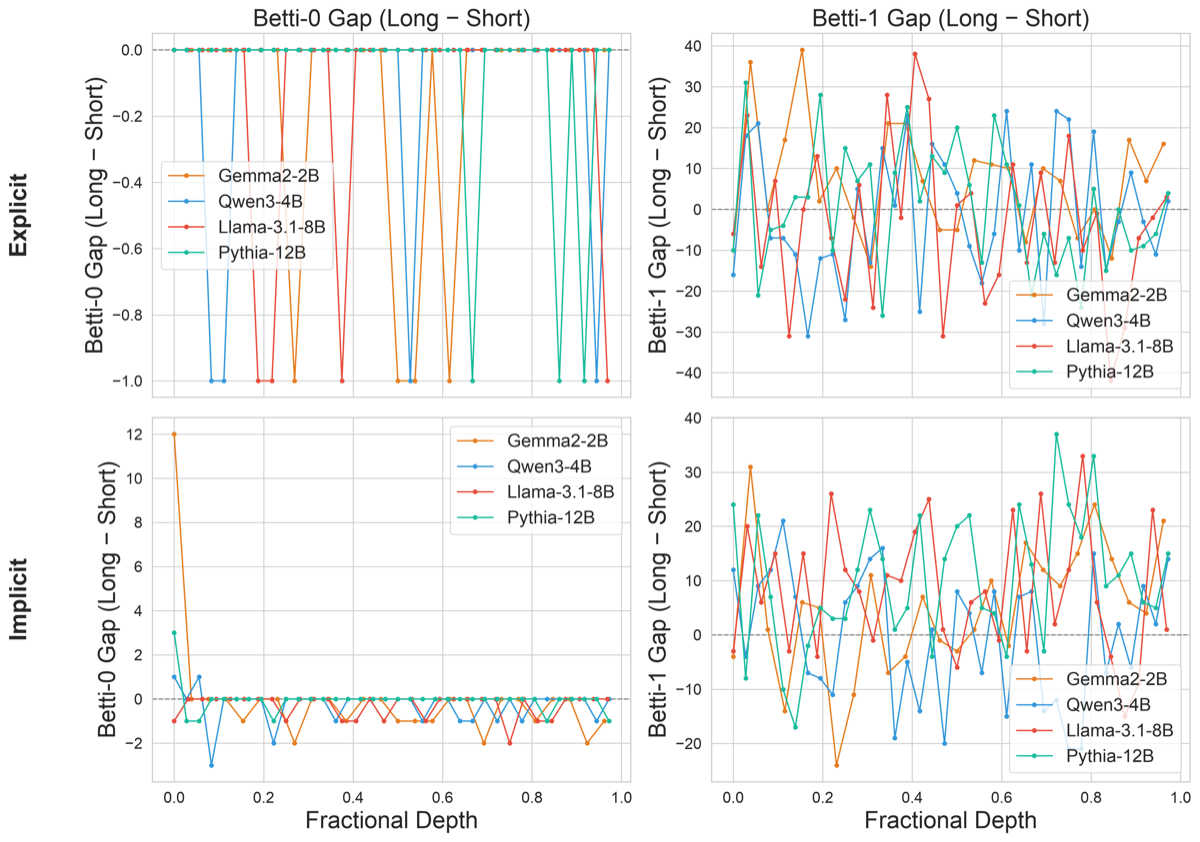}
        \caption{Across heterogeneous model families.}
        \label{fig:ph_betti_gap_across}
    \end{subfigure}

    \vspace{0.3cm}

    \begin{subfigure}[b]{0.9\linewidth}
        \centering
        \includegraphics[width=\linewidth]{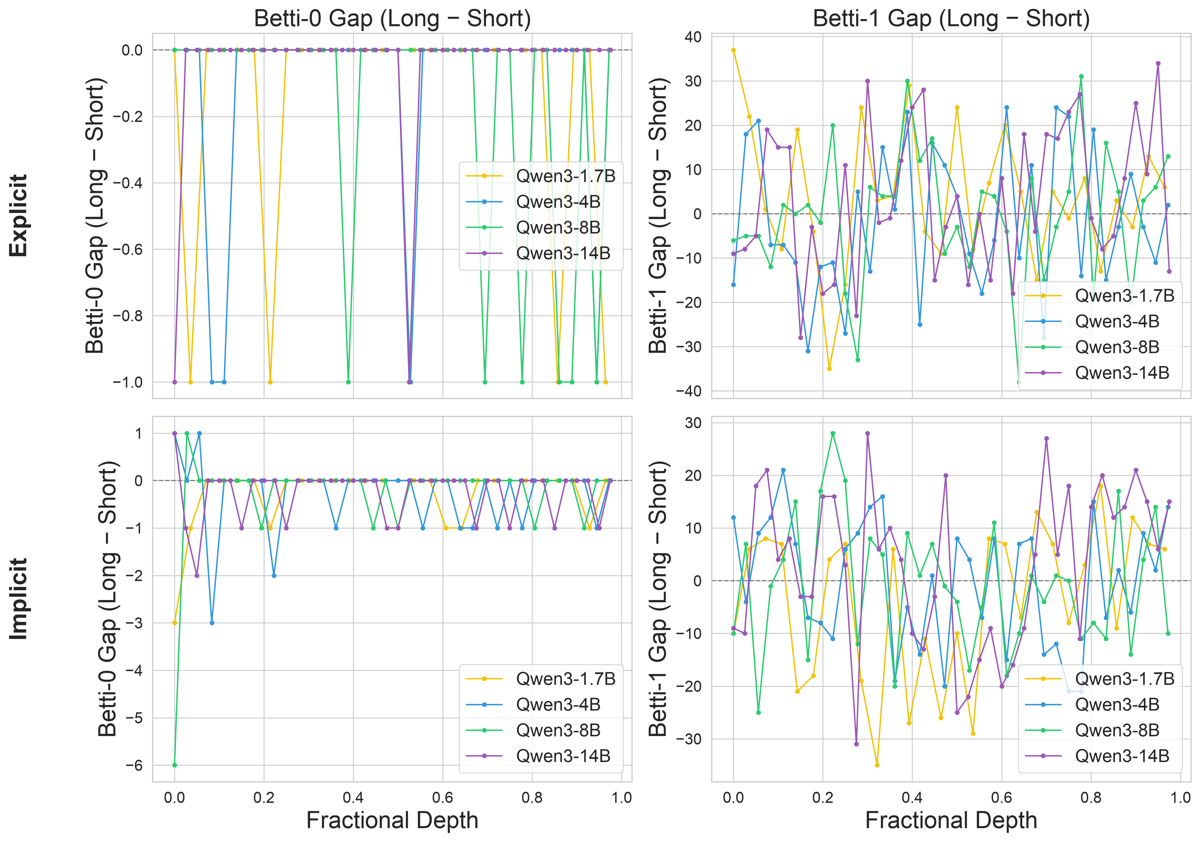}
        \caption{Within the Qwen3 model family.}
        \label{fig:ph_betti_gap_within}
    \end{subfigure}

    \caption[Betti-number gap across normalized layer depth for $H_0$ and $H_1$.]{Betti-number gap (long minus short) across normalized layer depth
    for $H_0$ and $H_1$. The $H_0$ gap is essentially zero throughout; the $H_1$
    gap oscillates about zero with substantial amplitude but no systematic depth
    trend. Neither shows a consistent difference in feature count between the
    two horizon conditions.}
    \label{fig:ph_betti_gap}
\end{figure}

\paragraph{Betti-number gap.}
Figure~\ref{fig:ph_betti_gap} shows the difference in persistent feature counts
between the long- and short-horizon clouds. The $H_0$ gap is essentially zero at
every depth in every model: the two conditions produce the same number of
persistent components. The $H_1$ gap oscillates around zero with no clear depth
trend. This is the weakest signal among the topological diagnostics, and we
report it because its flatness is informative --- whatever distinguishes the two
horizon classes, it is not a difference in the \emph{number} of topological
features.

\begin{figure}[htbp]
    \centering
    \begin{subfigure}[b]{0.9\linewidth}
        \centering
        \includegraphics[width=\linewidth]{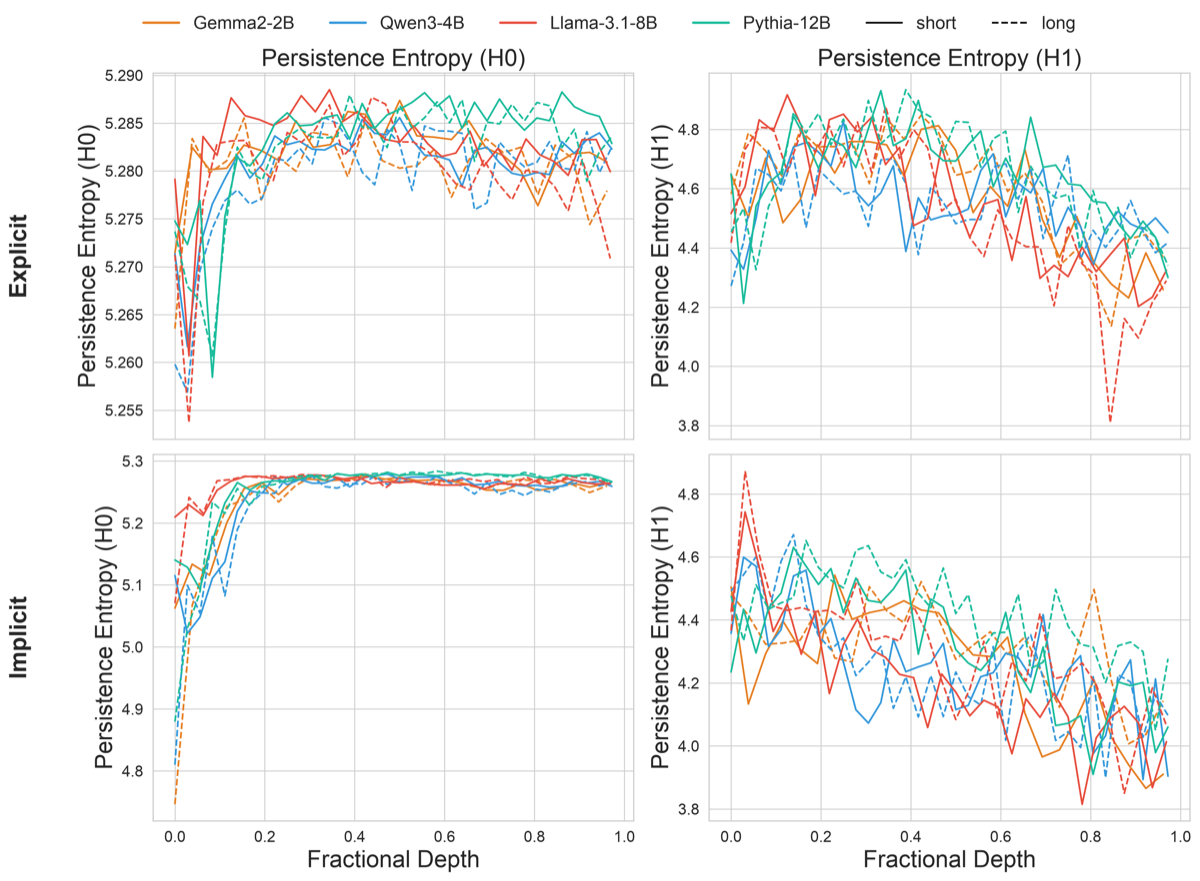}
        \caption{Across heterogeneous model families.}
        \label{fig:ph_entropy_across}
    \end{subfigure}

    \vspace{0.3cm}

    \begin{subfigure}[b]{0.9\linewidth}
        \centering
        \includegraphics[width=\linewidth]{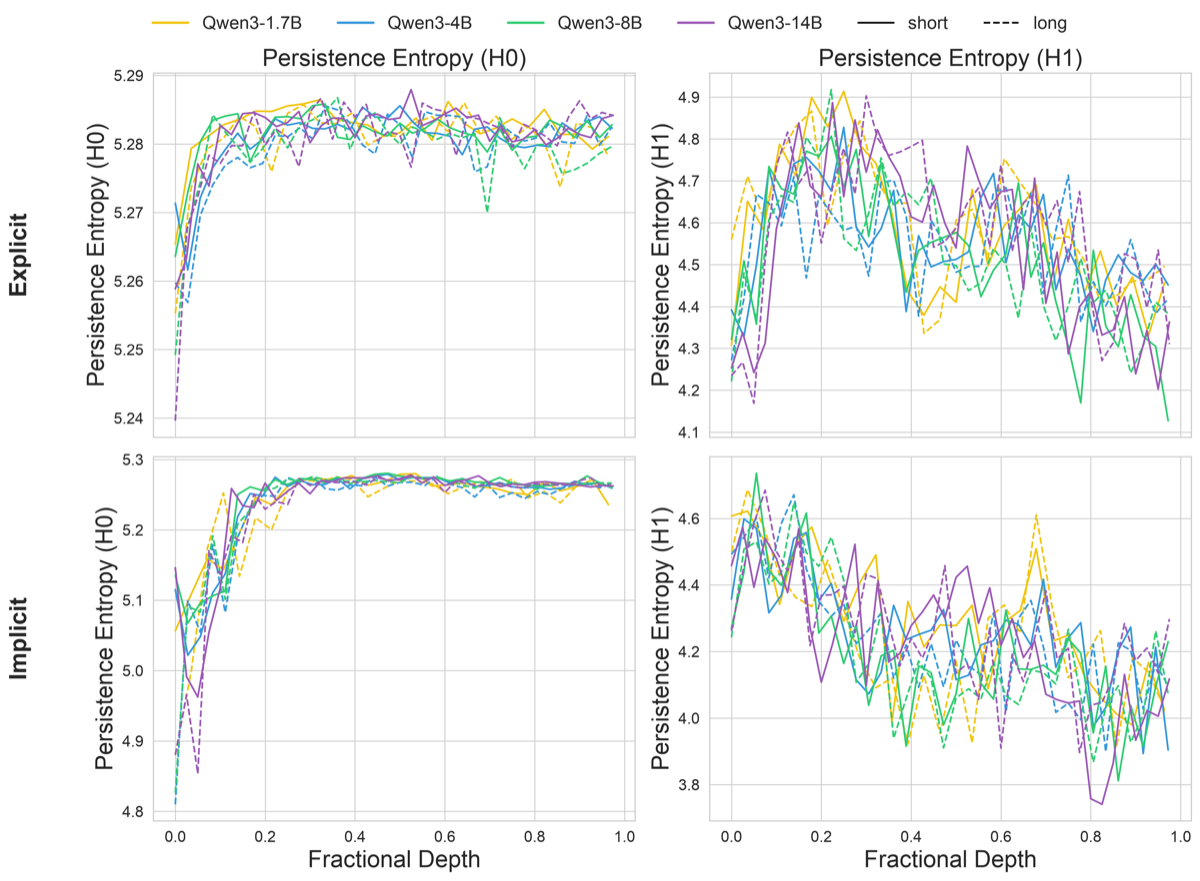}
        \caption{Within the Qwen3 model family.}
        \label{fig:ph_entropy_within}
    \end{subfigure}

    \caption{Persistence entropy across normalized layer depth. Low values
    indicate that a single topological feature dominates the lifetime
    distribution; high values indicate many comparably persistent features.}
    \label{fig:ph_entropy}
\end{figure}

\paragraph{Persistence entropy\citep{shannon1948mathematical}.}
Figure~\ref{fig:ph_entropy} reports the Shannon entropy of the normalized
feature-lifetime distribution, which is low when a single feature dominates and
high when many features are comparably important. The $H_0$ entropy saturates
almost immediately, by roughly a fifth of the way through the network, and stays
flat thereafter. The $H_1$ entropy rises and then decays toward the output
layers, with the short and long conditions tracking each other closely
throughout. As with the Betti gap, the two conditions are hard to tell apart on
this measure.

\begin{figure}[htbp]
    \centering
    \begin{subfigure}[b]{0.9\linewidth}
        \centering
        \includegraphics[width=\linewidth]{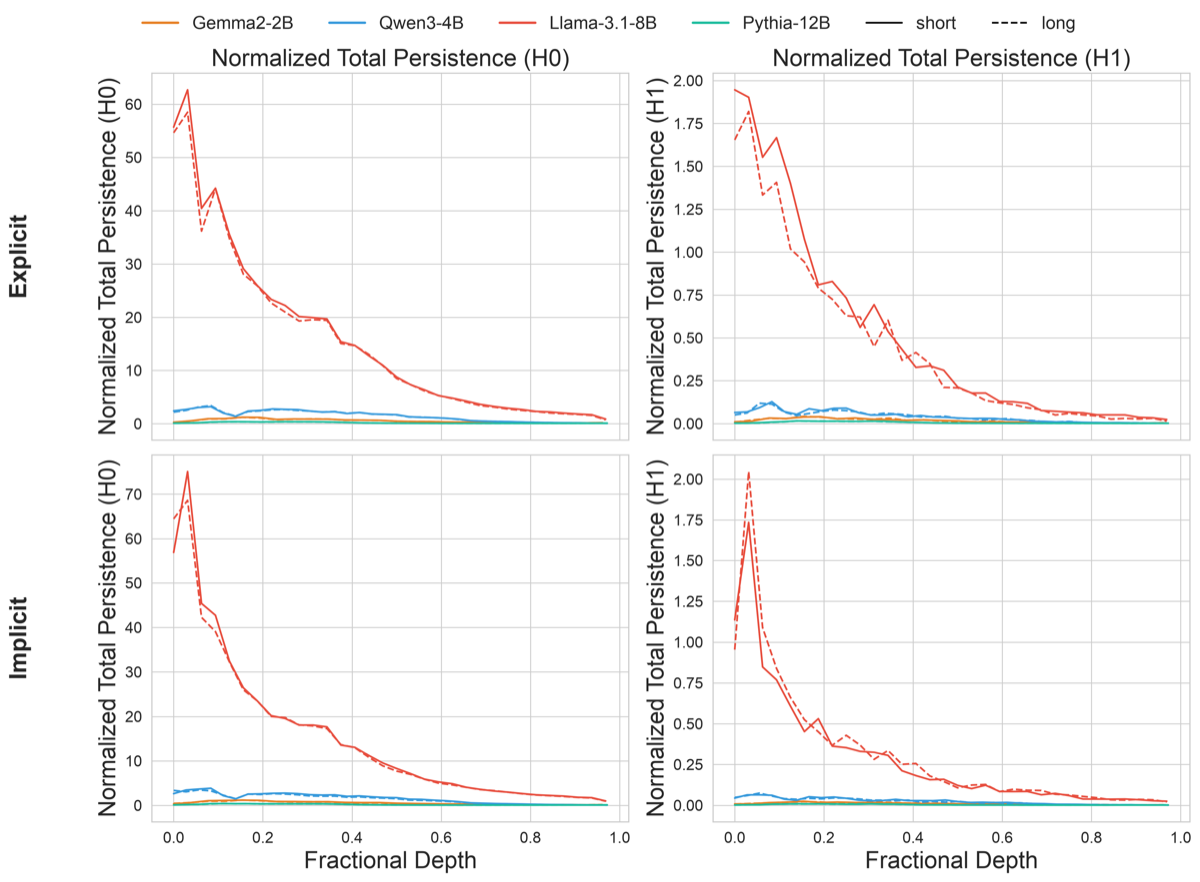}
        \caption{Across heterogeneous model families.}
        \label{fig:ph_normalized_across}
    \end{subfigure}

    \vspace{0.3cm}

    \begin{subfigure}[b]{0.9\linewidth}
        \centering
        \includegraphics[width=\linewidth]{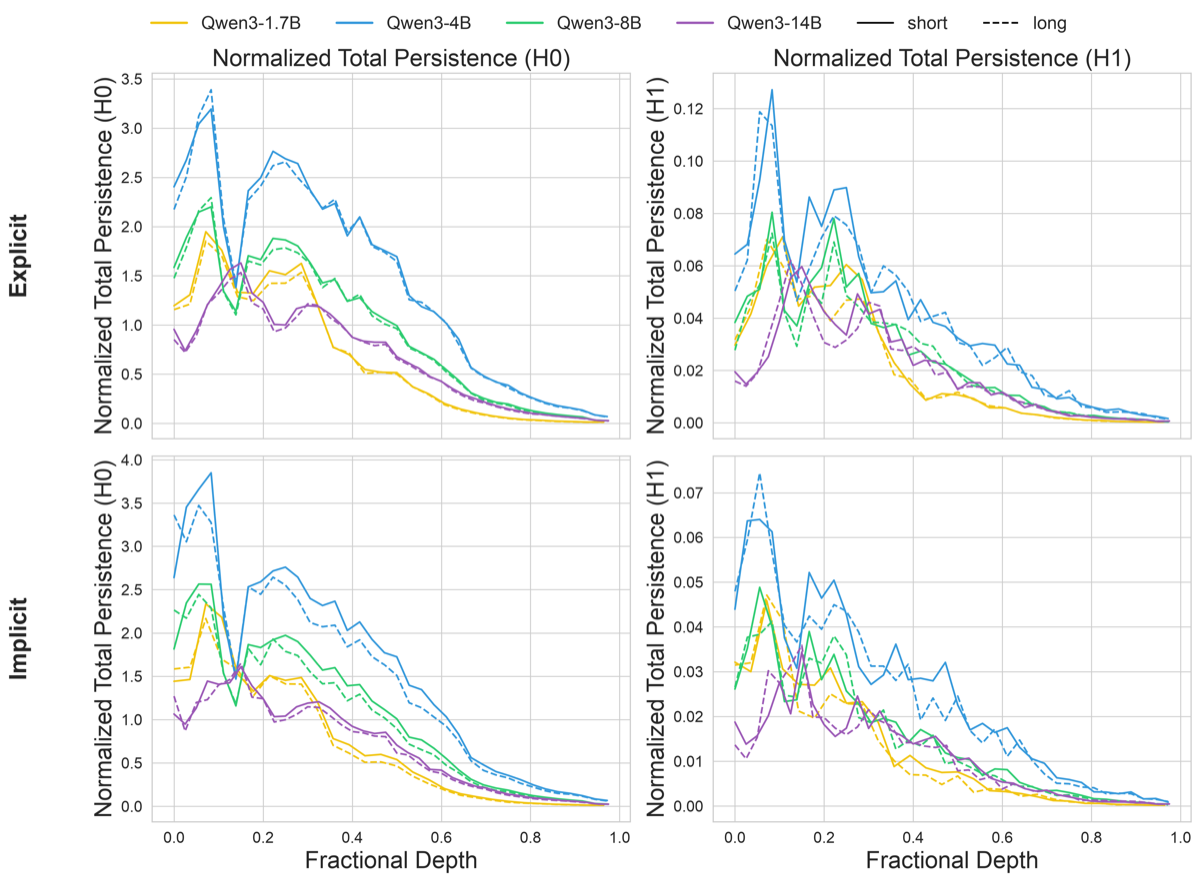}
        \caption{Within the Qwen3 model family.}
        \label{fig:ph_normalized_within}
    \end{subfigure}

    \caption{Total persistence normalized by mean activation norm, across
    normalized layer depth. After the norm confound is removed, topological
    structure peaks early and decays toward the output.}
    \label{fig:ph_normalized}
\end{figure}

\paragraph{Norm-normalized total persistence.}
Total persistence --- the sum of all finite feature lifetimes --- is a magnitude
quantity and therefore directly exposed to the norm blow-up documented in
Table~\ref{tab:raw_norm}. Figure~\ref{fig:ph_normalized} shows it after dividing
out the mean activation norm. Once the confound is removed the picture inverts
relative to the raw measure: persistence \emph{peaks early}, at roughly
$0.1$--$0.25$ fractional depth, and decays to near zero by the output. The same
``balloons in scale, flattens in topology'' pattern appears in the
susceptibility diagnostics of
Appendix~\ref{app:separability}, and the agreement between two independently
computed quantities is the reason we treat the raw late-layer growth as an
artifact rather than a finding.

\begin{figure}[htbp]
    \centering
    \begin{subfigure}[b]{0.9\linewidth}
        \centering
        \includegraphics[width=\linewidth]{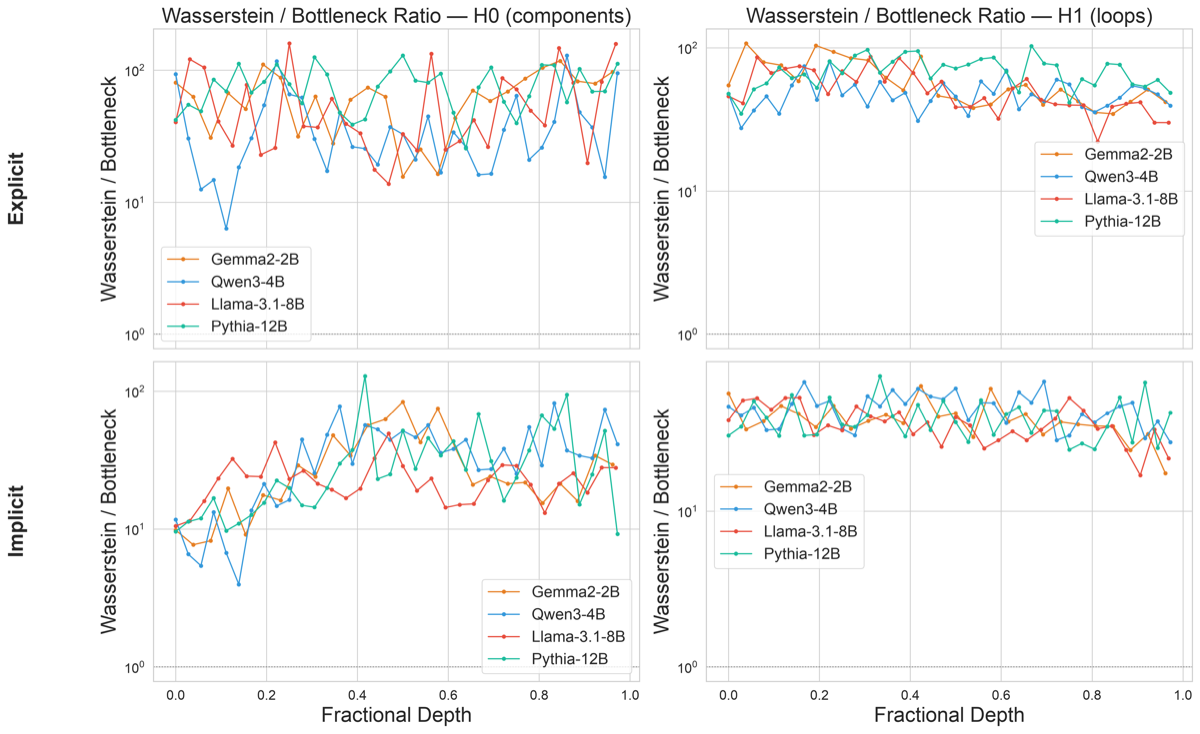}
        \caption{Across heterogeneous model families.}
        \label{fig:ph_ratio_across}
    \end{subfigure}

    \vspace{0.3cm}

    \begin{subfigure}[b]{0.9\linewidth}
        \centering
        \includegraphics[width=\linewidth]{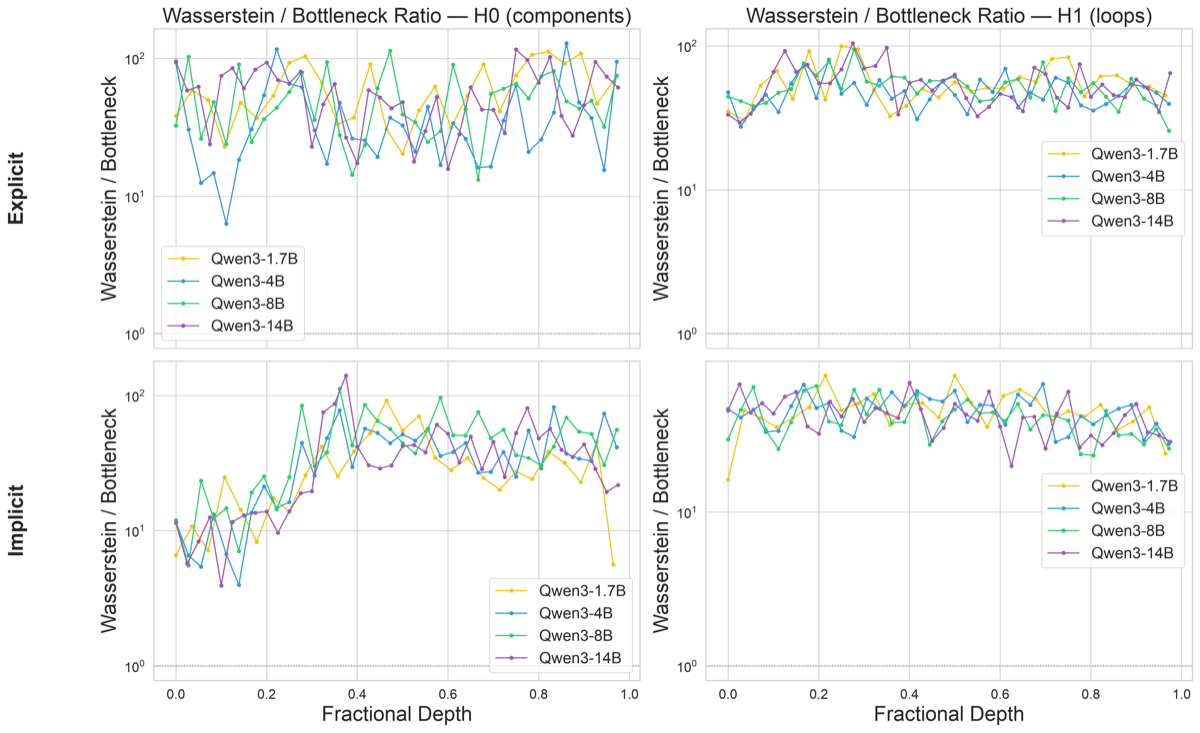}
        \caption{Within the Qwen3 model family.}
        \label{fig:ph_ratio_within}
    \end{subfigure}

    \caption{Ratio of Wasserstein to bottleneck distance across normalized layer
    depth, on a logarithmic axis. The ratio is bounded below by one; values well
    above one indicate that short-to-long topological divergence is distributed
    across many features rather than concentrated in a single one.}
    \label{fig:ph_ratio}
\end{figure}

\paragraph{Wasserstein-to-bottleneck ratio.}
The bottleneck distance measures the single worst feature-matching discrepancy
between two persistence diagrams, while the Wasserstein distance aggregates
mismatch over all features. Since the bottleneck distance is bounded above by
the Wasserstein distance, their ratio is at least one, and its magnitude is
diagnostic: a ratio near one means the divergence between the short and long
diagrams is dominated by a single feature, which would be the signature of an
abrupt topological transition, whereas a large ratio means the divergence is
spread across many small differences. Figure~\ref{fig:ph_ratio} shows the ratio
on a logarithmic axis. It remains well above one throughout, so the
short-to-long topological divergence is consistently distributed rather than
concentrated in one feature --- a further piece of evidence, from an independent
direction, that the transition is a smooth crossover.

Across all four diagnostics, no architecture produces a genuinely bimodal or
high-loop topology. Taken together with the low-rank probe result of
Appendix~\ref{app:probe_quality}, the picture is consistent: the horizon signal
lives in a low-dimensional, approximately linear subspace rather than in the
loop structure of the representation manifold.

\clearpage

\subsection{Intrinsic Dimensionality and Rank}
\label{app:geometry_intrinsic_dim}

Four quantities were measured per layer and separately for short-horizon,
long-horizon, and ambiguous items.

\paragraph{TwoNN intrinsic dimension\citep{twonninstrinsicrank}.}
For each point $i$ in a layer's token-level residual cloud, let $r_1(i)$ and
$r_2(i)$ be the distances to its first and second nearest neighbours and
$\mu_i = r_2(i) / r_1(i) \ge 1$. Under locally uniform density $\mu$ is
Pareto-distributed with tail index equal to the intrinsic dimension, giving the
maximum-likelihood estimate
\begin{equation}
    d_{\text{2NN}} = \frac{n}{\sum_{i=1}^{n} \log \mu_i} ,
    \label{eq:twonn}
\end{equation}
taken over the $n$ points with $r_1(i) > 0$. It is local and neighbour-based,
requiring no spectral decomposition.

\paragraph{Stable rank\citep{ipsen2024stablerankintrinsicdimension}.}
With $\sigma_1 \ge \dots \ge \sigma_m$ the singular values of the
condition-level residual matrix $\mathbf{A}$,
\begin{equation}
    \operatorname{srank}(\mathbf{A})
    = \frac{\lVert \mathbf{A} \rVert_F^2}{\lVert \mathbf{A} \rVert_2^2}
    = \frac{\sum_{i} \sigma_i^2}{\sigma_1^2} ,
    \label{eq:stable_rank}
\end{equation}
a continuous relaxation of matrix rank measuring how far the spectrum spreads
beyond its leading direction.

\paragraph{SVD effective rank\citep{svdeffectiverank}.}
Normalizing the same singular values to $p_i = \sigma_i / \sum_j \sigma_j$ and
exponentiating their Shannon entropy\citep{shannon1948mathematical},
\begin{equation}
    \operatorname{erank}(\mathbf{A}) = \exp\!\left( -\sum_{i} p_i \log p_i \right) ,
    \label{eq:effective_rank}
\end{equation}
which counts the directions carrying comparable spectral mass and returns the
true rank when the spectrum is flat. The remaining measure, the participation
ratio\citep{participationratio}, is
$\bigl(\sum_i \lambda_i\bigr)^2 \big/ \sum_i \lambda_i^2$ with
$\lambda_i = \sigma_i^2$. Equations~\ref{eq:stable_rank}
and~\ref{eq:effective_rank} are global, spectral quantities, whereas
Eq.~\ref{eq:twonn} is local; this distinction is what the divergence below turns
on.

\begin{figure}[htbp]
    \centering
    \begin{subfigure}[b]{0.9\linewidth}
        \centering
        \includegraphics[width=\linewidth]{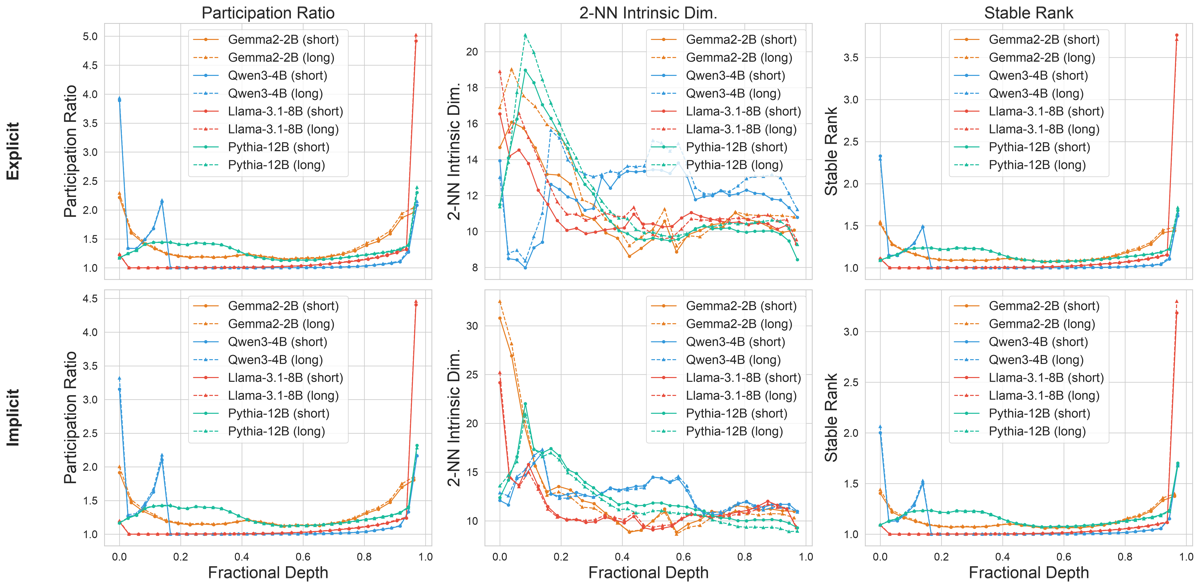}
        \caption{Across heterogeneous model families.}
        \label{fig:intrinsic_dim_across}
    \end{subfigure}

    \vspace{0.3cm}

    \begin{subfigure}[b]{0.9\linewidth}
        \centering
        \includegraphics[width=\linewidth]{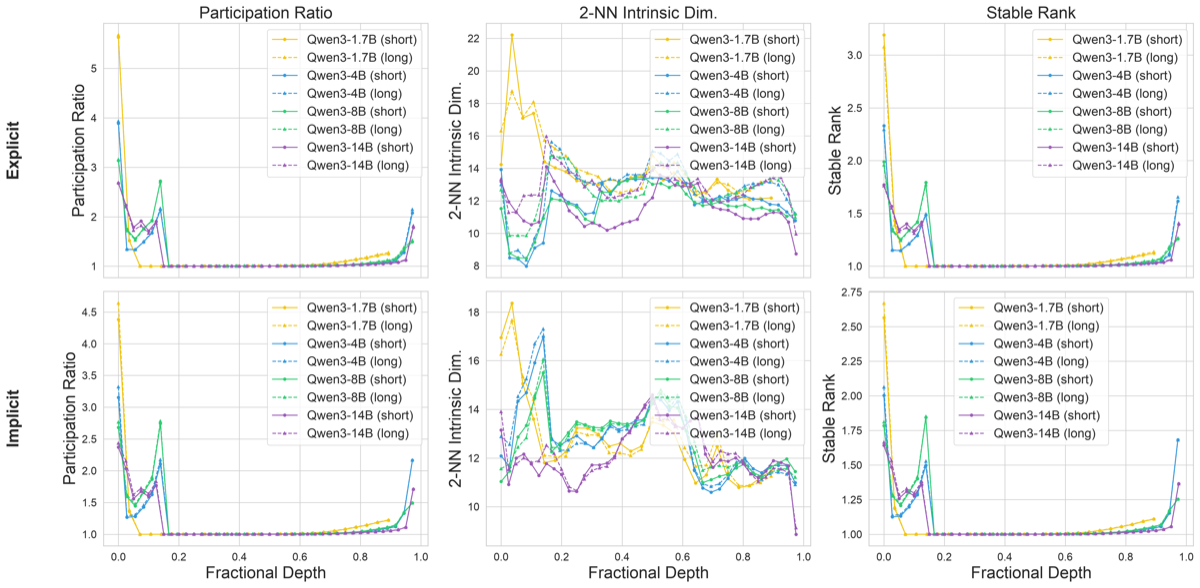}
        \caption{Within the Qwen3 model family.}
        \label{fig:intrinsic_dim_within}
    \end{subfigure}

    \caption{TwoNN intrinsic dimension and participation ratio across normalized
    layer depth. Local intrinsic dimension compresses with depth in most models,
    while the spectral-concentration measures follow a U-shaped profile.}
    \label{fig:intrinsic_dim}
\end{figure}

\begin{figure}[htbp]
    \centering
    \begin{subfigure}[b]{0.9\linewidth}
        \centering
        \includegraphics[width=\linewidth]{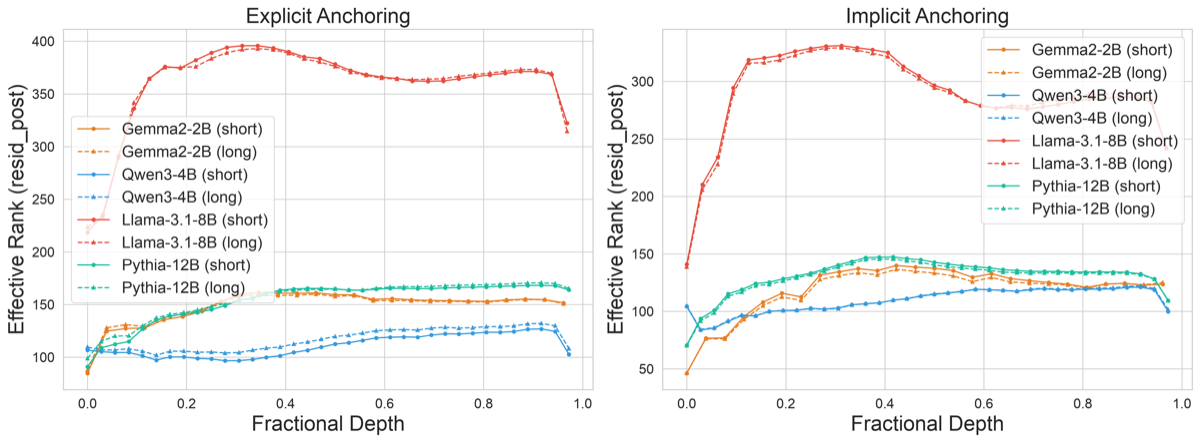}
        \caption{Across heterogeneous model families.}
        \label{fig:rank_divergence_across}
    \end{subfigure}

    \vspace{0.3cm}

    \begin{subfigure}[b]{0.9\linewidth}
        \centering
        \includegraphics[width=\linewidth]{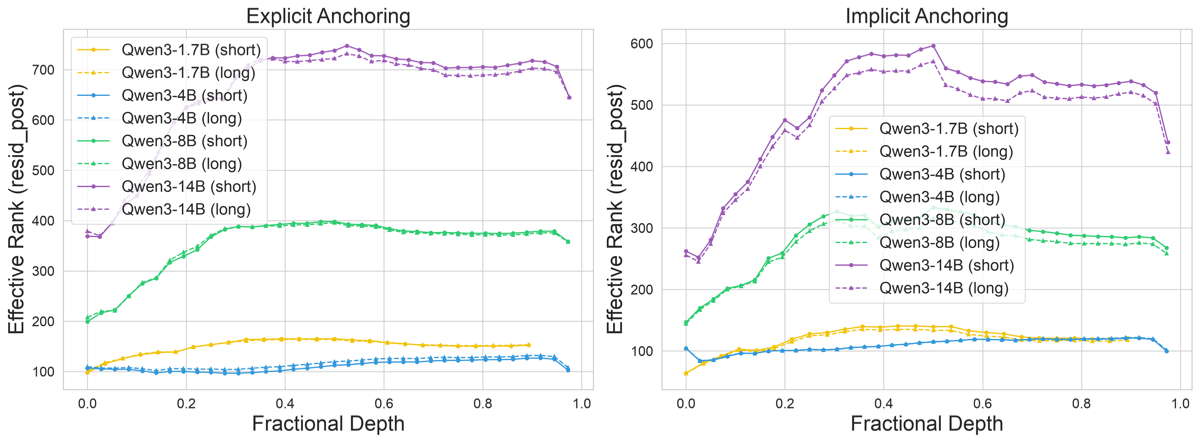}
        \caption{Within the Qwen3 model family.}
        \label{fig:rank_divergence_within}
    \end{subfigure}

    \caption{Effective-rank divergence between short- and long-horizon
    representations across normalized layer depth. Global effective rank rises
    with depth, the mirror image of the falling local intrinsic dimension.}
    \label{fig:rank_divergence}
\end{figure}

The headline is a divergence between local and global dimensionality: local
dimensionality \emph{compresses} with depth in five of seven models, while
global effective rank \emph{rises}. Deep layers pack points into
lower-dimensional local neighbourhoods even as the overall covariance spectrum
spreads out. The two metrics probe different scales of the geometry and should
not be conflated. Figures~\ref{fig:intrinsic_dim} and~\ref{fig:rank_divergence}
show both against depth.

Table~\ref{tab:twonn} reports TwoNN intrinsic dimension under explicit
anchoring. Every architecture places long-horizon items at higher local
dimension than short-horizon items, with gaps between $+0.30$ and $+1.38$, so
the effect is architecture-general rather than a Qwen artifact. Within the Qwen
family the gap scales cleanly and monotonically with model size, from $+0.30$ at
1.7B to $+1.38$ at 14B --- the clearest monotone scale law in this section.
Under implicit anchoring the same gap collapses to approximately $+0.06$ at all
four sizes, so the scaling law is specific to explicit anchoring.

\begin{table}[htbp]
\centering
\caption{TwoNN intrinsic dimension under explicit anchoring, averaged over each
third of the network, with the Spearman correlation against depth for
short-horizon items and the long-minus-short gap. Every model places
long-horizon items at higher local dimension.}
\label{tab:twonn}
\begin{tabular}{|l|l|l|r|r|}
\hline
\textbf{Model} & \textbf{Short (e/m/l)} & \textbf{Long (e/m/l)} & \textbf{$\rho_{\text{short}}$} & \textbf{Long $-$ Short} \\ \hline
Qwen3-1.7B   & 15.5 / 13.5 / 12.5 & 16.0 / 13.4 / 13.0 & $-0.83$ & $+0.30$ \\ \hline
Qwen3-4B     & 10.7 / 13.1 / 11.9 & 12.0 / 13.8 / 12.5 & $+0.10$ & $+0.89$ \\ \hline
Qwen3-8B     & 10.6 / 12.9 / 11.6 & 12.4 / 13.0 / 12.5 & $+0.13$ & $+0.98$ \\ \hline
Qwen3-14B    & 11.7 / 11.8 / 11.2 & 13.2 / 13.1 / 12.5 & $-0.05$ & $+1.38$ \\ \hline
Llama-3.1-8B & 12.1 / 10.5 / 10.4 & 13.5 / 10.4 / 10.6 & $-0.35$ & $+0.52$ \\ \hline
Gemma2-2B    & 13.5 /  9.5 / 10.4 & 15.5 / 10.0 / 10.6 & $-0.53$ & $+0.92$ \\ \hline
Pythia-12B   & 15.1 /  9.9 /  9.9 & 15.8 / 10.3 / 10.3 & $-0.73$ & $+0.53$ \\ \hline
\end{tabular}
\end{table}

The participation ratio and stable rank are nearly identical between the two
horizon classes, differing by at most $0.03$ everywhere, so the horizon does not
change how concentrated the spectrum is --- only where local neighbourhoods sit.
Both follow a U-shaped depth profile: high at the embeddings, a minimum near
unity mid-network, and a slight rebound at the output. The shape is
scale-invariant across the Qwen sweep. Two models depart from it. Llama-3.1-8B
inverts the U into a monotone late rise, holding minimal concentration through
the early and middle layers and expanding sharply only in the final third.
Pythia-12B is the one model trending downward, with concentration highest early
and eroding with depth, consistent with it also compressing local intrinsic
dimension hardest.

Table~\ref{tab:anova} reports the fraction of participation-ratio variance
explained by the three-way condition factor within each depth third. The result
is a sharp condition asymmetry. Under implicit anchoring the horizon is the
dominant driver of dimensionality, explaining roughly four-fifths of the
variance at every depth in every model, with remarkable uniformity across
architectures and scales. Under explicit anchoring the effect is far weaker and
much more variable, ranging from $0.02$ to $0.45$, and within the Qwen family it
is scale-dependent in a non-monotone way: strong at 1.7B, nearly absent at 4B
and 8B, partly recovering at 14B. The explicit-side value is the most
architecture-divergent quantity in this section.

\begin{table}[htbp]
\centering
\caption{Fraction of participation-ratio variance explained by the three-way
condition factor (short, long, ambiguous), reported per depth third. Implicit
anchoring makes the horizon the dominant axis of dimensional variation in every
model; explicit anchoring does not.}
\label{tab:anova}
\begin{tabular}{|l|c|c|}
\hline
\textbf{Model} & \textbf{Explicit (early/mid/late)} & \textbf{Implicit (early/mid/late)} \\ \hline
Qwen3-1.7B   & 0.45 / 0.36 / 0.33 & 0.84 / 0.79 / 0.78 \\ \hline
Qwen3-4B     & 0.08 / 0.02 / 0.03 & 0.79 / 0.74 / 0.77 \\ \hline
Qwen3-8B     & 0.05 / 0.07 / 0.10 & 0.79 / 0.77 / 0.80 \\ \hline
Qwen3-14B    & 0.12 / 0.22 / 0.25 & 0.78 / 0.78 / 0.80 \\ \hline
Llama-3.1-8B & 0.38 / 0.39 / 0.36 & 0.80 / 0.81 / 0.82 \\ \hline
Gemma2-2B    & 0.40 / 0.37 / 0.36 & 0.81 / 0.80 / 0.77 \\ \hline
Pythia-12B   & 0.13 / 0.31 / 0.15 & 0.79 / 0.76 / 0.71 \\ \hline
\end{tabular}
\end{table}

The mechanism behind this asymmetry is visible in the baseline that the horizon
marker acts upon. Under implicit anchoring the ambiguous, no-horizon Qwen
representations collapse to an intrinsic dimension of approximately two at every
size and every depth, and supplying the horizon restores them to roughly twelve
to fourteen dimensions. Under explicit anchoring the no-horizon baseline is
already full-dimensional at twelve to fifteen, and adding the horizon marker
mildly \emph{compresses} local dimension instead. Explicit situational text
pre-loads the dimensionality that an implicit prompt only acquires once the
temporal horizon is supplied. The implicit collapse is strongest in the Qwen
family; the non-Qwen models retain higher ambiguous baselines and correspondingly
smaller horizon-induced changes. We note that the TwoNN means are heavy-tailed,
so the implicit magnitudes should be read as approximate, though the direction is
robust and agrees with both the participation-ratio deltas and
Table~\ref{tab:anova}.

Two data gaps affect this subsection. Pythia-12B has no \texttt{resid\_mid} rank
files, so it is excluded from any mid-block rank comparison, and the Qwen3-1.7B
intrinsic-dimension run covers 26 of its 28 layers.

\subsection{Steering Supplements}
\label{app:geometry_steering}

This subsection reports the steering results not shown in the main text: the
sensitivity of the model to the steering coefficient, and a control experiment
that motivates the causal readout used throughout.

\paragraph{Steering coefficient sensitivity.}
Sweeping the steering coefficient from zero to one in ten steps and fitting a
two-parameter logistic per model, condition, and layer gives essentially exact
fits, with a median $R^2$ of $0.99999$ across all 468 fits. The sigmoid is
therefore the correct functional form, not merely a visual impression. Two
parameters summarize each fit: the coefficient required to reach probability
one-half, which measures how hard the model must be pushed, and the steepness,
which measures how sharply it switches.

Explicit anchoring is both sharper and more sensitive than implicit, as shown in
Table~\ref{tab:steering_condition}: median steepness is roughly twice as high,
and explicit is the steeper of the two in $99\%$ of layers when paired by model
and layer. Both conditions saturate near probability one, so the model is always
fully steerable --- the difference lies in sharpness and threshold, not in
reachability.

\begin{table}[htbp]
\centering
\caption{Steering-coefficient logistic parameters by anchoring condition, median
across all models and layers. Explicit anchoring produces a sharper switch at a
lower threshold, but both conditions reach full saturation.}
\label{tab:steering_condition}
\begin{tabular}{|l|r|r|}
\hline
\textbf{Quantity} & \textbf{Explicit} & \textbf{Implicit} \\ \hline
Steepness $k$ (median)                & 13.4  & 6.8   \\ \hline
Coefficient at $P = 0.5$ (median)     & 0.33  & 0.47  \\ \hline
Baseline $P(\text{long})$ at $c = 0$  & 0.017 & 0.057 \\ \hline
Saturation $P(\text{long})$ at $c = 1$ & 1.00  & 0.97  \\ \hline
\end{tabular}
\end{table}

\begin{figure}[htbp]
    \centering
    \begin{subfigure}[b]{0.9\linewidth}
        \centering
        \includegraphics[width=\linewidth]{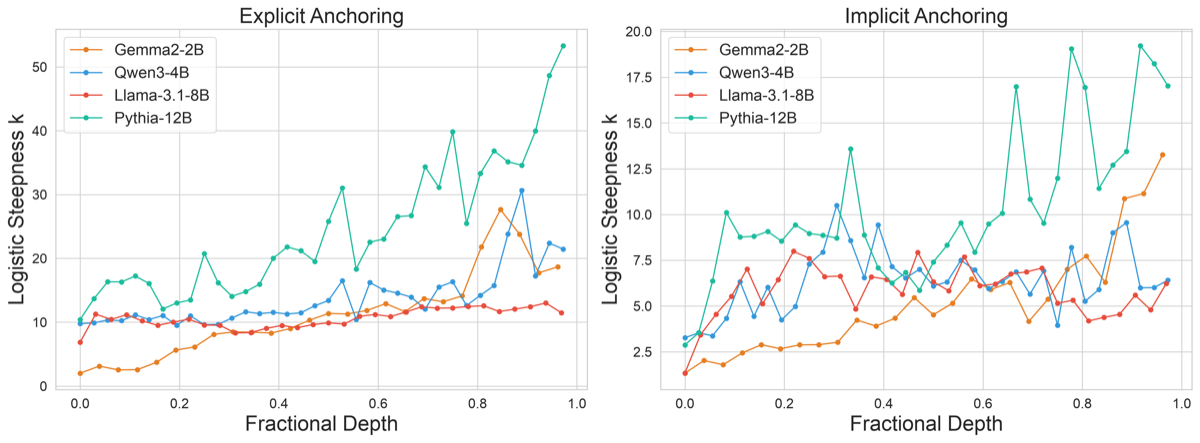}
        \caption{Across heterogeneous model families.}
        \label{fig:steering_steepness_across}
    \end{subfigure}

    \vspace{0.3cm}

    \begin{subfigure}[b]{0.9\linewidth}
        \centering
        \includegraphics[width=\linewidth]{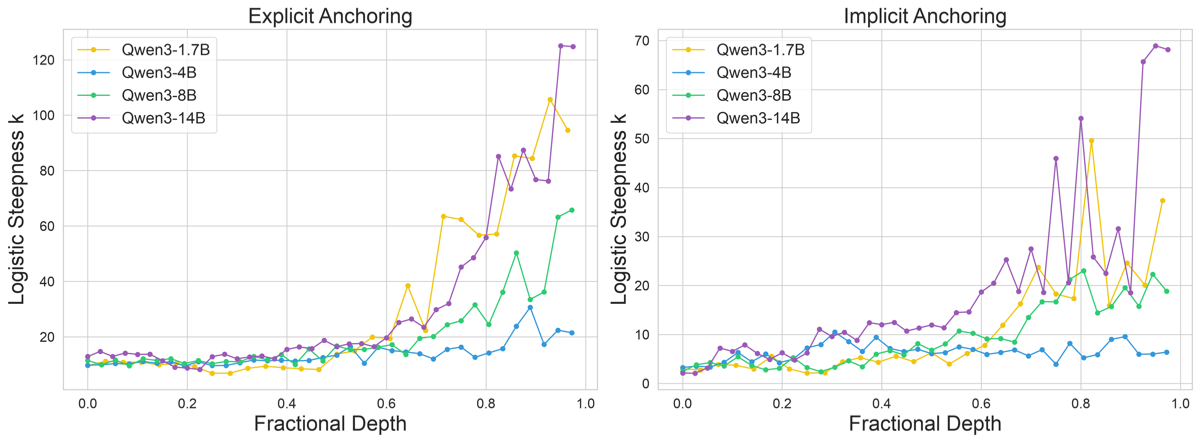}
        \caption{Within the Qwen3 model family.}
        \label{fig:steering_steepness_within}
    \end{subfigure}

    \caption{Logistic steepness of the steering dose-response against normalized
    layer depth. Steepness increases monotonically with depth in every model and
    condition.}
    \label{fig:steering_steepness}
\end{figure}

Steepness increases monotonically with depth in all fourteen model-condition
combinations, with every fitted slope positive and thirteen of fourteen
significant at $p < 0.001$. Table~\ref{tab:steering_depth} shows steepness
roughly tripling from the shallowest to the deepest quartile while the baseline
decays toward zero: deep layers implement a near-step-function switch, shallow
layers a soft ramp. Figure~\ref{fig:steering_steepness} gives the per-model
curves. Within the Qwen family the threshold is roughly scale-invariant while
peak steepness trends upward with size, so scale buys gain rather than a lower
threshold.

\begin{table}[htbp]
\centering
\caption{Logistic steepness and baseline by depth quartile, median across models.
Steepness roughly triples from the shallowest to the deepest quartile in both
conditions.}
\label{tab:steering_depth}
\begin{tabular}{|l|r|r|r|}
\hline
\textbf{Depth quartile} & \textbf{Explicit $k$} & \textbf{Implicit $k$} & \textbf{Explicit baseline} \\ \hline
Early (0--25\%)   & 10.7 &  4.7 & 0.064 \\ \hline
Mid (25--50\%)    & 11.5 &  6.6 & 0.059 \\ \hline
Late (50--75\%)   & 16.8 &  8.4 & 0.021 \\ \hline
Final (75--100\%) & 34.1 & 15.7 & 0.015 \\ \hline
\end{tabular}
\end{table}

One exception should be recorded. Gemma2-2B under implicit anchoring has a
baseline $P(\text{long})$ of $0.93$: it already predicts the long horizon with
no steering applied, so its threshold coefficient is undefined because the curve
never rises through one-half from below. Implicit-anchored Gemma2-2B carries a
strong long-horizon prior against which steering cannot be measured on the same
footing as the other models. This is the same model and condition that shows the
anomalous steering-energy behavior reported in the main text.

\paragraph{Self-referential steering control.}
The steering analysis in the main text reads out the effect of an intervention
by continuing the forward pass and applying the final-layer probe. A cheaper
alternative would be to re-project the injected activation onto the probe of the
same layer, without continuing the forward pass at all. We ran this
self-referential variant as a control, and it is degenerate in two ways that
together justify the causal readout.

\begin{figure}[htbp]
    \centering
    \begin{subfigure}[b]{0.9\linewidth}
        \centering
        \includegraphics[width=\linewidth]{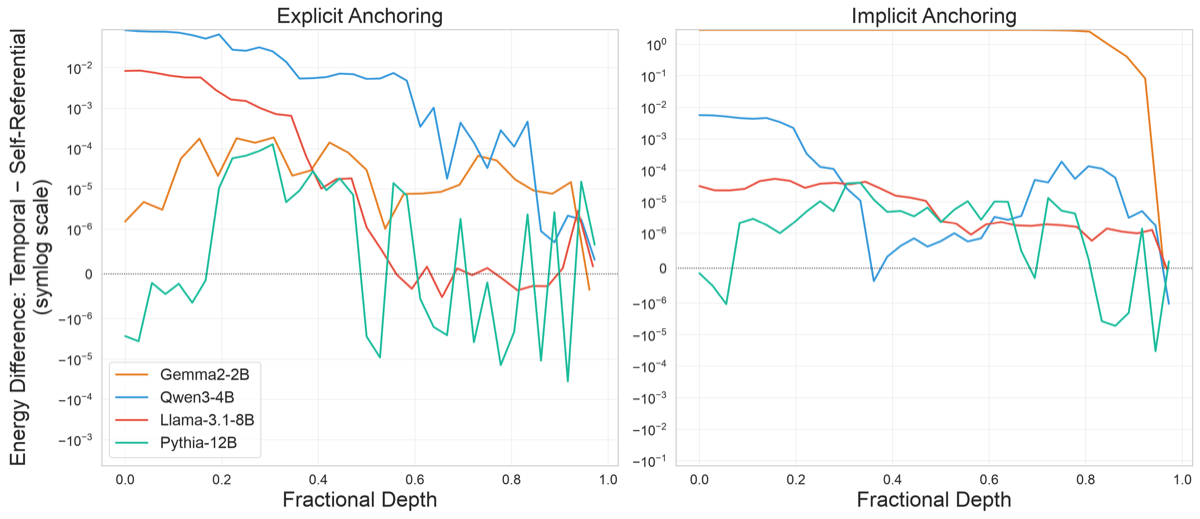}
        \caption{Across heterogeneous model families.}
        \label{fig:selfref_energy_across}
    \end{subfigure}

    \vspace{0.3cm}

    \begin{subfigure}[b]{0.9\linewidth}
        \centering
        \includegraphics[width=\linewidth]{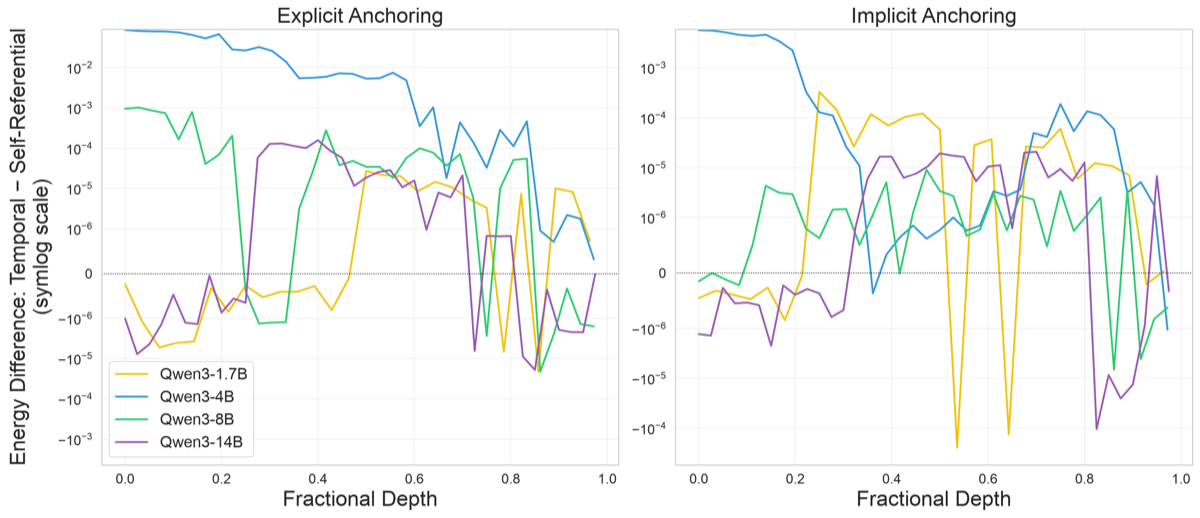}
        \caption{Within the Qwen3 model family.}
        \label{fig:selfref_energy_within}
    \end{subfigure}

    \caption{Log-difference between temporal-axis and self-referential steering
    energies across normalized layer depth. The self-referential readout yields
    energies orders of magnitude smaller, and is path-independent to float
    precision.}
    \label{fig:selfref_energy}
\end{figure}

First, its energies are approximately four orders of magnitude smaller than
those of the causal version, because the probe is re-projecting its own input
rather than measuring a propagated effect. Figure~\ref{fig:selfref_energy} shows
the comparison. Second, and more decisively, the linear and manifold steering
paths differ by roughly $10^{-7}$ everywhere under this readout, which is float
noise. Self-referential steering is entirely path-independent: because the probe
is a single linear map, it can only see the component of its input along its own
weight vector, and the linear and manifold paths share identical endpoints by
construction. The self-referential axis is therefore geometrically trivial, and
any comparison of steering geometries must use the causal readout, which
propagates the intervention through the remaining nonlinear layers.

\end{document}